\documentclass{article}

\title{
	Using Constraints to Discover Sparse and Alternative Subgroup Descriptions
}
\author{
	Jakob Bach~\orcidlink{0000-0003-0301-2798}\\
	\small Independent researcher\footnote{Most of the research for this article was carried out while the author was affiliated with the Karlsruhe Institute of Technology (KIT), Karlsruhe, Germany.}\\
	\small \href{mailto:jakob.bach.ka@gmail.com}{jakob.bach.ka@gmail.com}
}
\date{} 

\usepackage[style=numeric, backend=bibtex]{biblatex}
\usepackage[ruled,linesnumbered,vlined]{algorithm2e} 
\usepackage{amsmath} 
\usepackage{amssymb} 
\usepackage{amsthm} 
\usepackage{booktabs} 
\usepackage{enumitem} 
\usepackage{graphicx} 
\usepackage{multirow} 
\usepackage{orcidlink} 
\usepackage{subcaption} 
\usepackage{hyperref} 

\newtheorem{proposition}{Proposition}
\theoremstyle{definition}
\newtheorem{definition}{Definition}

\begin{document}

\maketitle

\begin{abstract}
Subgroup-discovery methods allow users to obtain simple descriptions of interesting regions in a dataset.
Using constraints in subgroup discovery can enhance interpretability even further.
In this article, we focus on two types of constraints:
First, we limit the number of features used in subgroup descriptions, making the latter sparse.
Second, we propose the novel optimization problem of finding alternative subgroup descriptions, which cover a similar set of data objects as a given subgroup but use different features.
We describe how to integrate both constraint types into heuristic subgroup-discovery methods.
Further, we propose a novel Satisfiability Modulo Theories (SMT) formulation of subgroup discovery as a white-box optimization problem, which allows solver-based search for subgroups and is open to a variety of constraint types.
Additionally, we prove that both constraint types lead to an $\mathcal{NP}$-hard optimization problem.
Finally, we employ 27 binary-classification datasets to compare algorithmic and solver-based search for unconstrained and constrained subgroup discovery.
We observe that heuristic search methods often yield high-quality subgroups within a short runtime, also in scenarios with constraints.
\end{abstract}

\section{Introduction}
\label{sec:csd:introduction}

\paragraph{Motivation}

The interpretability of prediction models has gained importance in recent years~\cite{carvalho2019machine, molnar2020interpretable}.
There are various ways to foster interpretability in machine-learning pipelines.
In particular, some machine-learning models are simple enough to be intrinsically interpretable~\cite{carvalho2019machine}, like subgroup descriptions.
Subgroup discovery aims to identify `interesting' subsets of a dataset~\cite{atzmueller2015subgroup}, such as data objects sharing a specific class label, that can be characterized by concise conditions on feature values.
Subgroup-discovery methods have recently been employed in various fields, such as chemistry~\cite{li2021subgroup}, medicine~\cite{esnault2020qfinder}, database engineering~\cite{remil2021makes}, decision making~\cite{zuheros2023explainable}, and social sciences~\cite{kiefer2022subgroup}.

\begin{figure}[t]
	\centering
	\includegraphics[width=\textwidth, trim=15 15 15 15, clip]{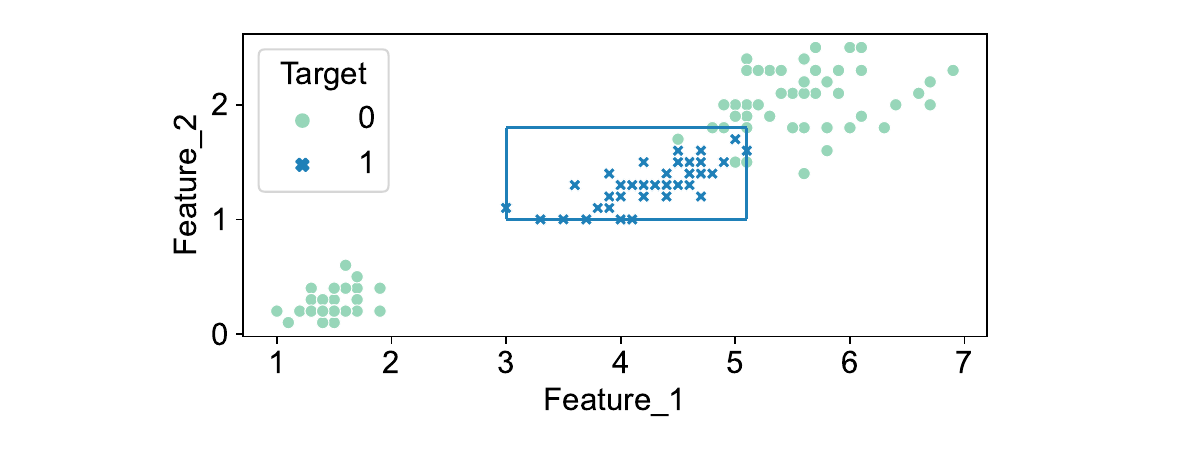}
	\caption{
		Exemplary subgroup description in the form of a rectangle for a dataset with two real-valued features and a binary prediction target.
	}
	\label{fig:csd:exemplary-subgroup}
\end{figure}

Figure~\ref{fig:csd:exemplary-subgroup} displays an exemplary rectangle-shaped subgroup description for a two-dimensional, real-valued dataset with a binary prediction target.
This subgroup is defined by $(\mathit{Feature\_1} \in [3.0, 5.1]) \land (\mathit{Feature\_2} \in [1.0, 1.8])$ and contains a considerably higher fraction of data objects with $\mathit{Target} = 1$ than the complete dataset.
While such subgroup descriptions already tend to be understandable for users, we see further potential to increase interpretability with the help of constraints.

\paragraph{Problem statement}

This article addresses the problem of constrained subgroup discovery.
In particular, we focus on two types of constraints related to the features used in subgroup descriptions:

First, \emph{feature-cardinality constraints} limit the number of selected features, i.e., features used in subgroup descriptions.
Thus, the subgroup descriptions become \emph{sparse}, which increases their interpretability at the potential expense of subgroup quality.
E.g., in Figure~\ref{fig:csd:exemplary-subgroup}, one can use bounds on either feature instead of both to define a subgroup still containing all data objects with $\mathit{Target} = 1$.
Such a description is simpler but covers more data objects with $\mathit{Target} = 0$.
In general, even intrinsically interpretable models may lose interpretability if they involve too many features~\cite{meeng2021real, molnar2020interpretable}.
Further, feature selection~\cite{guyon2003introduction, li2017feature} is common for other machine-learning tasks than subgroup discovery as well.

Second, we formulate constraints to search \emph{alternative subgroup descriptions}:
Given an \emph{original} subgroup, an alternative subgroup description should use different features but cover a similar set of data objects.
E.g., in Figure~\ref{fig:csd:exemplary-subgroup}, one may define a subgroup with an interval on one feature and then try to cover a similar set of data objects with the other feature.
With alternative subgroup descriptions, users obtain different explanations for the same subgroup.
Such alternative explanations are also popular in other explainable-AI techniques like counterfactuals~\cite{mothilal2020explaining, russell2019efficient}, e.g., to enable users to develop and test multiple hypotheses or foster trust in the predictions~\cite{kim2021multi, wang2019designing}.

\paragraph{Related work}

There are various search methods for subgroup discovery, exhaustive~\cite{atzmueller2006sd, bosc2018anytime, grosskreutz2009subgroup, lemmerich2010fast} as well as heuristic~\cite{friedman1999bump, leeuwen2012diverse, mampaey2012efficient, proencca2022robust} ones.
We see research gaps in three aspects:
First, all widely used subgroup-discovery methods are algorithmic in nature and only support a limited set of constraints, as the search routines need to be specifically adapted to particular constraint types.
Second, the number of features used in a subgroup description is a well-known measure for subgroup complexity~\cite{helal2016subgroup, herrera2011overview, ventura2018subgroup}.
However, there is a lack of systematic evaluations for this constraint type, particularly regarding evaluations with different cardinality thresholds and comparing multiple subgroup-discovery methods.
Third, various subgroup-discovery methods yield a diverse set of subgroups rather than only one subgroup, thereby providing alternative solutions~\cite{belfodil2019fssd, bosc2018anytime, leeuwen2012diverse, lemmerich2010fast, lucas2018ssdp+, proencca2022robust}.
However, this notion of alternatives targets at covering different subsets of data objects from the dataset.
In contrast, our notion of alternative subgroup descriptions tries to cover a similar set of data objects as in the original subgroup but with different features in the description.

\paragraph{Contributions}

Our contribution is fivefold:

First, we formalize subgroup discovery as a Satisfiability Modulo Theories (SMT) optimization problem.
This novel white-box formulation admits a solver-based search for subgroups and allows integrating and combining a variety of constraints in a declarative manner.

Second, we formalize two constraint types for this optimization problem, i.e., feature-cardinality constraints and alternative subgroup descriptions.
For the latter, we allow users to control alternatives with two parameters, i.e., the number of alternatives and a dissimilarity threshold.
We integrate both constraint types into our white-box formulation of subgroup discovery.

Third, we describe how to integrate these two constraint types into three existing heuristic search methods and two novel baselines for subgroup discovery.
The latter are faster and simpler than the former, so they may serve as additional reference points for future experimental studies on subgroup discovery.

Fourth, we analyze the time complexity of the subgroup-discovery problem with each of these two constraint types.
In particular, we prove several $\mathcal{NP}$-completeness results and thereby show that finding optimal solutions under these constraint types is computationally challenging.

Fifth, we conduct comprehensive experiments with 27 binary-classification datasets from the Penn Machine Learning Benchmarks (PMLB)~\cite{olson2017pmlb, romano2021pmlb}.
We compare solver-based and seven algorithmic subgroup-discovery methods in different experimental scenarios:
without constraints, with a feature-cardinality constraint, and for searching alternative subgroup descriptions.
In particular, we evaluate the runtime of subgroup discovery and the quality of the discovered subgroups.
We also analyze how the subgroup quality in solver-based search depends on the solver's timeout.
We publish all code and data (cf.~Section~\ref{sec:csd:experimental-design:implementation}).

\paragraph{Experimental results}

In our experimental scenario without constraints, the heuristic search methods yield similar subgroup quality as solver-based search.
On the test set, the heuristics may even be better since they show less overfitting, i.e., a lower gap between training-set quality and test-set quality.
Additionally, the solver-based search is one to two orders of magnitude slower.
Using a solver timeout, a large fraction of the final subgroup quality can be reached in a fraction of the runtime, though this quality is lower than for equally fast heuristics.

With a feature-cardinality constraint, heuristic search methods are still competitive quality-wise compared to solver-based search.
Further, subgroups that only use a few features show relatively high quality compared to unconstrained subgroups.
I.e., there is a decreasing marginal utility in selecting more features.
Additionally, feature-cardinality constraints reduce overfitting.

For alternative subgroup descriptions, heuristics also yield similar quality as solver-based search.
Our two user parameters for alternatives control the solutions as expected:
The similarity to the original subgroup and the quality of the alternatives decrease for more alternatives and a higher dissimilarity threshold.

\paragraph{Outline}

Section~\ref{sec:csd:fundamentals} introduces fundamentals.
Section~\ref{sec:csd:baselines} proposes two baselines for subgroup discovery.
Section~\ref{sec:csd:approach} describes and analyzes constrained subgroup discovery.
Section~\ref{sec:csd:experimental-design} outlines our experimental design, while Section~\ref{sec:csd:evaluation} presents the experimental results.
Section~\ref{sec:csd:related-work} reviews related work.
Section~\ref{sec:csd:conclusion} concludes and discusses future work.
Appendix~\ref{sec:csd:appendix} contains supplementary materials.

\paragraph{Other versions}

There is a conference version of this article~\cite{bach2025subgroup}, which retains the most important experimental and theoretical results.
It evaluates the same run of the experimental pipeline as this arXiv version (v2).

The dissertation~\cite{bach2025leveraging} contains a shortened rendition of the first arXiv version, though longer than the conference version.
It lacks the comparison against the exhaustive algorithmic search methods \emph{BSD} and \emph{SD-Map}.

Appendix~\ref{sec:csd:appendix:further-encodings:smt-categorical}, \ref{sec:csd:appendix:further-encodings:max-sat}, and~\ref{sec:csd:appendix:competitor-runtime} appear exclusively in this arXiv version.

\section{Fundamentals of Subgroup Discovery}
\label{sec:csd:fundamentals}

In this section, we describe fundamentals.
First, we introduce the optimization problem of subgroup discovery (cf.~Section~\ref{sec:csd:fundamentals:problem}).
Second, we describe common algorithmic search methods to solve this problem (cf.~Section~\ref{sec:csd:fundamentals:search}).

\subsection{Problem of Subgroup Discovery}
\label{sec:csd:fundamentals:problem}

\paragraph{Context}

In general, subgroup discovery involves finding descriptions of interesting subsets of a dataset~\cite{atzmueller2015subgroup}.
There are multiple options to define the type of dataset, the kind of subgroup description, and the criterion of interestingness.
In the following, we formalize the notion of subgroup discovery that we tackle in this article.
For broader surveys, see~\cite{atzmueller2015subgroup, helal2016subgroup, herrera2011overview, ventura2018subgroup}.

\paragraph{Dataset}

We focus on tabular, real-valued data.
In particular, $X \in \mathbb{R}^{m \times n}$ stands for a dataset in the form of a matrix.
Each row is a data object, and each column is a feature.
We assume that categorical features have been made numeric, e.g., via a one-hot or an ordinal encoding~\cite{matteucci2023benchmark}.
There are also subgroup-discovery methods that only process categorical data and require continuous features to be discretized~\cite{herrera2011overview, meeng2021real}.
$X_{i \cdot} \in \mathbb{R}^n$ denotes the values of all features for the $i$-th data object,
while $X_{\cdot j} \in \mathbb{R}^m$ denotes the values of the $j$-th feature for all data objects.
$y \in Y^m$ represents the prediction target with domain $Y$, e.g., $Y=\{0,1\}$ for binary classification or $Y=\mathbb{R}$ for regression.
To harmonize formalization and evaluation, we focus on binary-classification scenarios in this article.
In general, one may also conduct subgroup discovery in multi-class, multi-target, or regression scenarios~\cite{atzmueller2015subgroup}.

\paragraph{Subgroup (description)}

A subgroup description typically comprises a conjunction of conditions on individual features~\cite{meeng2021real}.
For real-valued data, the conditions constitute intervals.
Thus, a subgroup description defines a hyperrectangle.
In particular, the subgroup description comprises a lower and upper bound for each feature.
The bounds for a feature may also be infinite to leave it unrestricted.
A data object resides in the subgroup if all its feature values are in the intervals formed by lower and upper bounds:
\begin{definition}[Subgroup (description)]
	Given a dataset~$X \in \mathbb{R}^{m \times n}$, a \emph{subgroup} is described by its lower bounds~$\mathit{lb} \in \{\mathbb{R} \cup \{-\infty, +\infty\}\}^n$ and upper bounds~$\mathit{ub} \in \{\mathbb{R} \cup \{-\infty, +\infty\}\}^n$.
	Data object $X_{i \cdot}$ is a \emph{member} of this subgroup if $\forall j \in \{1, \dots, n\}:~ \left( X_{ij} \geq \mathit{lb}_j \right) \land \left( X_{ij} \leq \mathit{ub}_j \right)$.
	\label{def:csd:subgroup}
\end{definition}
For categorical features, one may replace the inequality comparisons with equality comparisons against categorical feature values~\cite{atzmueller2015subgroup}.

Throughout this article, we often use the terms \emph{subgroup} and \emph{subgroup description} interchangeably.
In a more strict sense, one may use the former term to denote the subgroup's members and the latter for the subgroup's bounds~\cite{atzmueller2015subgroup}.

\paragraph{Subgroup discovery}

Framing subgroup discovery as an optimization problem requires a notion of subgroup quality, i.e., interestingness of the subgroup.
A function $Q(\mathit{lb}, \mathit{ub}, X, y)$ shall return the quality of a subgroup on a particular dataset.
Without loss of generality, we assume a maximization problem:
\begin{definition}[Subgroup discovery]
	Given a dataset~$X \in \mathbb{R}^{m \times n}$ with prediction target~$y \in \{0, 1\}^m$,
	\emph{subgroup discovery} is the problem of finding a subgroup (cf.~Definition~\ref{def:csd:subgroup}) with bounds~$\mathit{lb}, \mathit{ub} \in \{\mathbb{R} \cup \{-\infty, +\infty\}\}^n$ that maximizes a given notion of subgroup quality~$Q(\mathit{lb}, \mathit{ub}, X, y)$.
	\label{def:csd:subgroup-discovery}
\end{definition}
While this definition refers to one subgroup, some subgroup-discovery methods return a set of subgroups~\cite{atzmueller2015subgroup}.

\paragraph{Subgroup quality}

For binary-classification scenarios, interesting subgroups should typically contain many data objects from one class but few from the other class.
While traditional classification tries to characterize the dataset globally, subgroup discovery follows a local paradigm, i.e., focuses on the data objects in the subgroup~\cite{meeng2021real}.
Without loss of generality, we assume that the class with label~`1' is the class of interest, also called \emph{positive} class.
Weighted Relative Accuracy (WRAcc)~\cite{lavravc1999rule} is a popular metric for subgroup quality~\cite{meeng2021real}:
\begin{equation}
	\text{WRAcc} = \frac{m_b}{m} \cdot \left( \frac{m_b^+}{m_b} - \frac{m^+}{m} \right)
	\label{eq:csd:wracc}
\end{equation}
Besides the total number of data objects~$m$, this metric considers the number of positive data objects~$m^+$, the number of data objects in the subgroup~$m_b$, and the number of positive data objects in the subgroup~$m_b^+$.
In particular, WRAcc is the product of two factors:
$m_b / m$ expresses the generality of the subgroup as the relative frequency of subgroup membership.
The second factor measures the relative accuracy of the subgroup, i.e., the difference in the relative frequency of the positive class between the subgroup and the whole dataset.
If the subgroup contains the same fraction of positive data objects as the whole dataset, WRAcc is zero.
The theoretical maximum and minimum of WRAcc depend on the class frequencies in the dataset.
In particular, the maximum WRAcc for a dataset equals the product of the relative frequencies of positive and negative data objects in the dataset~\cite{mathonat2021anytime}:
\begin{equation}
	\text{WRAcc}_{\text{max}} = \frac{m^+}{m} \cdot \left( 1 - \frac{m^+}{m} \right)
	\label{eq:csd:wracc-max}
\end{equation}
This maximum is reached if all positive data objects are in the subgroup and all negative data objects are outside, i.e., $m_b^+ = m_b = m^+$.
Depending on the feature values of the dataset, a corresponding subgroup description may not exist.
Further, the maximum value of this expression is 0.25 if both classes occur with equal frequency but becomes smaller the more imbalanced the classes are.
Thus, it makes sense to normalize WRAcc when working with datasets with different class frequencies.
One normalization, which we use in our experiments, is a max-normalization to the range $[-1, 1]$~\cite{mathonat2021anytime}:
\begin{equation}
	\text{nWRAcc} = \frac{\text{WRAcc}}{\text{WRAcc}_{\text{max}}} = \frac{m_b^+ \cdot m - m^+ \cdot m_b}{m^+ \cdot (m - m^+)}
	\label{eq:csd:wracc-normalized}
\end{equation}
Alternatively, one can also min-max-normalize the range to~$[0, 1]$~\cite{carmona2018unifying, ventura2018subgroup}.

\subsection{Search Methods}
\label{sec:csd:fundamentals:search}

To discover subgroups, there are heuristic search methods (cf.~Section~\ref{sec:csd:fundamentals:search:heuristics}), like PRIM~\cite{friedman1999bump} and Best Interval~\cite{mampaey2012efficient}, as well as exhaustive search methods (cf.~Section~\ref{sec:csd:fundamentals:search:exhaustive}), like SD-Map~\cite{atzmueller2009fast, atzmueller2006sd}, MergeSD~\cite{grosskreutz2009subgroup}, and BSD~\cite{lemmerich2016fast, lemmerich2010fast}.
In this section, we discuss five search methods that are relevant for our experiments; see~\cite{atzmueller2015subgroup, helal2016subgroup, herrera2011overview, ventura2018subgroup} for comprehensive surveys on subgroup-discovery methods.

\subsubsection{Heuristic Search Methods}
\label{sec:csd:fundamentals:search:heuristics}

\begin{algorithm}[p]
	\DontPrintSemicolon
	\KwIn{Dataset~$X \in \mathbb{R}^{m \times n}$, \newline
		Prediction target~$y \in \{0, 1\}^m$, \newline
		Subgroup-quality function~$Q(\mathit{lb}, \mathit{ub}, X, y)$, \newline
		Peeling fraction~$\alpha \in (0, 1)$, \newline
		Support threshold~$\beta_0 \in [0, 1]$
	}
	\KwOut{Subgroup bounds~$\mathit{lb}, \mathit{ub} \in \{\mathbb{R} \cup \{-\infty, +\infty\}\}^n$}
	\BlankLine
	\For(\tcp*[f]{Start with unrestricted subgroup}){$j \leftarrow 1$ \KwTo $n$}{ \label{al:csd:prim:line:initialization-start}
		$(\mathit{lb}^{\text{opt}}_j,~ \mathit{ub}^{\text{opt}}_j) \leftarrow (-\infty, +\infty)$\;
	}
	$Q^{\text{opt}} \leftarrow Q(\mathit{lb}^{\text{opt}}, \mathit{ub}^{\text{opt}}, X, y)$\;
	$(\mathit{lb}^{\text{peel}}, \mathit{ub}^{\text{peel}}) \leftarrow (\mathit{lb}^{\text{opt}},~ \mathit{ub}^{\text{opt}})$\; \label{al:csd:prim:line:initialization-end}
	\While(\tcp*[f]{Support threshold satisfied}){$\frac{m_b}{m} > \beta_0$}{ \label{al:csd:prim:line:stop} \label{al:csd:prim:iteration-start}
		$Q^{\text{cand}} \leftarrow -\infty$\;
		\For{$j \in$ get\_permissible\_feature\_idxs(\dots)}{ \label{al:csd:prim:line:peel-start} \label{al:csd:prim:line:permissible-features}
			$(\mathit{lb},~ \mathit{ub}) \leftarrow (\mathit{lb}^{\text{peel}},~ \mathit{ub}^{\text{peel}})$ \tcp*{Try peeling lower bound}
			$\mathit{lb}_j \leftarrow$ quantile($X_{\cdot j}$, $\mathit{lb}$, $\mathit{ub}$, $\alpha$)\;
			\If{$Q(\mathit{lb}, \mathit{ub}, X, y) > Q^{\text{cand}}$}{
				$(\mathit{lb}^{\text{cand}}, \mathit{ub}^{\text{cand}}) \leftarrow (\mathit{lb}, \mathit{ub})$\;
			}
			$(\mathit{lb},~ \mathit{ub}) \leftarrow (\mathit{lb}^{\text{peel}},~ \mathit{ub}^{\text{peel}})$ \tcp*{Try peeling upper bound}
			$\mathit{ub}_j \leftarrow$ quantile($X_{\cdot j}$, $\mathit{lb}$, $\mathit{ub}$, $1-\alpha$)\;
			\If{$Q(\mathit{lb}, \mathit{ub}, X, y) > Q^{\text{cand}}$}{
				$(\mathit{lb}^{\text{cand}}, \mathit{ub}^{\text{cand}}) \leftarrow (\mathit{lb}, \mathit{ub})$\; \label{al:csd:prim:line:peel-end}
			}
		}
		$(\mathit{lb}^{\text{peel}}, \mathit{ub}^{\text{peel}}) \leftarrow (\mathit{lb}^{\text{cand}}, \mathit{ub}^{\text{cand}})$ \tcp*{Retain best candidate} \label{al:csd:prim:line:best-peel-selection}
		\If(\tcp*[f]{Update optimum}){$Q(\mathit{lb}^{\text{peel}}, \mathit{ub}^{\text{peel}}, X, y) > Q^{\text{opt}}$}{ \label{al:csd:prim:line:opt-check-start}
			$Q^{\text{opt}} \leftarrow Q(\mathit{lb}^{\text{peel}}, \mathit{ub}^{\text{peel}}, X, y)$\;
			$(\mathit{lb}^{\text{opt}},~ \mathit{ub}^{\text{opt}}) \leftarrow (\mathit{lb}^{\text{peel}},~ \mathit{ub}^{\text{peel}})$\; \label{al:csd:prim:line:opt-check-end} \label{al:csd:prim:iteration-end}
		}
	}
	$(\mathit{lb}, \mathit{ub}) \leftarrow (\mathit{lb}^{\text{opt}}, \mathit{ub}^{\text{opt}})$\; \label{al:csd:prim:line:select-optimal-bounds}
	\For(\tcp*[f]{Reset non-excluding bounds}){$j \leftarrow 1$ \KwTo $n$}{ \label{al:csd:prim:line:bounds-infty-start}
		\lIf{$\mathit{lb}_j = \min_{i \in \{1, \dots, m\}} X_{ij}$}{$\mathit{lb}_j \leftarrow -\infty$}
		\lIf{$\mathit{ub}_j = \max_{i \in \{1, \dots, m\}} X_{ij}$}{$\mathit{ub}_j \leftarrow +\infty$} \label{al:csd:prim:line:bounds-infty-end}
	}
	\Return{$\mathit{lb}, \mathit{ub}$}
	\caption{\emph{PRIM} for subgroup discovery.}
	\label{al:csd:prim}
\end{algorithm}

\paragraph{PRIM}

\emph{Patient Rule Induction Method (PRIM)}~\cite{friedman1999bump} is an iterative search algorithm.
In its basic form, it consists of a peeling phase and a pasting phase.
Peeling restricts the bounds of the subgroup iteratively, while pasting expands them.
Algorithm~\ref{al:csd:prim} outlines the peeling phase for finding one subgroup, which is the flavor of PRIM we consider in this article and denote as \emph{PRIM}.
Pasting may have little effect on the subgroup quality and is often left out~\cite{arzamasov2021reds}.
Further, we do not discuss extensions of PRIM like bumping~\cite{friedman1999bump, kwakkel2016improving}, which uses bagging of multiple PRIM runs to improve subgroup quality, or covering~\cite{friedman1999bump}, which returns a sequence of subgroups covering different data objects.

The algorithm \emph{PRIM} starts with a subgroup containing all data objects, which is the initial solution candidate (Lines~\ref{al:csd:prim:line:initialization-start}--\ref{al:csd:prim:line:initialization-end}).
It continues peeling until the current solution candidate contains at most a fraction~$\beta_0$ of data objects (Line~\ref{al:csd:prim:line:stop}).
The support threshold~$\beta_0 \in [0, 1]$ is a user parameter.
The returned subgroup is the optimal solution candidate over all peeling iterations (Line~\ref{al:csd:prim:line:select-optimal-bounds}).
In our \emph{PRIM} implementation, we add a small post-processing step after peeling:
We set \emph{non-excluding bounds} to infinity (Lines~\ref{al:csd:prim:line:bounds-infty-start}--\ref{al:csd:prim:line:bounds-infty-end}).
These are bounds that do not exclude any data objects from the subgroup, i.e., lower/upper bounds that equal the minimum/maximum feature value over all data objects.
There are two reasons behind this post-processing:
First, we ensure that these bounds remain non-excluding for any new data, where global feature minima/maxima may differ.
Second, it becomes easier to see which features are selected in the subgroup description and which are not.

In the iterative peeling procedure (Lines~\ref{al:csd:prim:iteration-start}--\ref{al:csd:prim:iteration-end}), the algorithm generates new solution candidates by trying to restrict each \emph{permissible feature} (Lines~\ref{al:csd:prim:line:peel-start}--\ref{al:csd:prim:line:peel-end}).
In unconstrained subgroup discovery, each feature is permissible, but the function \emph{get\_permissible\_feature\_idxs(\dots)} will become useful once we introduce constraints.
For each Feature~$j$, the algorithm tests a new lower bound at the $\alpha$-quantile of feature values in the subgroup and a new upper bound at the $1-\alpha$-quantile of feature values in the subgroup.
The peeling fraction $\alpha \in (0, 1)$ is a user parameter.
It describes which fraction of data objects gets excluded from the subgroup in each peeling iteration.
Having tested two new bounds for each feature, the algorithm takes the subgroup with the highest associated quality (Line~\ref{al:csd:prim:line:best-peel-selection}) and continues peeling it in the next iteration.
Further, if this solution candidate improves upon the optimal solution candidate from all prior iterations, it is stored as the new optimum (Lines~\ref{al:csd:prim:line:opt-check-start}--\ref{al:csd:prim:line:opt-check-end}).

\begin{algorithm}[p]
	\DontPrintSemicolon
	\KwIn{Dataset~$X \in \mathbb{R}^{m \times n}$, \newline
		Prediction target~$y \in \{0, 1\}^m$, \newline
		Subgroup-quality function~$Q(\mathit{lb}, \mathit{ub}, X, y)$, \newline
		Beam width~$w \in \mathbb{N}$
	}
	\KwOut{Subgroup bounds~$\mathit{lb}, \mathit{ub} \in \{\mathbb{R} \cup \{-\infty, +\infty\}\}^n$}
	\BlankLine
	\For(\tcp*[f]{Initialize beam}){$l \leftarrow 1$ \KwTo $w$}{ \label{al:csd:generic-beam-search:line:initialization-start}
		\For{$j \leftarrow 1$ \KwTo $n$}{
			$(\mathit{lb}^{(\text{beam, } l)}_j,~ \mathit{ub}^{(\text{beam, } l)}_j) \leftarrow (-\infty, +\infty)$ \tcp*{Unrestricted}
		}
		$\mathit{cand\_has\_changed}^{(l)} \leftarrow$ \textbf{true} \tcp*{Subgroup should be updated}
		$Q^{(l)} \leftarrow Q(\mathit{lb}^{(\text{beam, } l)}, \mathit{ub}^{(\text{beam, } l)}, X, y)$\; \label{al:csd:generic-beam-search:line:initialization-end}
	}
	\While(\tcp*[f]{Beam has changed}){$\left( \sum_{l=1}^w \mathit{cand\_has\_changed}^{(l)} \right) > 0$}{ \label{al:csd:generic-beam-search:line:iteration-start} \label{al:csd:generic-beam-search:line:stop}
		$\mathit{prev\_cand\_changed\_idxs} \leftarrow \{l \mid \mathit{cand\_has\_changed}^{(l)} \}$\;
		\For(\tcp*[f]{Create temporary solution candidates}){$l \leftarrow 1$ \KwTo $w$}{
			$(\mathit{lb}^{(\text{cand, } l)},~ \mathit{ub}^{(\text{cand, } l)}) \leftarrow$ $(\mathit{lb}^{(\text{beam, } l)},~ \mathit{ub}^{(\text{beam, } l)})$\;
			$\mathit{cand\_has\_changed}^{(l)} \leftarrow$ \textbf{false}\;
		}
		\For(\tcp*[f]{Prepare beam updates}){$l \in \mathit{prev\_cand\_changed\_idxs}$}{ \label{al:csd:generic-beam-search:line:update-start}
			\For{$j \in$ get\_permissible\_feature\_idxs(\dots)}{ \label{al:csd:generic-beam-search:line:permissible-features}
				evaluate\_subgroup\_updates(\dots) \tcp*{Algorithm~\ref{al:csd:beam-search-subgroup-update} or~\ref{al:csd:best-interval-subgroup-update}} \label{al:csd:generic-beam-search:line:update-end}
			}
		}
		\For(\tcp*[f]{Update beam}){$l \leftarrow 1$ \KwTo $w$}{
			$(\mathit{lb}^{(\text{beam, } l)},~ \mathit{ub}^{(\text{beam, } l)}) \leftarrow$ $(\mathit{lb}^{(\text{cand, } l)},~ \mathit{ub}^{(\text{cand, } l)})$\; \label{al:csd:generic-beam-search:line:iteration-end}
		}
	}
	$l \leftarrow \arg\max_{l \in \{1, \dots, w\}} Q^{(l)}$ \tcp*{Select best subgroup from beam} \label{al:csd:generic-beam-search:line:finalization-start}
	$(\mathit{lb},~ \mathit{ub}) \leftarrow (\mathit{lb}^{(\text{beam, } l)},~ \mathit{ub}^{(\text{beam, } l)})$\;
	\For(\tcp*[f]{Reset non-excluding bounds}){$j \leftarrow 1$ \KwTo $n$}{ \label{al:csd:generic-beam-search:line:bounds-infty-start}
		\lIf{$\mathit{lb}_j = \min_{i \in \{1, \dots, m\}} X_{ij}$}{$\mathit{lb}_j \leftarrow -\infty$}
		\lIf{$\mathit{ub}_j = \max_{i \in \{1, \dots, m\}} X_{ij}$}{$\mathit{ub}_j \leftarrow +\infty$} \label{al:csd:generic-beam-search:line:bounds-infty-end}
	}
	\Return{$\mathit{lb}, \mathit{ub}$}  \label{al:csd:generic-beam-search:line:finalization-end}
	\caption{Generic beam search for subgroup discovery.}
	\label{al:csd:generic-beam-search}
\end{algorithm}

\begin{algorithm}[t]
	\DontPrintSemicolon
	\KwIn{Parameters and variables from Algorithm~\ref{al:csd:generic-beam-search}}
	\KwOut{None; modifies variables from Algorithm~\ref{al:csd:generic-beam-search} in-place}
	$(\mathit{lb},~ \mathit{ub}) \leftarrow (\mathit{lb}^{(\text{beam, } l)},~ \mathit{lb}^{(\text{beam, } l)})$ \tcp*{Next, update lower bound} \label{al:csd:beam-search-subgroup-update:line:lb-start}
	\For{$b \in $ sort(unique(get\_feature\_values($X$, $j$, $\mathit{lb}^{(\text{beam, } l)}$, $\mathit{ub}^{(\text{beam, } l)}$)))}{
		$\mathit{lb}_j \leftarrow b$\;
		\If{$\left( Q(\mathit{lb}, \mathit{ub}, X, y) > \min_{l \in \{1, ..., w\}} Q^{(l)} \right)$ \textbf{and} $(\mathit{lb},~ \mathit{ub}) \notin \{(\mathit{lb}^{(\text{cand, } l)},~ \mathit{ub}^{(\text{cand, } l)}) \mid l \in \{1, \dots, w\}\}$ }{ \label{al:csd:beam-search-subgroup-update:line:lb-replace-start}
			$l \leftarrow \arg\min_{l \in \{1, \dots, w\}} Q^{(l)}$ \tcp*{Replace worst candidate}
			$(\mathit{lb}^{(\text{cand, } l)},~ \mathit{ub}^{(\text{cand, } l)}) \leftarrow (\mathit{lb}, \mathit{ub})$\;
			$\mathit{cand\_has\_changed}^{(l)} \leftarrow$ \textbf{true}\;
			$Q^{(l)} \leftarrow Q(\mathit{lb}, \mathit{ub}, X, y)$\; \label{al:csd:beam-search-subgroup-update:line:lb-replace-end} \label{al:csd:beam-search-subgroup-update:line:lb-end}
		}
	}
	$(\mathit{lb},~ \mathit{ub}) \leftarrow (\mathit{lb}^{(\text{beam, } l)},~ \mathit{lb}^{(\text{beam, } l)})$ \tcp*{Next, update upper bound} \label{al:csd:beam-search-subgroup-update:line:ub-start}
	\For{$b \in $ sort(unique(get\_feature\_values($X$, $j$, $\mathit{lb}^{(\text{beam, } l)}$, $\mathit{ub}^{(\text{beam, } l)}$)))}{
		$\mathit{ub}_j \leftarrow b$\;
		\If{$\left( Q(\mathit{lb}, \mathit{ub}, X, y) > \min_{l \in \{1, ..., w\}} Q^{(l)} \right)$ \textbf{and} $(\mathit{lb},~ \mathit{ub}) \notin \{(\mathit{lb}^{(\text{cand, } l)},~ \mathit{ub}^{(\text{cand, } l)}) \mid l \in \{1, \dots, w\}\}$ }{ \label{al:csd:beam-search-subgroup-update:line:ub-replace-start}
			$l \leftarrow \arg\min_{l \in \{1, \dots, w\}} Q^{(l)}$ \tcp*{Replace worst candidate}
			$(\mathit{lb}^{(\text{cand, } l)},~ \mathit{ub}^{(\text{cand, } l)}) \leftarrow (\mathit{lb}, \mathit{ub})$\;
			$\mathit{cand\_has\_changed}^{(l)} \leftarrow$ \textbf{true}\;
			$Q^{(l)} \leftarrow Q(\mathit{lb}, \mathit{ub}, X, y)$\; \label{al:csd:beam-search-subgroup-update:line:ub-replace-end} \label{al:csd:beam-search-subgroup-update:line:ub-end}
		}
	}
	\caption{evaluate\_subgroup\_updates(\dots) for \emph{Beam Search}.}
	\label{al:csd:beam-search-subgroup-update}
\end{algorithm}

\paragraph{Beam Search}

Beam search is a generic search strategy that is also common in subgroup discovery~\cite{atzmueller2005exploiting}.
It maintains a set of currently best solution candidates, i.e., the beam, which it iteratively updates.
The number of solution candidates in the beam is a user parameter, i.e., the beam width~$w \in \mathbb{N}$.
We outline one way to implement it in Algorithms~\ref{al:csd:generic-beam-search} and~\ref{al:csd:beam-search-subgroup-update}, which we refer to as \emph{Beam Search} in the following.
It is an adapted version of the beam-search implementation in the Python package \emph{pysubgroup}~\cite{lemmerich2019pysubgroup}.

First, the algorithm \emph{Beam Search} initializes the beam by creating $w$ unrestricted subgroups (Lines~\ref{al:csd:generic-beam-search:line:initialization-start}--\ref{al:csd:generic-beam-search:line:initialization-end}).
Further, it stores the quality of each of these subgroups.
Additionally, it records which subgroups changed in the previous iteration (Lines~\ref{al:csd:generic-beam-search:line:iteration-start}--\ref{al:csd:generic-beam-search:line:iteration-end}) of the search.
In particular, it stops once all subgroups in the beam remain unchanged (Line~\ref{al:csd:generic-beam-search:line:stop}).
Subsequently, it returns the best subgroup from the beam (Lines~\ref{al:csd:generic-beam-search:line:finalization-start}--\ref{al:csd:generic-beam-search:line:finalization-end}).
As for Algorithm~\ref{al:csd:prim}, we replace all non-excluding bounds with infinity as a post-processing step.

The main loop (Lines~\ref{al:csd:generic-beam-search:line:iteration-start}--\ref{al:csd:generic-beam-search:line:iteration-end}) updates the beam.
In particular, for each subgroup that changed in the previous iteration, the algorithm creates new solution candidates by attempting to update the bounds of each feature separately (Lines~\ref{al:csd:generic-beam-search:line:update-start}--\ref{al:csd:generic-beam-search:line:update-end}).
There are different options for this update step.
Algorithm~\ref{al:csd:beam-search-subgroup-update} outlines the update procedure for \emph{Beam Search}, while \emph{Best Interval} uses a slightly different one (cf.~Algorithm~\ref{al:csd:best-interval-subgroup-update}).
For \emph{Beam Search}, the procedure tries to refine the lower bound (Lines~\ref{al:csd:beam-search-subgroup-update:line:lb-start}--\ref{al:csd:beam-search-subgroup-update:line:lb-end}) and the upper bound (Lines~\ref{al:csd:beam-search-subgroup-update:line:ub-start}--\ref{al:csd:beam-search-subgroup-update:line:ub-end}) for a given Feature~$j$ separately by replacing it with another feature value from data objects in the subgroup.
In particular, it iterates over all these unique feature values.
Each solution candidate that improves upon the minimum subgroup quality from the beam replaces the corresponding subgroup, unless it already is part of the beam due to another update action (Lines~\ref{al:csd:beam-search-subgroup-update:line:lb-replace-start}--\ref{al:csd:beam-search-subgroup-update:line:lb-replace-end} and~\ref{al:csd:beam-search-subgroup-update:line:ub-replace-start}--\ref{al:csd:beam-search-subgroup-update:line:ub-replace-end}).

\begin{algorithm}[t]
	\DontPrintSemicolon
	\KwIn{Parameters and variables from Algorithm~\ref{al:csd:generic-beam-search}}
	\KwOut{None; modifies variables from Algorithm~\ref{al:csd:generic-beam-search} in-place}
	$(\mathit{lb},~ \mathit{ub}) \leftarrow (\mathit{lb}^{(\text{beam, } l)},~ \mathit{lb}^{(\text{beam, } l)})$ \tcp*{Value at index~$j$ will change}  \label{al:csd:best-interval-subgroup-update:line:main-start}
	$(\mathit{lb}^{\text{opt}},~ \mathit{ub}^{\text{opt}}) \leftarrow (\mathit{lb}^{(\text{beam, } l)},~ \mathit{lb}^{(\text{beam, } l)})$ \tcp*{$(l, r)$ in~\cite{mampaey2012efficient}}
	$Q^{\text{opt}} \leftarrow Q(\mathit{lb}^{\text{opt}}, \mathit{ub}^{\text{opt}}, X, y)$ \tcp*{$\textit{WRAcc}_{\text{max}}$ in~\cite{mampaey2012efficient}}
	$Q^{\text{temp}} \leftarrow -\infty$ \tcp*{$h_{\text{max}}$ in~\cite{mampaey2012efficient}}
	$\mathit{lb}^{\text{temp}}_j \leftarrow -\infty$ \tcp*{$t_{\text{max}}$ in~\cite{mampaey2012efficient}}
	\For{$b \in $ sort(unique(get\_feature\_values($X$, $j$, $\mathit{lb}^{(\text{beam, } l)}$, $\mathit{ub}^{(\text{beam, } l)}$)))}{
		$\mathit{lb}_j \leftarrow b$\;
		$\mathit{ub}_j \leftarrow \mathit{ub}^{(\text{beam, } l)}_j$\;
		\If{$Q(\mathit{lb}, \mathit{ub}, X, y) > Q^{\text{temp}}$}{
			$\mathit{lb}^{\text{temp}}_j \leftarrow b$\;
			$Q^{\text{temp}} \leftarrow Q(\mathit{lb}, \mathit{ub}, X, y)$\;
		}
		$\mathit{lb}_j \leftarrow \mathit{lb}^{\text{temp}}_j$\;
		$\mathit{ub}_j \leftarrow b$\;
		\If{$Q(\mathit{lb}, \mathit{ub}, X, y) > Q^{\text{opt}}$}{
			$(\mathit{lb}^{\text{opt}},~ \mathit{ub}^{\text{opt}}) \leftarrow (\mathit{lb},~ \mathit{ub})$\;
			$Q^{\text{opt}} \leftarrow Q(\mathit{lb}, \mathit{ub}, X, y)$\; \label{al:csd:best-interval-subgroup-update:line:main-end}
		}
	}
	\If{$\left( Q^{\text{opt}} > \min_{l \in \{1, ..., w\}} Q^{(l)} \right)$ \textbf{and} $(\mathit{lb}^{\text{opt}},~ \mathit{ub}^{\text{opt}}) \notin \{(\mathit{lb}^{(\text{cand, } l)},~ \mathit{ub}^{(\text{cand, } l)}) \mid l \in \{1, \dots, w\}\}$ }{ \label{al:csd:best-interval-subgroup-update:line:replace-start}
		$l \leftarrow \arg\min_{l \in \{1, \dots, w\}} Q^{(l)}$ \tcp*{Replace worst candidate}
		$(\mathit{lb}^{(\text{cand, } l)},~ \mathit{ub}^{(\text{cand, } l)}) \leftarrow (\mathit{lb}^{\text{opt}}, \mathit{ub}^{\text{opt}})$\;
		$\mathit{cand\_has\_changed}^{(l)} \leftarrow$ \textbf{true}\;
		$Q^{(l)} \leftarrow Q^{\text{opt}}$\; \label{al:csd:best-interval-subgroup-update:line:replace-end}
	}
	\caption{evaluate\_subgroup\_updates(\dots) for \emph{Best Interval}.}
	\label{al:csd:best-interval-subgroup-update}
\end{algorithm}

\paragraph{Best Interval}

\emph{Best Interval}~\cite{mampaey2012efficient} offers an update procedure for subgroups (cf.~Algorithm~\ref{al:csd:best-interval-subgroup-update}) that is tailored towards WRAcc (cf.~Equation~\ref{eq:csd:wracc}) as the subgroup-quality function.
This update procedure can be used within a generic beam-search strategy (cf.~Algorithm~\ref{al:csd:generic-beam-search}).
As before, the best new solution candidate from an update step becomes part of the beam if it improves upon the worst subgroup quality there and is not a duplicate (Lines~\ref{al:csd:best-interval-subgroup-update:line:replace-start}--\ref{al:csd:best-interval-subgroup-update:line:replace-end}).

However, solution candidates are generated differently than in the update procedure of \emph{Beam Search} (cf.~Algorithm~\ref{al:csd:beam-search-subgroup-update}).
In particular, \emph{Best Interval} updates lower and upper bounds for a given Feature~$j$ simultaneously rather than separately (Lines~\ref{al:csd:best-interval-subgroup-update:line:main-start}--\ref{al:csd:best-interval-subgroup-update:line:main-end}).
Thus, this procedure optimizes over all potential combinations of lower and upper bounds.
However, it still only requires one pass over the unique values of Feature~$j$ rather than quadratic cost, due to theoretical properties of the WRAcc function~\cite{mampaey2012efficient}.

\subsubsection{Exhaustive Search Methods}
\label{sec:csd:fundamentals:search:exhaustive}

\paragraph{SD-Map}

\emph{SD-Map}~\cite{atzmueller2006sd} is an exhaustive search method based on the pattern-mining algorithm \emph{FP-growth}~\cite{han2000mining}.
It assumes discretized features and produces subgroup descriptions with equality conditions of the form $X_{ij} = v$ instead of the numeric intervals we focus on (Definition~\ref{def:csd:subgroup}).
\emph{SD-Map*}~\cite{atzmueller2009fast} adds support for numeric targets and quality-based pruning of the search space.

\paragraph{Bitset-based Subgroup Discovery (BSD)}

\emph{BSD}~\cite{lemmerich2010fast} is another exhaustive search method that produces equality conditions in the subgroup description.
It combines a branch-and-bound strategy with a special binary data representation and pruning techniques to speed up the search.
\emph{NumBSD}~\cite{lemmerich2016fast} can handle numeric targets, though the features are still assumed to be discrete.

\section{Baselines}
\label{sec:csd:baselines}

In this section, we propose and analyze two baselines for subgroup discovery, \emph{MORS} (cf.~Section~\ref{sec:csd:baselines:mors}) and \emph{Random Search} (cf.~Section~\ref{sec:csd:baselines:random-search}).
They are conceptually simpler than the heuristic search methods (cf.~Section~\ref{sec:csd:fundamentals:search:heuristics}) and serve as further reference points in our experiments.
While they technically also are heuristics, we use the term \emph{baselines} to refer to these two methods specifically.

\begin{algorithm}[t]
	\DontPrintSemicolon
	\KwIn{Dataset~$X \in \mathbb{R}^{m \times n}$, \newline
		Prediction target~$y \in \{0, 1\}^m$
	}
	\KwOut{Subgroup bounds~$\mathit{lb}, \mathit{ub} \in \{\mathbb{R} \cup \{-\infty, +\infty\}\}^n$}
	\BlankLine
	\For{$j \leftarrow 1$ \KwTo $n$}{
		$\mathit{lb}_j \leftarrow \min\limits_{\substack{i \in \{1, \dots, m\} \\ y_i = 1}} X_{ij}$\; \label{al:csd:mors:line:bounds-start}
		$\mathit{ub}_j \leftarrow \max\limits_{\substack{i \in \{1, \dots, m\} \\ y_i = 1}} X_{ij}$\; \label{al:csd:mors:line:bounds-end}
		\lIf{$\mathit{lb}_j = \min_{i \in \{1, \dots, m\}} X_{ij}$}{$\mathit{lb}_j \leftarrow -\infty$} \label{al:csd:mors:line:bounds-infty-start}
		\lIf{$\mathit{ub}_j = \max_{i \in \{1, \dots, m\}} X_{ij}$}{$\mathit{ub}_j \leftarrow + \infty$} \label{al:csd:mors:line:bounds-infty-end}
	}
	\For{$j \notin \text{get\_permissible\_feature\_idxs(\dots)}$}{ \label{al:csd:mors:line:reset-start} \label{al:csd:mors:line:permissible-features}
		$(\mathit{lb}_j,~ \mathit{ub}_j) \leftarrow (-\infty, +\infty)$\; \label{al:csd:mors:line:reset-end}
	}
	\Return{$\mathit{lb}, \mathit{ub}$}
	\caption{\emph{MORS} for subgroup discovery.}
	\label{al:csd:mors}
\end{algorithm}

\subsection{MORS}
\label{sec:csd:baselines:mors}

This baseline builds on the following definition:
\begin{definition}[Minimal Optimal-Recall Subgroup (MORS)]
	Given a dataset $X \in \mathbb{R}^{m \times n}$ with prediction target~$y \in \{0, 1\}^m$,
	the \emph{Minimal Optimal-Recall Subgroup (MORS)} is the subgroup (cf.~Definition~\ref{def:csd:subgroup}) whose lower and upper bounds of each feature correspond to the minimum and maximum value of that feature over all positive data objects (i.e., with $y_i = 1$) from the dataset~$X$.
	\label{def:csd:mors}
\end{definition}
The definition ensures that all positive data objects are contained in the subgroup.
Thus, the evaluation metric \emph{recall}, i.e., the fraction of positive data objects becoming subgroup members, reaches its \emph{optimal} value of~1.
At the same time, raising the lower bounds or lowering the upper bounds would exclude positive data objects from the subgroup.
In this sense, the set of subgroup members is \emph{minimal}.
The corresponding subgroup description is unique and implicitly solves the following variant of the subgroup-discovery problem:
\begin{definition}[Minimal-optimal-recall-subgroup discovery]
	Given a dataset $X \in \mathbb{R}^{m \times n}$ with prediction target~$y \in \{0, 1\}^m$,
	\emph{minimal-optimal-recall-subgroup discovery} is the problem of finding a subgroup (cf.~Definition~\ref{def:csd:subgroup}) that contains as few negative data objects (i.e., with $y_i = 0$) as possible but all positive data objects (i.e., with $y_i = 1$) from the dataset~$X$.
	\label{def:csd:minimal-optimal-recall-subgroup-discovery}
\end{definition}
I.e., the problem targets at minimizing the number of false positives subject to producing no false negatives.
With the constraint on the positive data objects, this problem is equivalent to maximizing the number of true negatives, i.e., negative data objects excluded from the subgroup.

Algorithm~\ref{al:csd:mors} outlines the procedure to determine the \emph{MORS} bounds.
Slightly deviating from Definition~\ref{def:csd:mors}, but consistent to Algorithm~\ref{al:csd:prim}, \emph{MORS} replaces all non-excluding bounds with infinity (Lines~\ref{al:csd:mors:line:bounds-infty-start}--\ref{al:csd:mors:line:bounds-infty-end}).
Further, if only certain features are permissible to be bounded, as we discuss later, we reset the bounds of the remaining features (Lines~\ref{al:csd:mors:line:reset-start}--\ref{al:csd:mors:line:reset-end}).

Since \emph{MORS} only needs to iterate over all data objects and features once to determine the minima and maxima, the time complexity of this algorithm is~$O(m \cdot n)$.
This places minimal-optimal-recall-subgroup discovery in complexity class~$\mathcal{P}$:
\begin{proposition}[Complexity of minimal-optimal-recall-subgroup discovery]
	The problem of minimal-optimal-recall-subgroup discovery (cf.~Definition~\ref{def:csd:minimal-optimal-recall-subgroup-discovery}) can be solved in~$O(m \cdot n)$.
	\label{prop:csd:complexity-unconstrained-minimal-optimal-recall-subgroup-discovery}
\end{proposition}

For complexity proofs later in our article, we define another variant of the subgroup-discovery problem based on another particular type of subgroups~\cite{meeng2014rocsearch}:
\begin{definition}[Perfect subgroup]
	Given a dataset~$X \in \mathbb{R}^{m \times n}$ with prediction target~$y \in \{0, 1\}^m$,
	a \emph{perfect subgroup} is a subgroup (cf.~Definition~\ref{def:csd:subgroup}) that contains all positive data objects (i.e., with $y_i = 1$) but no negative data objects (i.e., with $y_i = 0$) from the dataset~$X$.
	\label{def:csd:perfect-subgroup}
\end{definition}
Perfect subgroups reach the theoretical maximum WRAcc for a dataset (cf.~Equation~\ref{eq:csd:wracc-max}).
Next, we define a corresponding search problem:
\begin{definition}[Perfect-subgroup discovery]
	Given a dataset~$X \in \mathbb{R}^{m \times n}$ with prediction target~$y \in \{0, 1\}^m$,
	\emph{perfect-subgroup discovery} is the problem of finding a perfect subgroup (cf.~Definition~\ref{def:csd:perfect-subgroup}) if it exists or determining that it does not exist.
	\label{def:csd:perfect-subgroup-discovery}
\end{definition}
Since the algorithm \emph{MORS} helps solving this problem in~$O(m \cdot n)$, we obtain the following complexity result:
\begin{proposition}[Complexity of perfect-subgroup discovery]
	The problem of perfect-subgroup discovery (cf.~Definition~\ref{def:csd:perfect-subgroup-discovery}) can be solved in~$O(m \cdot n)$.
	\label{prop:csd:complexity-unconstrained-perfect-subgroup}
\end{proposition}
In particular, after \emph{MORS} (cf.~Algorithm~\ref{al:csd:mors}) has found a subgroup, one only needs to check whether the subgroup contains any negative data objects.
If the found subgroup does not contain negative data objects, then it is perfect.
If it does, then no perfect subgroup exists.
In particular, the bounds found by \emph{MORS} cannot be made tighter to exclude negative data objects from the subgroup without also excluding positive data objects, thereby violating perfection.

\begin{algorithm}[t]
	\DontPrintSemicolon
	\KwIn{Dataset~$X \in \mathbb{R}^{m \times n}$, \newline
		Prediction target~$y \in \{0, 1\}^m$, \newline
		Subgroup-quality function~$Q(\mathit{lb}, \mathit{ub}, X, y)$, \newline
		Number of iterations~$\mathit{n\_iters} \in \mathbb{N}$
	}
	\KwOut{Subgroup bounds~$\mathit{lb}, \mathit{ub} \in \{\mathbb{R} \cup \{-\infty, +\infty\}\}^n$}
	\BlankLine
	$Q^{\text{opt}} \leftarrow - \infty$\;
	\For{$\mathit{iters} \leftarrow 1$ \KwTo $\mathit{n\_iters}$}{
		\For{$j \leftarrow 1$ \KwTo $n$}{ \label{al:csd:random-search:line:sampling-start}
			$(\mathit{lb}_j,~ \mathit{ub}_j) \leftarrow (-\infty, +\infty)$\;
		}
		\For{$j \in$ get\_permissible\_feature\_idxs(\dots)}{ \label{al:csd:random-search:line:permissible-features}
			$(\mathit{lb}_j,~\mathit{ub}_j) \leftarrow$ sample\_uniformly(unique($X_{\cdot j}$)) \label{al:csd:random-search:line:sampling-end}
		}
		\If{$Q(\mathit{lb}, \mathit{ub}, X, y) > Q^{\text{opt}}$}{ \label{al:csd:random-search:line:optimum-update-start}
			$Q^{\text{opt}} \leftarrow Q(\mathit{lb}, \mathit{ub}, X, y)$\;
			$(\mathit{lb}^{\text{opt}},~ \mathit{ub}^{\text{opt}}) \leftarrow (\mathit{lb},~ \mathit{ub})$\; \label{al:csd:random-search:line:optimum-update-end}
		}
	}
	\For{$j \leftarrow 1$ \KwTo $n$}{ \label{al:csd:random-search:line:bounds-infty-start}
		\lIf{$\mathit{lb}_j^{\text{opt}} = \min_{i \in \{1, \dots, m\}} X_{ij}$}{$\mathit{lb}_j^{\text{opt}} \leftarrow -\infty$} 
		\lIf{$\mathit{ub}_j^{\text{opt}} = \max_{i \in \{1, \dots, m\}} X_{ij}$}{$\mathit{ub}_j^{\text{opt}} \leftarrow +\infty$} \label{al:csd:random-search:line:bounds-infty-end}
	}
	\Return{$\mathit{lb}^{\text{opt}},~ \mathit{ub}^{\text{opt}}$}
	\caption{\emph{Random Search} for subgroup discovery.}
	\label{al:csd:random-search}
\end{algorithm}

\subsection{Random Search}
\label{sec:csd:baselines:random-search}

Algorithm~\ref{al:csd:random-search} outlines a randomized search procedure that constitutes the second baseline.
\emph{Random Search} generates and evaluates subgroups for a fixed number of iterations, which the user controls with the parameter~$\mathit{n\_iters} \in \mathbb{N}$.
Hereby, subgroup generation samples a lower bound and an upper bound uniformly random from the unique values for each permissible feature, leaving the remaining features unrestricted (Lines~\ref{al:csd:random-search:line:sampling-start}--\ref{al:csd:random-search:line:sampling-end}).
The algorithm tracks the best generated subgroup so far over the iterations (Lines~\ref{al:csd:random-search:line:optimum-update-start}--\ref{al:csd:random-search:line:optimum-update-end}) and finally returns the subgroup with the highest quality.
Like Algorithm~\ref{al:csd:prim}, \emph{Random Search} replaces all non-excluding bounds with infinity (Lines~\ref{al:csd:random-search:line:bounds-infty-start}--\ref{al:csd:random-search:line:bounds-infty-end}).

\section{Constrained Subgroup Discovery}
\label{sec:csd:approach}

In this section, we discuss subgroup discovery with constraints.
First, we frame subgroup discovery as an SMT optimization problem (cf.~Section~\ref{sec:csd:approach:smt}).
Second, we give a brief overview of potential constraint types (cf.~Section~\ref{sec:csd:approach:constraint-types}).
Third, we formalize and analyze feature-cardinality constraints (cf.~Section~\ref{sec:csd:approach:cardinality}).
Fourth, we formalize and analyze alternative subgroup descriptions (cf.~Section~\ref{sec:csd:approach:alternatives}).

\subsection{SMT Encoding of Subgroup Discovery}
\label{sec:csd:approach:smt}

To find optimal subgroups exactly, one can encode subgroup discovery as a white-box optimization problem and employ a solver.
Here, we propose a Satisfiability Modulo Theories (SMT)~\cite{barrett2018satisfiability, nieuwenhuis2006sat} encoding, which is straightforward given the problem definition (cf.~Definition~\ref{def:csd:subgroup-discovery}).
SMT allows expressions in first-order logic with particular interpretations, e.g., arrays, arithmetic, or bit vectors~\cite{barrett2018satisfiability}.
Our encoding of subgroup discovery uses linear real arithmetic (LRA).
Complementing this SMT encoding, Appendix~\ref{sec:csd:appendix:further-encodings} describes further encodings:
SMT for categorical features (cf.~Section~\ref{sec:csd:appendix:further-encodings:smt-categorical}), mixed integer linear programming (cf.~Section~\ref{sec:csd:appendix:further-encodings:milp}), and maximum satisfiability (cf.~Section~\ref{sec:csd:appendix:further-encodings:max-sat}).

The optimization problem consists of an objective function and constraints.

\paragraph{Objective function}

We use WRAcc as the objective function, which should be maximized.
In the formula for WRAcc (cf.~Equation~\ref{eq:csd:wracc}), $m$ and $m^+$ are constants, while $m_b$ and $m_b^+$ depend on the decision variables.
The previously provided formula seems to be non-linear in the decision variables since $m_b$ appears in the numerator and denominator.
However, one can reformulate the expression by multiplying its two factors, obtaining the following expression:
\begin{equation}
	\text{WRAcc} = \frac{m_b^+}{m} - \frac{m_b \cdot m^+}{m^2} = \frac{m_b^+ \cdot m - m_b \cdot m^+}{m^2}
	\label{eq:csd:smt-wracc}
\end{equation}
In this new expression, the denominators are constant, and the factor~$m^+$ in the numerator is constant as well.
Thus, the whole expression is linear in~$m_b^+$ and~$m_b$.
We define these two quantities as linear expressions from binary decision variables~$b \in \{0, 1\}^m$ that denote subgroup membership.
I.e., $b_i$~expresses whether the $i$-th data object is in the subgroup or not:
\begin{equation}
	\begin{aligned}
		 m_b &:= \sum_{i=1}^{m} b_i \\
		 m_b^+ &:= \sum_{\substack{i \in \{1, \dots, m\} \\ y_i = 1 }} b_i \\
	\end{aligned}
	\label{eq:csd:smt-constraint-m-as-sum}
\end{equation}
Since the values of the target variable~$y$ are fixed, the expression for~$m_b^+$ only sums over the positive data objects.
Further, one may define $m_b^+$ and~$m_b$ as separate integer variables or directly insert their expressions into Equation~\ref{eq:csd:smt-wracc}.
We chose the latter formulation in our implementation and therefore wrote $:=$ in Equation~\ref{eq:csd:smt-constraint-m-as-sum} instead of using a proper propositional operator like $\leftrightarrow$.

The formula for nWRAcc (cf.~Equation~\ref{eq:csd:wracc-normalized}) is linear as well, having the same enumerator as Equation~\ref{eq:csd:smt-wracc} and a different constant in the denominator.

\paragraph{Constraints}

The subgroup membership~$b_i$ of a data object depends on the bounds of the subgroup (cf.~Definition~\ref{def:csd:subgroup}).
Thus, we define real-valued decision variables $\mathit{lb}, \mathit{ub} \in \{\mathbb{R} \cup \{-\infty, +\infty\}\}^n$ for the latter.
In particular, there is one lower bound and one upper bound for each of the $n$~features.
The upper bounds naturally need to be at least as high as the lower bounds:
\begin{equation}
	\forall j \in \{1, \dots, n\}:~ \mathit{lb}_j\leq \mathit{ub}_j
	\label{eq:csd:smt-constraint-bounds-monotonic}
\end{equation}
A data object is a member of the subgroup if all its feature values are contained within the bounds:
\begin{equation}
	\forall i \in \{1, \dots, m\}:~ b_i\leftrightarrow \bigwedge_{j \in \{1, \dots, n\}} \left( \left( X_{ij} \geq \mathit{lb}_j \right) \land \left( X_{ij} \leq \mathit{ub}_j \right) \right)
	\label{eq:csd:smt-constraint-subgroup-membership}
\end{equation}
Instead of defining separate decision variables~$b_i$ and binding them to the bounds with an equivalence constraint, one could also insert the Boolean expression into the right-hand-side of Equation~\ref{eq:csd:smt-constraint-m-as-sum} directly.
In particular, $\mathit{lb}_j$ and $\mathit{ub}_j$ are the only decision variables strictly necessary for the optimization problem.
However, for formulating some constraint types on subgroups (cf.~Section~\ref{sec:csd:approach:constraint-types}), it is helpful to be able to refer to~$b_i$.

\paragraph{Complete optimization problem}

Combining all prior definitions of decision variables, constraints, and the objective function, we obtain the following SMT optimization problem:

\begin{equation}
	\begin{aligned}
		\max &\quad & Q_{\text{WRAcc}} &= \frac{m_b^+}{m} - \frac{m_b \cdot m^+}{m^2} \\
		\text{s.t.:} &\quad & m_b &:= \sum_{i=1}^{m} b_i \\
		&\quad & m_b^+ &:= \sum_{\substack{i \in \{1, \dots, m\} \\ y_i = 1 }} b_i \\
		&\quad \forall i \in \{1, \dots, m\} & b_i &\leftrightarrow \bigwedge_{j \in \{1, \dots, n\}} \left( \left( X_{ij} \geq \mathit{lb}_j \right) \land \left( X_{ij} \leq \mathit{ub}_j \right) \right) \\
		&\quad \forall j \in \{1, \dots, n\} & \mathit{lb}_j &\leq \mathit{ub}_j \\
		&\quad & b &\in \{0, 1\}^m \\
		&\quad & \mathit{lb}, \mathit{ub} &\in \{\mathbb{R} \cup \{-\infty, +\infty\}\}^n
	\end{aligned}
	\label{eq:csd:smt-problem-unconstrained-complete}
\end{equation}

We call this optimization problem \emph{unconstrained subgroup discovery} in the following since it only contains constraints that are technically necessary to define subgroup discovery properly but no additional constraints (cf.~Section~\ref{sec:csd:approach:constraint-types}).

\paragraph{Post-processing}

In our implementation, we add a small post-processing step.
In particular, we do not use the solver-determined values of the variables $\mathit{lb}_j$ and $\mathit{ub}_j$ when evaluating subgroup quality.
Instead, we set the lower and upper bounds to the minimum and maximum feature values of all data objects in the subgroup (i.e., with $b_i=1$).
Thus, we ensure that the bounds correspond to actual feature values.
This guarantee is not formally necessary but consistent with the subgroup descriptions returned by heuristic search methods and baselines.
Also, we avoid potential minor numerical issues caused by extracting the values of real variables from the solver.
Finally, if the subgroup does not contain any data objects, we use invalid bounds (i.e., $\mathit{ub}_j = -\infty < \infty = \mathit{lb}_j$) to ensure that the subgroup remains empty even for arbitrary new data objects.

\subsection{Overview of Constraint Types}
\label{sec:csd:approach:constraint-types}

A white-box formulation of subgroup discovery, like our SMT encoding, supports directly integrating a variety of constraint types in a declarative manner.
In contrast, algorithmic search methods for subgroups need to explicitly check constraints or implicitly ensure that generated solution candidates satisfy constraints.
Such implicit guarantees may only hold for particular constraint types or may require adapting the search method accordingly.

\paragraph{Domain knowledge}

Constraints can express firm knowledge or hypotheses from the domain that solutions of machine-learning techniques should adhere to~\cite{bach2022empirical}.
Since the definition of such constraints depends on the use case, we do not give domain-specific examples here.
\cite{atzmueller2006methodological, atzmueller2005exploiting, atzmueller2007using} provide a taxonomy and examples for knowledge-based constraint types in subgroup discovery.
In a white-box formulation of subgroup discovery, such constraints may restrict the values of decision variables, e.g., lower and upper bounds, subgroup membership (cf.~Equation~\ref{eq:csd:smt-problem-unconstrained-complete}), or selected features (cf.~Equation~\ref{eq:csd:smt-constraint-feature-selection}).
For example, certain bound values or feature combinations used in the subgroup description may contradict domain knowledge and should therefore be prevented with constraints.
In our formulation of subgroup discovery as an SMT problem with linear real arithmetic (cf.~Section~\ref{sec:csd:approach:smt}), one can employ propositional logic, basic arithmetic operators, and inequalities to express constraints over the decision variables~\cite{barrett2018satisfiability}.

\paragraph{Secondary objectives}

Various notions of subgroup quality can serve as an objective in subgroup discovery~\cite{atzmueller2015subgroup, herrera2011overview}.
If one wants to consider several quality metrics simultaneously, one option is multi-objective optimization.
However, the latter typically requires using different search methods than single-objective optimization.
Also, there may be not one but a set of Pareto-optimal solutions, or users may need to define trade-offs between objectives manually.
Alternatively, one can keep a single primary objective and add the other objectives as inequality constraints, e.g., enforcing that their values are below or above user-defined thresholds.
According to~\cite{meeng2021real}, such lower bounds on subgroup quality are a common constraint type in subgroup discovery.
Without constraints, one may prune the set of discovered subgroups as a post-processing step~\cite{atzmueller2015subgroup}.
Finally, quality-based pruning can also reduce the search space during subgroup discovery, e.g., using optimistic estimates in exhaustive search~\cite{atzmueller2015subgroup, atzmueller2009fast, grosskreutz2009subgroup}.
However, such automatically determined bounds on subgroup quality relate to the primary optimization objective rather than being user-provided constraints.

\paragraph{Regularization}

Regularization aims to control the complexity of machine-learning models, preventing overfitting and increasing interpretability.
In subgroup discovery, we see three directions for regularization:
the data objects via subgroup membership, the feature values via the subgroup's bounds, and the features via the subgroup's feature selection.

Regarding \emph{subgroup membership}, one can introduce lower or upper bounds on the number of data objects in the subgroup, using the decision variables~$b_i$ (cf.~Equation~\ref{eq:csd:smt-constraint-subgroup-membership}).
Such constraints are particularly useful for notions of subgroup quality that incentivize including all data objects in the subgroup, like recall, or including very few data objects, like precision.
In contrast, WRAcc (cf.~Equation~\ref{eq:csd:wracc}) automatically considers the number of data objects in the subgroup.
According to~\cite{meeng2021real}, lower bounds on subgroup membership are a common constraint type in subgroup discovery.

Regarding \emph{bounds}, one can define minimum or maximum values for the range of a feature in the subgroup, i.e., the difference between lower and upper bound, using the decision variables~$\mathit{lb}_j$ and~$\mathit{ub}_j$.
Such constraints can prevent choosing ranges that are too small or too large for user needs.
If the features are normalized, one can also constrain the volume of the subgroup, i.e., the product of all ranges, or the density, i.e., the number of data objects per volume.
Generally, however, a feature's bounded value range need not indicate how many data objects are excluded from the subgroup.
Alternatively, one can constrain the subgroup membership implied by individual features' bounds, e.g., enforcing that selected features exclude at least a certain fraction of data objects.
The latter constraint type may prevent setting oversensitive bounds that only exclude few data objects and do not generalize to unseen data.

Regarding \emph{feature selection}, one can limit the number of features used in the subgroup description, which is a common constraint type~\cite{meeng2021real} and also a metric for subgroup complexity~\cite{helal2016subgroup, herrera2011overview, ventura2018subgroup}, already proposed in the article introducing PRIM~\cite{friedman1999bump}.
Section~\ref{sec:csd:approach:cardinality} discusses such feature-cardinality constraints in detail.
Instead of using this constraint type, one may also post-process subgroups to eliminate irrelevant features after search~\cite{friedman1999bump}.
Further, instead of limiting the total number of used features, one may also introduce constraints to remove individual irrelevant features based on the number of data objects they represent correctly in absolute terms or relative to other features~\cite{lavrac2006relevancy}.

\paragraph{Alternatives}

As for regularization, constraints for alternatives may relate to data objects, features, or feature values.
We see two major notions of alternatives, which we call \emph{alternative subgroups} and \emph{alternative subgroup descriptions}.

\emph{Alternative subgroups} aim to contain different sets of data objects.
Since subgroups only intend to cover specific regions of the data, it is natural to search for multiple subgroups to cover multiple regions; see~\cite{atzmueller2015subgroup} for an overview of subgroup-set selection.
One can search for multiple subgroups sequentially or simultaneously.
The `covering' approach allows sequential search for any subgroup-discovery method by removing all data objects contained in previous subgroups and repeating subgroup discovery~\cite{friedman1999bump}.
Alternatively, one may reweigh~\cite{gamberger2002expert, lavrac2004subgroup} or resample~\cite{scholz2005sampling} data objects based on their subgroup membership.
In contrast, simultaneous search requires subgroup-discovery methods that specifically target at multiple solutions~\cite{leeuwen2012diverse, leeuwen2013discovering, lemmerich2010fast, lucas2018ssdp+, proencca2022robust}.
E.g., a heuristic search may retain multiple solution candidates at each step.
The notion of subgroup quality typically becomes broader:
Besides measures for predictive quality like WRAcc, the diversity of subgroups~\cite{belfodil2019fssd, leeuwen2012diverse, leeuwen2013discovering, lucas2018ssdp+}, e.g., to which extent they contain different data objects, and their number~\cite{helal2016subgroup, herrera2011overview, ventura2018subgroup} may also serve as metrics.
One may also filter redundant subgroups as a post-processing step~\cite{bosc2018anytime, grosskreutz2012enhanced, hudson2023subgroup, leeuwen2013discovering}.
In a white-box formulation of subgroup discovery, one can enforce diversity with appropriate constraints on subgroup membership (cf.~Equation~\ref{eq:csd:smt-constraint-subgroup-membership}), e.g., limiting the number of data objects that can be members of two subgroups simultaneously.

In contrast, \emph{alternative subgroup descriptions} explicitly aim to contain a similar set of data objects but use different subgroup descriptions, e.g., a different feature selection.
Section~\ref{sec:csd:approach:alternatives} discusses this constraint type in detail.
Related to this concept, \cite{boley2009non}~introduces the notion of equivalent subgroup descriptions of minimal length, which cover exactly the same set of data objects.

\subsection{Feature-Cardinality Constraints}
\label{sec:csd:approach:cardinality}

In this section, we discuss feature-cardinality constraints for subgroup discovery.
First, we motivate and formalize them (cf.~Section~\ref{sec:csd:approach:cardinality:concept}).
Next, we describe how to integrate them into our SMT encoding of subgroup discovery (cf.~Section~\ref{sec:csd:approach:cardinality:smt}), heuristic search methods (cf.~Section~\ref{sec:csd:approach:cardinality:heuristics}), and baselines (cf.~Section~\ref{sec:csd:approach:cardinality:baselines}).
Finally, we analyze the time complexity of subgroup discovery with this constraint type (cf.~Section~\ref{sec:csd:approach:cardinality:complexity}).

\subsubsection{Concept}
\label{sec:csd:approach:cardinality:concept}

Feature-cardinality constraints are a constraint type that regularizes subgroup descriptions (cf.~Section~\ref{sec:csd:approach:constraint-types}).
In particular, this constraint type limits the number of features used in the subgroup description, rendering the latter less complex and easier to interpret~\cite{meeng2021real}.
To formalize this constraint type, we define feature selection~\cite{guyon2003introduction, li2017feature} in the context of subgroup discovery as follows:
\begin{definition}[Feature selection in subgroups]
	Given a dataset~$X \in \mathbb{R}^{m \times n}$ and a subgroup (cf.~Definition~\ref{def:csd:subgroup}) with bounds~$\mathit{lb}, \mathit{ub} \in \{\mathbb{R} \cup \{-\infty, +\infty\}\}^n$,
	Feature~$j$ is \emph{selected} if the bounds exclude at least one data object of~$X$ from the subgroup, i.e., $\exists i \in \{1, \dots, m\}:~ \left( X_{ij} < \mathit{lb}_j \right) \lor \left( X_{ij} > \mathit{ub}_j \right)$.
	\label{def:csd:feature-selection}
\end{definition}
The bounds of unselected features can be considered infinite, effectively removing these features from the subgroup description.
The \emph{feature cardinality} of the subgroup is the number of selected features.
Related work also uses the terms \emph{depth} \cite{meeng2021real} or \emph{length} \cite{atzmueller2015subgroup, helal2016subgroup}, though partly referring to the number of conditions in the subgroup description rather than selected features.
I.e., if there is a lower and an upper bound for a feature, some related work counts this feature twice instead of once.

To formulate a feature-cardinality constraint, users provide an upper bound on the number of selected features:
\begin{definition}[Feature-cardinality constraint]
	Given a cardinality threshold $k \in \mathbb{N}$,
	a \emph{feature-cardinality constraint} for a subgroup (cf.~Definition~\ref{def:csd:subgroup}) requires the subgroup to have at most $k$~features selected (cf.~Definition~\ref{def:csd:feature-selection}).
	\label{def:csd:feature-cardinality-constraint}
\end{definition}
In practice, less than $k$ features may be selected if selecting more features does not improve the subgroup quality.

\subsubsection{SMT Encoding}
\label{sec:csd:approach:cardinality:smt}

We first need to encode whether a feature is selected or not.
Thus, we introduce binary decision variables $s, s^{\text{lb}}, s^{\text{ub}} \in \{0, 1\}^n$.
A feature is selected if its bounds exclude at least one data object from the subgroup (cf.~Definition~\ref{def:csd:feature-selection}), i.e., the lower bound is higher than the minimum feature value or the upper bound is lower than the maximum feature value:
\begin{equation}
	\begin{aligned}
		\forall j: & & s^{\text{lb}}_j &\leftrightarrow \left( \mathit{lb}_j > \min_{i \in \{1, \dots, m\}} X_{ij} \right) \\
		\forall j: & &s^{\text{ub}}_j &\leftrightarrow \left( \mathit{ub}_j < \max_{i \in \{1, \dots, m\}} X_{ij} \right) \\
		\forall j: & & s_j &\leftrightarrow \left( s^{\text{lb}}_j \lor s^{\text{ub}}_j \right) \\
		\text{with index:} & & j &\in \{1, \dots, n\}
	\end{aligned}
	\label{eq:csd:smt-constraint-feature-selection}
\end{equation}
In this equation, minimum and maximum feature values are constants that can be determined before formulating the optimization problem.

Given the definition of~$s_j$, setting an upper bound on the number of selected features (cf.~Definition~\ref{def:csd:feature-cardinality-constraint}) is straightforward:
\begin{equation}
	\sum_{j=1}^n s_j \leq k
	\label{eq:csd:smt-constraint-feature-cardinalty}
\end{equation}
Instead of explicitly defining the decision variables $s_j$, $s^{\text{lb}}_j$, and $s^{\text{ub}}_j$, one could also insert the corresponding expressions into Equation~\ref{eq:csd:smt-constraint-feature-cardinalty} directly.
However, we will also use~$s_j$ for alternative subgroup descriptions (cf.~Section~\ref{sec:csd:approach:alternatives:smt}), so we define corresponding variables in our implementation.

The overall SMT encoding of subgroup discovery with a feature-cardinality constraint is the SMT encoding of unconstrained subgroup discovery (cf.~Equation~\ref{eq:csd:smt-problem-unconstrained-complete}) supplemented by the variables and constraints from Equations~\ref{eq:csd:smt-constraint-feature-selection} and~\ref{eq:csd:smt-constraint-feature-cardinalty}.

In our implementation, we also add a post-processing step that sets non-excluding lower bounds (i.e., with $s^{\text{lb}}_j = 0$) to $-\infty$ and non-excluding upper bounds (i.e., with $s^{\text{ub}}_j = 0$) to $+\infty$.
This step is consistent with the heuristic search methods and baselines (e.g., cf.~Lines~\ref{al:csd:prim:line:bounds-infty-start}--\ref{al:csd:prim:line:bounds-infty-end} in Algorithm~\ref{al:csd:prim}).

\subsubsection{Integration into Heuristic Search Methods}
\label{sec:csd:approach:cardinality:heuristics}

The feature-cardinality constraint (cf.~Equation~\ref{eq:csd:smt-constraint-feature-cardinalty}) has the form~$|F_s| \leq k$ for the feature set~$F_s$ induced by the selection decisions~$s \in \{0, 1\}^n$.
Thus, the constraint is \emph{antimonotonic}~\cite{ng1998exploratory} regarding the set of selected features.
In particular, the empty feature set satisfies the constraint for any~$k \geq 0$.
If a selected feature set satisfies the constraint, all its subsets also satisfy it.
Vice versa, if a feature set violates the constraint, all its supersets violate it as well.

This property allows to easily integrate the constraint into the three heuristic search methods from Section~\ref{sec:csd:fundamentals:search:heuristics}, i.e., \emph{PRIM} (cf.~Algorithm~\ref{al:csd:prim}), \emph{Beam Search} (cf.~Algorithms~\ref{al:csd:generic-beam-search} and~\ref{al:csd:beam-search-subgroup-update}), and \emph{Best Interval} (cf.~Algorithms~\ref{al:csd:generic-beam-search} and~\ref{al:csd:best-interval-subgroup-update}), which all iteratively enlarge the selected feature set.
In particular, they start with unrestricted subgroup bounds, i.e., an empty feature set.
Each iteration may either add bounds on one further feature, thereby increasing the feature-set size by one, or refine the bounds on an already selected feature, so the feature-set size remains constant.
Features cannot be deselected, so the feature-set size is non-decreasing overall.
Thus, one can prevent generation of invalid solution candidates by defining the function \emph{get\_permissible\_feature\_idxs(\dots)} (cf.~Line~\ref{al:csd:prim:line:permissible-features} in Algorithm~\ref{al:csd:prim} and Line~\ref{al:csd:generic-beam-search:line:permissible-features} in Algorithm~\ref{al:csd:generic-beam-search}) as follows:
If already $k$~features are selected, only these features are permissible, i.e., only their bounds may be refined.
If less than $k$~features are selected, all features are permissible to be bounded, as in the unconstrained search.

\subsubsection{Integration into Baselines}
\label{sec:csd:approach:cardinality:baselines}

\paragraph{MORS}

\emph{MORS} calls the function \emph{get\_permissible\_feature\_idxs(\dots)} in Line~\ref{al:csd:mors:line:permissible-features} of Algorithm~\ref{al:csd:mors}.
To instantiate this function, we employ a univariate, quality-based heuristic for feature selection:
For each feature, we evaluate what would happen if only this feature was restricted according to \emph{MORS} (cf.~Definition~\ref{def:csd:mors}).
In particular, we determine the number of false positives, i.e., negative data objects in the subgroup, defined by each feature's \emph{MORS} bounds (Lines~\ref{al:csd:mors:line:bounds-start}--\ref{al:csd:mors:line:bounds-end}).
We select the $k$~features with the lowest number of false positives.

This heuristic is equivalent to selecting the features with the highest WRAcc for univariate \emph{MORS} bounds:
Due to \emph{MORS}, not only~$m$ and~$m^+$ are constant in Equation~\ref{eq:csd:wracc} but also the number of positive data objects in the subgroup~$m_b^+$, which equals~$m^+$.
In particular, one can rephrase Equation~\ref{eq:csd:wracc} as follows:
\begin{equation}
	\text{WRAcc}_{\text{MORS}} = \frac{m_b^+}{m} - \frac{m_b \cdot m^+}{m^2} = \frac{m^+}{m} - \frac{m_b \cdot m^+}{m^2} = \frac{m^+}{m} \cdot \left( 1 - \frac{m_b}{m} \right)
	\label{eq:csd:wracc-mors}
\end{equation}
Thus, maximizing WRAcc corresponds to minimizing~$m_b / m$, i.e., the relative frequency of the data objects in the subgroup.
Since the number of positive data objects in the subgroup is fixed, this objective amounts to including as few negative data objects as possible in the subgroup, i.e., minimizing the number of false positives, which is what our univariate heuristic does as well.

The proposed heuristic only entails a linear runtime in the number of features, like the unconstrained \emph{MORS}, since it evaluates each feature independently.
With quadratic runtime, one can also consider interactions between features and thereby potentially increase subgroup quality.
In particular, one could select features sequentially instead of simultaneously.
In each iteration, one would select the feature whose \emph{MORS} bounds, combined with the \emph{MORS} bounds of all features selected in previous iterations, yield the lowest number of false positives.
This sequential procedure mimics an existing greedy heuristic for the \textsc{Maximum Coverage} problem~\cite{chekuri2004maximum} (cf.~Sections~\ref{sec:csd:appendix:proofs:complexity-cardinality-np-perfect-subgroup} and~\ref{sec:csd:appendix:proofs:complexity-cardinality-np}).

\paragraph{Random Search}

\emph{Random Search} calls \emph{get\_permissible\_feature\_idxs(\dots)} in Line~\ref{al:csd:random-search:line:permissible-features} of Algorithm~\ref{al:csd:random-search}).
To observe a feature-cardinality threshold~$k$, we simply sample~$k$ out of $n$ features uniformly random without replacement.
The bounds for these features will be restricted in the next step of the algorithm, while all remaining features remain unrestricted.

\subsubsection{Time Complexity}
\label{sec:csd:approach:cardinality:complexity}

We analyze three aspects of time complexity:
the size of the search space for exhaustive search, parameterized complexity, and $\mathcal{NP}$-hardness.

\paragraph{Exhaustive search}

Before addressing feature-cardinality constraints, we analyze the unconstrained case.
In general, the search space of subgroup discovery depends on the number of relevant candidate values for lower and upper bounds.
With $m$~data objects, each real-valued feature may have up to $m$~unique values.
It suffices to treat these unique values as bound candidates since any bounds between feature values or outside the feature's range do not change the subgroup membership during optimization, though the prediction on a test set with further data objects may vary.
Thus, there are $O(m^2)$ relevant lower-upper-bound combinations per feature.
Since we need to combine bounds over all $n$~features, the size of the search space is $O(m^{2n})$:
\begin{proposition}[Complexity of exhaustive search for subgroup discovery]
	An exhaustive search for subgroup discovery (cf.~Definition~\ref{def:csd:subgroup-discovery}) needs to evaluate $O(m^{2n})$ subgroups.
	\label{prop:csd:complexity-unconstrained-exhaustive}
\end{proposition}
For each of these candidate subgroups, the cost of evaluating a quality function like WRAcc (cf.~Equation~\ref{eq:csd:wracc}) typically is $O(m \cdot n)$ since it requires a constant number of passes over the dataset.
Additionally, the number of potential subgroups should be seen as an upper bound:
More efficient exhaustive search methods employ quality-based pruning to not explicitly evaluate all solution candidates while still implicitly covering the full search space~\cite{atzmueller2015subgroup}.

Next, we adapt the result from Proposition~\ref{prop:csd:complexity-unconstrained-exhaustive} to feature-cardinality constraints.
Instead of combining bounds from all $n$~features, there are $\binom{n}{k} \leq n^k$ feature sets of size~$k$ with $O(m^{2k})$ bound candidates each:
\begin{proposition}[Complexity of exhaustive search for subgroup discovery with feature-cardinality constraint]
	An exhaustive search for subgroup discovery (cf. Definition~\ref{def:csd:subgroup-discovery}) with a feature-cardinality constraint (cf.~Definition~\ref{def:csd:feature-cardinality-constraint}) needs to evaluate $O(n^k \cdot m^{2k})$ subgroups.
	\label{prop:csd:complexity-cardinality-exhaustive}
\end{proposition}
For the special case $k=1$, the size of the search space becomes $O(m^2 \cdot n)$, which is leveraged by heuristic search methods that only consider updating the bounds of each feature separately instead of jointly (cf.~Section~\ref{sec:csd:fundamentals:search:heuristics}).
With the update procedure of \emph{Best Interval} (cf.~Algorithm~\ref{al:csd:best-interval-subgroup-update}), the cost for~$k=1$ even reduces to $O(m \cdot n)$ since it only requires one pass over the unique values of each feature to evaluate all lower-upper-bound combinations for WRAcc implicitly.
The update procedure of \emph{Beam Search} (cf.~Algorithm~\ref{al:csd:beam-search-subgroup-update}) also requires $O(m \cdot n)$, by only checking updates of either lower or upper bound.

\paragraph{Parameterized complexity}

For unconstrained subgroup discovery, the complexity term from Proposition~\ref{prop:csd:complexity-unconstrained-exhaustive} is polynomial in~$m$ if we consider~$n$ to be a small constant.
In particular, the term takes the form $O(f(n) \cdot m^{g(n)})$ with parameter~$n$ and polynomial functions~$f(\cdot)$ and~$g(\cdot)$~\cite{downey1997parameterized}.
Thus, the problem of subgroup discovery belongs to the parameterized complexity class $\mathcal{XP}$:
\begin{proposition}[Parameterized complexity of subgroup discovery]
	The problem of subgroup discovery (cf.~Definition~\ref{def:csd:subgroup-discovery}) resides in the parameterized complexity class~$\mathcal{XP}$ for the parameter~$n$.
	\label{prop:csd:complexity-unconstrained-xp}
\end{proposition}
Due to the exponent~$2n$ in Proposition~\ref{prop:csd:complexity-unconstrained-exhaustive}, an exhaustive search may be infeasible in practice, even for a small, constant~$n$.
Further, the complexity remains exponential in~$n$ if $m$~is fixed.
I.e., the number of features has an exponential impact on the size of the search space, while the number of data objects has a polynomial impact.

With a feature-cardinality constraint, the problem retains its $\mathcal{XP}$ membership.
Considering the size of the search space from Proposition~\ref{prop:csd:complexity-cardinality-exhaustive}, the parameter is~$k$ instead of~$n$ now:
\begin{proposition}[Parameterized complexity of subgroup discovery with feature-cardinality constraint]
	The problem of subgroup discovery (cf.~Definition~\ref{def:csd:subgroup-discovery}) with a feature-cardinality constraint (cf.~Definition~\ref{def:csd:feature-cardinality-constraint}) resides in the parameterized complexity class~$\mathcal{XP}$ for the parameter~$k$.
	\label{prop:csd:complexity-cardinality-xp}
\end{proposition}

\paragraph{NP-Hardness}

\cite{boley2009non} showed that it is an $\mathcal{NP}$-hard problem to find a subgroup description with minimum feature cardinality that induces exactly the same subgroup membership as a given subgroup.
We transfer this result to optimizing subgroup quality under a feature-cardinality constraint.
First, we tackle the search problem for perfect subgroups (cf.~Appendix~\ref{sec:csd:appendix:proofs:complexity-cardinality-np-perfect-subgroup} for the proof):
\begin{proposition}[Complexity of perfect-subgroup discovery with feature-cardinality constraint]
	 The problem of perfect-subgroup discovery (cf.~Definition~\ref{def:csd:perfect-subgroup-discovery}) with a feature-cardinality constraint (cf.~Definition~\ref{def:csd:feature-cardinality-constraint}) is $\mathcal{NP}$-complete.
	\label{prop:csd:complexity-cardinality-np-perfect-subgroup}
\end{proposition}
This hardness result under a feature-cardinality constraint contrasts with the polynomial runtime of unconstrained perfect-subgroup discovery (cf.~Proposition~\ref{prop:csd:complexity-unconstrained-perfect-subgroup}), which corresponds to a cardinality constraint with $k = n$.

Generalizing Proposition~\ref{prop:csd:complexity-cardinality-np-perfect-subgroup}, the optimization problem of subgroup discovery with a feature-cardinality constraint is $\mathcal{NP}$-complete under a reasonable assumption on the notion of subgroup quality (cf.~Appendix~\ref{sec:csd:appendix:proofs:complexity-cardinality-np} for the proof):
\begin{proposition}[Complexity of subgroup discovery with feature-cardinality constraint]
	Assuming a subgroup-quality function~$Q(\mathit{lb}, \mathit{ub}, X, y)$ for which only perfect subgroups (cf.~Definition~\ref{def:csd:perfect-subgroup}) reach its maximal value,
	the problem of subgroup discovery (cf.~Definition~\ref{def:csd:subgroup-discovery}) with a feature-cardinality constraint (cf.~Definition~\ref{def:csd:feature-cardinality-constraint}) is $\mathcal{NP}$-complete.
	\label{prop:csd:complexity-cardinality-np}
\end{proposition}
WRAcc as the subgroup-quality function satisfies this assumption since only perfect subgroups yield the theoretical maximum WRAcc (cf.~Equation~\ref{eq:csd:wracc-max}).

\subsection{Alternative Subgroup Descriptions}
\label{sec:csd:approach:alternatives}

In this section, we propose the optimization problem of discovering alternative subgroup descriptions.
First, we motivate and formalize the problem (cf.~Section~\ref{sec:csd:approach:alternatives:concept}).
Next, we describe how to phrase it within our SMT encoding of subgroup discovery (cf.~Section~\ref{sec:csd:approach:alternatives:smt}), heuristic search methods (cf.~Section~\ref{sec:csd:approach:alternatives:heuristics}), and baselines (cf.~Section~\ref{sec:csd:approach:alternatives:baselines}).
Finally, we analyze the time complexity of this problem (cf.~Section~\ref{sec:csd:approach:alternatives:complexity}).

\subsubsection{Concept}
\label{sec:csd:approach:alternatives:concept}

\paragraph{Overview}

For alternative subgroup descriptions, we assume to have an \emph{original subgroup} given, with subgroup membership~$b^{(0)} \in \{0, 1\}^m$ of data objects and with feature selection~$s^{(0)} \in \{0, 1\}^n$.
This subgroup may originate from any subgroup-discovery method.
When searching alternatives, we do not optimize subgroup quality but the similarity to the original subgroup.
We express this similarity in terms of subgroup membership.
If this similarity is very high, then the subgroup quality should also be similar since evaluation metrics for subgroup quality typically base on subgroup membership.

Additionally, we constrain the new subgroup description to be alternative enough.
We express this dissimilarity in terms of the subgroups' feature selection.
The user chooses a dissimilarity threshold~$\tau \in \mathbb{R}_{\geq 0}$ and can thereby control alternatives.
Further, we recommend employing a feature-cardinality constraint (cf.~Definition~\ref{def:csd:feature-cardinality-constraint}) when determining the original subgroup, so there are sufficiently many features left for the alternative description.
The alternative may be feature-cardinality-constrained as well, to increase its interpretability.

In a nutshell, alternative subgroup descriptions should produce similar predictions as the original subgroup but use different features.

\paragraph{Sequential search}

One can search for multiple alternative subgroup descriptions sequentially.
After determining the original subgroup, each iteration yields one further alternative.
The user may prescribe a number of alternatives~$a \in \mathbb{N}$ a priori or interrupt the procedure whenever the alternatives are not interesting anymore, e.g., too dissimilar to the original subgroup.
Each alternative should have a similar subgroup membership as the original subgroup but a dissimilar feature selection compared to all \emph{existing subgroups}, i.e., subgroups found in prior iterations.
The following definition captures this optimization problem:
\begin{definition}[Alternative-subgroup-description discovery]
	Given
	\begin{itemize}[noitemsep]
		\item a dataset $X \in \mathbb{R}^{m \times n}$,
		\item $a \in \mathbb{N}$ existing subgroups with subgroup membership~$b^{(l)} \in \{0, 1\}^m$ and feature selection~$s^{(l)} \in \{0, 1\}^n$ for $l \in \{0, \dots, a - 1\}$,
		\item a similarity measure $\text{sim}(\cdot)$ for subgroup-membership vectors,
		\item a dissimilarity measure $\text{dis}(\cdot)$ for feature-selection vectors of subgroups,
		\item and a dissimilarity threshold~$\tau \in \mathbb{R}_{\geq 0}$,
	\end{itemize}
	\emph{alternative-subgroup-description discovery} is the problem of finding a subgroup (cf.~Definition~\ref{def:csd:subgroup}) with membership~$b^{(a)} \in \{0, 1\}^m$ and feature selection~$s^{(a)} \in \{0, 1\}^n$ that maximizes the subgroup-membership similarity $\text{sim}(b^{(a)}, b^{(0)})$ to the original subgroup while being dissimilar to all existing subgroups regarding the feature selection, i.e., $\forall l \in \{0, \dots, a-1\}:~\text{dis}(s^{(a)}, s^{(l)}) \geq \tau$.
	\label{def:csd:alternative-subgroup-description-discovery}
\end{definition}
Compared to the conventional subgroup-discovery problem (cf.~Definition~\ref{def:csd:subgroup-discovery}), $b^{(0)}$ replaces the prediction target~$y$, and $\text{sim}(\cdot)$ replaces the subgroup-quality function~$Q(\mathit{lb}, \mathit{ub}, X, y)$.
In the following, we discuss $\text{sim}(\cdot)$ and $\text{dis}(\cdot)$.

\paragraph{Similarity in objective function}

There are various options to quantify the similarity of subgroup membership, i.e., between two binary vectors.
For example, the Hamming distance counts how many vector entries differ~\cite{choi2010survey}.
We turn this distance into a similarity measure by counting identical vector entries.
Further, we normalize the similarity with the vector length, i.e., number of data objects~$m$, to obtain the following \emph{normalized Hamming similarity} for two subgroup-membership vectors~$b', b'' \in \{0, 1\}^m$:
\begin{equation}
	\text{sim}_{\text{nHamm}}(b', b'') = \frac{1}{m} \cdot \sum_{i=1}^{m} (b'_i = b''_i)
	\label{eq:csd:hamming-general}
\end{equation}
If either $b'$ or $b''$ is constant, then this similarity measure is linear in its remaining argument, as discussed later (cf.~Equation~\ref{eq:csd:smt-hamming}).
Further, if one considers one vector to be a prediction and the other to be the ground truth, Equation~\ref{eq:csd:hamming-general} equals prediction accuracy for classification.

Another popular similarity measure for sets or binary vectors is the Jaccard index~\cite{choi2010survey}, which relates the overlap of positive vector entries to their union:
\begin{equation}
	\text{sim}_{\text{Jacc}}(b', b'') = \frac{\sum_{i=1}^{m} (b'_i \land b''_i)}{\sum_{i=1}^{m} (b'_i \lor b''_i)}
	\label{eq:csd:jaccard}
\end{equation}
However, this similarity measure is not linear in~$b'$ and~$b''$, which prevents its use in certain white-box solvers.
Thus, we use the normalized Hamming similarity as the objective function.

\paragraph{Dissimilarity in constraints}

There are various options to quantify the dissimilarity between feature-selection vectors.
We employ the following \emph{deselection dissimilarity} in combination with an adapted dissimilarity threshold:
\begin{equation}
	\text{dis}_{\text{des}}(s^{\text{new}}, s^{\text{old}}) = \sum_{j=1}^n (\lnot s^{\text{new}}_j \land s^{\text{old}}_j) \geq \min \left( \tau_{\text{abs}},~k^{\text{old}} \right)
	\label{eq:csd:constraint-dissimilarity}
\end{equation}
This dissimilarity counts how many of the previously selected features are \emph{not} selected in the new subgroup description.
These features may either be replaced by other features, or the total number of selected features may be reduced.
The constraint ensures that at least $\tau_{\text{abs}} \in \mathbb{N}$ features are deselected but never more than there were selected before ($k^{\text{old}}$), which would be infeasible.
For maximum dissimilarity, none of the previously selected features may be selected again.
Note that this dissimilarity measure is asymmetric, i.e., $\text{dis}_{\text{des}}(s^{\text{new}}, s^{\text{old}}) \neq \text{dis}_{\text{des}}(s^{\text{old}}, s^{\text{new}})$.
While this property would be an issue in a simultaneous search for multiple alternatives, i.e., without an explicit ordering, it is acceptable for sequential search, where `old' and `new' are well-defined.

Conceptually, one could also employ a more common dissimilarity measure like the Jaccard distance or the Dice dissimilarity~\cite{choi2010survey}.
The latter two are even symmetric and normalized to~$[0,1]$.
However, our deselection dissimilarity has two advantages:
First, if $s^{\text{old}}$ is constant, the dissimilarity is linear in $s^{\text{new}}$, as it amounts to a simple sum, even if the exact number of newly selected features is unknown yet.
This property is useful for solver-based search (cf.~Section~\ref{sec:csd:approach:alternatives:smt}).
In contrast, Jaccard distance and Dice dissimilarity involve a ratio and are therefore non-linear.
Second, the constraint from Equation~\ref{eq:csd:constraint-dissimilarity} is antimonotonic in the new feature selection, which is useful for heuristic search (cf.~Section~\ref{sec:csd:approach:alternatives:heuristics}).
Using the Jaccard distance or Dice dissimilarity in the constraint violates this property.
In particular, these dissimilarities can increase by selecting features that were not selected in the existing subgroup, i.e., an invalid feature set can become valid instead of remaining invalid by selecting further features.

\subsubsection{SMT Encoding}
\label{sec:csd:approach:alternatives:smt}

We only need to reformulate Equation~\ref{eq:csd:hamming-general} slightly to obtain a linear objective function regarding the alternative subgroup-membership vector~$b^{(a)}$:
\begin{equation}
	\begin{aligned}
		\text{sim}_{\text{nHamm}}(b^{(a)}, b^{(0)}) &= \frac{1}{m} \cdot \sum_{i=1}^m \left( b_i^{(a)} \leftrightarrow b_i^{(0)} \right) \\
		&= \frac{1}{m} \cdot \Big( \sum\limits_{\substack{i \in \{1, \dots, m\} \\ b_i^{(0)} = 1}} b_i^{(a)} + \sum\limits_{\substack{i \in \{1, \dots, m\} \\ b_i^{(0)} = 0}} \lnot b_i^{(a)} \Big)
	\end{aligned}
	\label{eq:csd:smt-hamming}
\end{equation}
In particular, since $b^{(0)}$~is known and therefore constant, we employ the expression from the second line, i.e., without the logical equivalence operator.
Instead, we compute two sums, one for data objects that are members of the original subgroup and one for non-members.
The negated expression $\lnot b_i^{(a)}$ may be expressed as $1 - b_i^{(a)}$.

To formulate the dissimilarity constraints, we leverage that the feature-selection vector~$s^{(l)}$ and the corresponding number of selected features~$k^{(l)}$ are known for all existing subgroups as well.
Thus, we instantiate and adapt Equation~\ref{eq:csd:constraint-dissimilarity} as follows:
\begin{equation}
	\forall l \in \{0, \dots, a-1\}:~ \text{dis}_{\text{des}}(s^{(a)}, s^{(l)}) = \sum_{\substack{j \in \{1, \dots, n\} \\ s^{(l)}_j = 1}} \lnot s^{(a)}_j \geq \min \left( \tau_{\text{abs}},~k^{(l)} \right)
	\label{eq:csd:smt-constraint-dissimilarity}
\end{equation}
In particular, we only sum over features that were selected in the existing subgroup and check whether they are deselected now.
To tie the variables~$s^{(a)}_j$ to the subgroup's bounds, we use Equation~\ref{eq:csd:smt-constraint-feature-selection}, which we already employed for feature cardinality constraints.

Finally, the overall SMT encoding of alternative-subgroup-description discovery (cf.~Definition~\ref{def:csd:alternative-subgroup-description-discovery}) combines the similarity objective (cf.~Equation~\ref{eq:csd:smt-hamming}) and dissimilarity constraints (cf.~Equation~\ref{eq:csd:smt-constraint-dissimilarity}) for alternatives with the previously introduced variables and constraints for bounds (cf.~Equation~\ref{eq:csd:smt-constraint-bounds-monotonic}), subgroup membership (cf.~Equation~\ref{eq:csd:smt-constraint-subgroup-membership}), and feature selection (cf.~Equation~\ref{eq:csd:smt-constraint-feature-selection}).
Optionally, one may add a feature-cardinality constraint (cf.~Equation~\ref{eq:csd:smt-constraint-feature-cardinalty}).

\subsubsection{Integration into Heuristic Search Methods}
\label{sec:csd:approach:alternatives:heuristics}

The situation here is similar to integrating feature-cardinality constraints into heuristic search methods (cf.~Section~\ref{sec:csd:approach:cardinality:heuristics}).
In particular, the constraint for alternatives based on the deselection dissimilarity (cf.~Equation~\ref{eq:csd:constraint-dissimilarity}) is antimonotonic in the subgroup's selected feature set.
I.e., the dissimilarity constraint is satisfied for an empty set of selected features, and once it is violated for a feature set, it remains violated for any superset.
Thus, the constraint type is suitable for heuristic search that iteratively enlarges the set of selected features, like \emph{PRIM} (cf.~Algorithm~\ref{al:csd:prim}), \emph{Beam Search} (cf.~Algorithms~\ref{al:csd:generic-beam-search} and~\ref{al:csd:beam-search-subgroup-update}), and \emph{Best Interval} (cf.~Algorithms~\ref{al:csd:generic-beam-search} and~\ref{al:csd:best-interval-subgroup-update}).
We only need to adapt the function \emph{get\_permissible\_feature\_idxs(\dots)} (cf.~Line~\ref{al:csd:prim:line:permissible-features} in Algorithm~\ref{al:csd:prim} and Line~\ref{al:csd:generic-beam-search:line:permissible-features} in Algorithm~\ref{al:csd:generic-beam-search}) to check the constraint.
I.e., the function should return the indices of all features that may be selected into the subgroup without violating the dissimilarity constraint (cf.~Equation~\ref{eq:csd:constraint-dissimilarity}).
In particular, once $k^{(l)} - \tau_{\text{abs}}$~features from an existing subgroup with $k^{(l)}$~features are selected again, no further features from this subgroup may be selected.

\subsubsection{Integration into Baselines}
\label{sec:csd:approach:alternatives:baselines}

Adapting our two baselines to alternative subgroup descriptions is less straightforward than to feature-cardinality constraints (cf.~Section~\ref{sec:csd:approach:cardinality:baselines}) since the optimization objective changes and the search space under the dissimilarity constraint (cf.~Equation~\ref{eq:csd:constraint-dissimilarity}) is harder to describe.
Thus, we did not implement and evaluate concrete adaptations but still discuss possible ideas in the following.

\paragraph{MORS}

A major issue for adapting \emph{MORS} (cf.~Algorithm~\ref{al:csd:mors}) is that \emph{MORS} is tailored to a particular objective, i.e., perfect subgroup quality in terms of recall.
In contrast, alternative subgroup descriptions should optimize subgroup-membership similarity to an original subgroup.
Also, the normalized Hamming similarity (cf.~Equation~\ref{eq:csd:hamming-general}) for alternatives measures accuracy rather than recall, i.e., it considers all data objects rather than only the positive ones.

For the dissimilarity constraint, we would like to enforce a valid feature set by implementing the function \emph{get\_permissible\_feature\_idxs(\dots)} in Line~\ref{al:csd:mors:line:permissible-features} of Algorithm~\ref{al:csd:mors} appropriately.
The univariate, quality-based selection heuristic we proposed for feature-cardinality constraints (cf.~Section~\ref{sec:csd:approach:cardinality:baselines}) may produce an invalid solution.
To alleviate this issue, we could adapt this heuristic as follows:
Still order the features by their univariate quality and iteratively select them in this order, but check the dissimilarity constraint in each iteration and skip over features that violate it.

\paragraph{Random Search}

For \emph{Random Search} (cf.~Algorithm~\ref{al:csd:random-search}), changing the optimization objective from subgroup quality to subgroup-membership similarity is not an issue since the objective is treated as a black-box function for evaluating randomly generated subgroups (cf.~Line~\ref{al:csd:random-search:line:optimum-update-start} of Algorithm~\ref{al:csd:random-search}).
For the dissimilarity constraint, we would like to implement \emph{get\_permissible\_feature\_idxs(\dots)} in Line~\ref{al:csd:random-search:line:permissible-features} of Algorithm~\ref{al:csd:random-search}) by uniformly sampling from the constrained search space.
In general, uniform sampling from a constrained space is a computationally hard problem~\cite{ermon2012uniform}, though it may be feasible for the particular constraint type.
We could also sample from the unconstrained space and then check the dissimilarity constraint, repeating sampling till a valid feature set is found.
However, this strategy may produce a high fraction of invalid solution candidates, depending on how strong the constraint is for the particular dataset and user parameters.
Another option would be using non-uniform sampling, e.g., only sample features not selected in any existing subgroup.
This guarantees constraint satisfaction but does not cover the entire constrained search space since it ignores the feature-set overlap allowed by the dissimilarity threshold~$\tau$.

\subsubsection{Time Complexity}
\label{sec:csd:approach:alternatives:complexity}

As for the feature-cardinality constraint (cf.~Section~\ref{sec:csd:approach:cardinality:complexity}), we analyze three aspects of time complexity:
the size of the search space for exhaustive search, parameterized complexity, and $\mathcal{NP}$-hardness.

\paragraph{Exhaustive search}

The search for an alternative subgroup description can iterate over the same solution candidates as the search for an original subgroup description, i.e., all combinations of bound values over features.
Thus, the results from Propositions~\ref{prop:csd:complexity-unconstrained-exhaustive} and~\ref{prop:csd:complexity-cardinality-exhaustive} remain valid:
\begin{proposition}[Complexity of exhaustive search for alternative-subgroup-description discovery]
	An exhaustive search for alternative-subgroup-description discovery (cf.~Definition~\ref{def:csd:alternative-subgroup-description-discovery})
	needs to evaluate $O(m^{2n})$ subgroups for each alternative in general or $O(n^k \cdot m^{2k})$ subgroups for each alternative if a feature-cardinality constraint (cf.~Definition~\ref{def:csd:feature-cardinality-constraint}) is employed.
	\label{prop:csd:complexity-alternatives-exhaustive}
\end{proposition}
The evaluation of solution candidates differs from the original subgroup descriptions but has a similar time complexity, i.e., $O(m \cdot n + a \cdot n)$ instead of $O(m \cdot n)$.
In particular, evaluating the subgroup-membership-similarity-based objective function (e.g.,~Equation~\ref{eq:csd:hamming-general}) should typically have a cost of~$O(m \cdot n)$, like subgroup-quality functions have.
Unlike the unconstrained case, some solution candidates violate the dissimilarity constraint (e.g.,~Equation~\ref{eq:csd:constraint-dissimilarity}) and need not be evaluated.
The corresponding constraint check requires determining the selected features and computing the dissimilarity.
The former (cf.~Definition~\ref{def:csd:feature-selection}) runs in~$O(n)$ if the minimum and maximum feature values of the dataset are precomputed.
The latter should typically entail a cost of~$O(n)$ per existing subgroup description for reasonably simple dissimilarity functions.

\paragraph{Parameterized complexity}

Due to the similar search space as for original subgroup descriptions, the parameterized-complexity results from Propositions~\ref{prop:csd:complexity-unconstrained-xp} and~\ref{prop:csd:complexity-cardinality-xp} apply to discovering alternative subgroup descriptions as well:
\begin{proposition}[Parameterized complexity of alternative-subgroup-description discovery]
	The problem of alternative-subgroup-description discovery (cf. Definition~\ref{def:csd:alternative-subgroup-description-discovery}) resides in the parameterized complexity class~$\mathcal{XP}$ for the parameter~$n$ in general and for the parameter~$k$ if a feature-cardinality constraint (cf.~Definition~\ref{def:csd:feature-cardinality-constraint}) is employed.
	\label{prop:csd:complexity-alternatives-xp}
\end{proposition}

\paragraph{NP-Hardness}

We prove $\mathcal{NP}$-completeness for a special case of alternative-subgroup-description discovery (cf.~Definition~\ref{def:csd:alternative-subgroup-description-discovery}) first.
To this end, we introduce the following definition:
\begin{definition}[Perfect alternative subgroup description]
	Given
	\begin{itemize}[noitemsep]
		\item a dataset $X \in \mathbb{R}^{m \times n}$,
		\item an original subgroup with subgroup membership~$b^{(0)} \in \{0, 1\}^m$ and feature selection~$s^{(0)} \in \{0, 1\}^n$,
		\item a dissimilarity measure $\text{dis}(\cdot)$ for feature-selection vectors of subgroups,
		\item and a dissimilarity threshold~$\tau \in \mathbb{R}_{\geq 0}$,
	\end{itemize}
	a \emph{perfect alternative subgroup description} defines a subgroup (cf.~Definition~\ref{def:csd:subgroup}) with membership~$b^{(a)} \in \{0, 1\}^m$ and feature selection~$s^{(a)} \in \{0, 1\}^n$ that exactly replicates the subgroup membership of the original subgroup, i.e., $b^{(a)} = b^{(0)}$, while being dissimilar regarding the feature selection, i.e., $\text{dis}(s^{(a)}, s^{(0)}) \geq \tau$.
	\label{def:csd:perfect-alternative}
\end{definition}
In particular, the value of the subgroup-membership similarity is fixed here rather than an optimization objective.
Similar to perfect subgroups (cf.~Definition~\ref{def:csd:perfect-subgroup}), perfect alternative subgroup descriptions only exist in some datasets.
Next, we define a corresponding search problem:
\begin{definition}[Perfect-alternative-subgroup-description discovery]
	Given
	\begin{itemize}[noitemsep]
		\item a dataset $X \in \mathbb{R}^{m \times n}$,
		\item an original subgroup with subgroup membership~$b^{(0)} \in \{0, 1\}^m$ and feature selection~$s^{(0)} \in \{0, 1\}^n$,
		\item a dissimilarity measure $\text{dis}(\cdot)$ for feature-selection vectors of subgroups,
		\item and a dissimilarity threshold~$\tau \in \mathbb{R}_{\geq 0}$,
	\end{itemize}
	\emph{perfect-alternative-subgroup-description discovery} is the problem of finding a perfect alternative subgroup description (cf.~Definition~\ref{def:csd:perfect-alternative}) if it exists or determining that it does not exist.
	\label{def:csd:perfect-alternative-subgroup-description-discovery}
\end{definition}
Next, we prove the following hardness result for this search problem with a perfect original subgroup and under a reasonable assumption on the notion of feature-selection dissimilarity (cf.~Appendix~\ref{sec:csd:appendix:proofs:complexity-perfect-alternatives-np-perfect-subgroup} for the proof):
\begin{proposition}[Complexity of perfect-alternative-subgroup-description discovery with feature-cardinality constraint and perfect original subgroup]
	Assuming
	\begin{itemize}[noitemsep]
		\item a combination of a dissimilarity measure~$\text{dis}(\cdot)$ and a dissimilarity threshold~$\tau \in \mathbb{R}_{\geq 0}$ that prevents selecting any selected feature from the original subgroup description again,
		\item and the original subgroup being perfect (cf.~Definition~\ref{def:csd:perfect-subgroup}),
	\end{itemize}
	the problem of perfect-alternative-subgroup-description discovery (cf.~Definition \ref{def:csd:perfect-alternative-subgroup-description-discovery}) with a feature-cardinality constraint (cf.~Definition~\ref{def:csd:feature-cardinality-constraint}) is $\mathcal{NP}$-complete.
	\label{prop:csd:complexity-perfect-alternatives-np-perfect-subgroup}
\end{proposition}
Our deselection dissimilarity (cf.~Equation~\ref{eq:csd:constraint-dissimilarity}) as~$\text{dis}(\cdot)$ satisfies the dissimilarity assumption if we choose a dissimilarity threshold~$\tau_{\text{abs}} \geq k^{\text{old}}$.
Other dissimilarity measures should typically also have such a threshold value that enforces zero overlap between the sets of selected features.

The problem naturally remains $\mathcal{NP}$-complete when dropping the two assumptions in Proposition~\ref{prop:csd:complexity-perfect-alternatives-np-perfect-subgroup}.
Nevertheless, we explicitly extend this result to imperfect original subgroups (cf.~Appendix~\ref{sec:csd:appendix:proofs:complexity-perfect-alternatives-np-imperfect-subgroup} for the proof):
\begin{proposition}[Complexity of perfect-alternative-subgroup-description discovery with feature-cardinality constraint and imperfect original subgroup]
	Assuming
	\begin{itemize}[noitemsep]
		\item a combination of a dissimilarity measure~$\text{dis}(\cdot)$ and a dissimilarity threshold~$\tau \in \mathbb{R}_{\geq 0}$ that prevents selecting any selected feature from the original subgroup description again,
		\item and the original subgroup \emph{not} being perfect (cf.~Definition~\ref{def:csd:perfect-subgroup}),
	\end{itemize}
	the problem of perfect-alternative-subgroup-description discovery (cf.~Definition \ref{def:csd:perfect-alternative-subgroup-description-discovery}) with a feature-cardinality constraint (cf.~Definition~\ref{def:csd:feature-cardinality-constraint}) is $\mathcal{NP}$-complete.
	\label{prop:csd:complexity-perfect-alternatives-np-imperfect-subgroup}
\end{proposition}
Finally, we switch from the search problem for perfect alternatives to the optimization problem of alternative-subgroup-description discovery.
We establish $\mathcal{NP}$-completeness under a reasonable assumption on the notion of subgroup-membership similarity (cf.~Appendix~\ref{sec:csd:appendix:proofs:complexity-alternatives-np} for the proof):
\begin{proposition}[Complexity of alternative-subgroup-description discovery with feature-cardinality constraint]
	Assuming
	\begin{itemize}[noitemsep]
		\item a combination of a dissimilarity measure~$\text{dis}(\cdot)$ and a dissimilarity threshold~$\tau \in \mathbb{R}_{\geq 0}$ that prevents selecting any selected feature from the original subgroup description again,
		\item and a similarity measure~$\text{sim}(\cdot)$ for which only perfect alternative subgroup descriptions (cf.~Definition~\ref{def:csd:perfect-alternative}) reach its maximal value regarding the original subgroup,
	\end{itemize}
	the problem of alternative-subgroup-description discovery (cf.~Definition~\ref{def:csd:alternative-subgroup-description-discovery}) with a feature-cardinality constraint (cf. Definition~\ref{def:csd:feature-cardinality-constraint}) is $\mathcal{NP}$-complete.
	\label{prop:csd:complexity-alternatives-np}
\end{proposition}
Normalized Hamming similarity (cf.~Equation~\ref{eq:csd:hamming-general}) as~$\text{sim}(\cdot)$ satisfies the similarity assumption since only perfect alternative subgroup descriptions yield the theoretical maximum similarity of~1 to the original subgroup description.

\section{Experimental Design}
\label{sec:csd:experimental-design}

In this section, we introduce our experimental design.
After a brief overview of its components (cf.~Section~\ref{sec:csd:experimental-design:overview}), we describe subgroup-discovery methods (cf.~Section~\ref{sec:csd:experimental-design:methods}), experimental scenarios (cf.~Section~\ref{sec:csd:experimental-design:scenarios}), evaluation metrics (cf.~Section~\ref{sec:csd:experimental-design:metrics}), and datasets (cf.~Section~\ref{sec:csd:experimental-design:datasets}).
Finally, we briefly outline our implementation (cf.~Section~\ref{sec:csd:experimental-design:implementation}).

\subsection{Overview}
\label{sec:csd:experimental-design:overview}

In our experiments, we evaluate eight subgroup-discovery methods on 27 binary-classification datasets.
We measure the subgroup quality in terms of nWRAcc and also record the methods' runtime.
We analyze four \emph{experimental scenarios}:
First, we compare all subgroup-discovery methods without constraints.
Second, we vary the timeout in solver-based search.
Third, we compare all subgroup-discovery methods with a feature-cardinality constraint, varying the cardinality threshold~$k$.
Fourth, we search for alternative subgroup descriptions with one solver-based and one heuristic search method.
We vary the number of alternatives~$a$ and the dissimilarity threshold~$\tau_{\text{abs}}$.

\subsection{Subgroup-Discovery Methods}
\label{sec:csd:experimental-design:methods}

We employ eight subgroup-discovery methods:
a solver-based one using our SMT encoding (cf.~Section~\ref{sec:csd:approach:smt}),
two exhaustive search methods from related work (cf.~Section~\ref{sec:csd:fundamentals:search:exhaustive}),
three heuristic search methods from related work (cf.~Section~\ref{sec:csd:fundamentals:search:heuristics}), and our two baselines (cf.~Section~\ref{sec:csd:baselines}).

\paragraph{Solver-based search}

For solver-based search, denoted as \emph{SMT}, we employ the solver \emph{Z3}~\cite{bjorner2015nuz, deMoura2008z3} with our SMT encoding of subgroup discovery (cf.~Equation~\ref{eq:csd:smt-problem-unconstrained-complete}).
This search method is exhaustive, i.e., it finds the global optimum, if granted sufficient time.
In practice, however, we set solver timeouts to control the runtime (cf.~Section~\ref{sec:csd:experimental-design:scenarios}), so the method may not be truly exhaustive.

\paragraph{Exhaustive algorithmic search}

We evaluate two exhaustive algorithmic search methods from related work (cf.~Section~\ref{sec:csd:fundamentals:search:exhaustive}), i.e., \emph{BSD}~\cite{lemmerich2010fast} and \emph{SD-Map}~\cite{atzmueller2006sd}, based on their implementation in the package \emph{SD4Py}~\cite{hudson2023subgroup}.
We chose these two methods after comparing the runtimes of several algorithms from four Python packages for subgroup discovery (cf.~Section~\ref{sec:csd:appendix:competitor-runtime}).
Both methods require discretizing numeric features, which makes them non-exhaustive regarding the full numeric search space in our experiments.
We use their built-in equal-width discretization and tune it by choosing the best subgroup quality out of ten different bin counts. i.e., $\{2, 3, 4, 5, 10, 15, 20, 30, 40, 50\}$.
Each subgroup description may comprise at most one bin for each feature.

\paragraph{Heuristic search}

We evaluate three heuristic search methods from related work (cf.~Section~\ref{sec:csd:fundamentals:search:heuristics}):
\emph{PRIM}~\cite{friedman1999bump} (cf.~Algorithm~\ref{al:csd:prim}), \emph{Beam Search} (cf.~Algorithms~\ref{al:csd:generic-beam-search} and~\ref{al:csd:beam-search-subgroup-update}), subsequently called \emph{Beam}, and \emph{Best Interval}~\cite{mampaey2012efficient} (cf.~Algorithms~\ref{al:csd:generic-beam-search} and~\ref{al:csd:best-interval-subgroup-update}), subsequently called \emph{BI}.
We set the peeling fraction of \emph{PRIM} to~$\alpha = 0.05$, consistent with other implementations~\cite{arzamasov2021reds, kwakkel2017exploratory} and within the recommended value range proposed by its authors~\cite{friedman1999bump}.
Further, we set the support threshold to~$\beta_0 = 0$, so there is no constraint on the subgroup's size.
For \emph{Beam} and \emph{BI}, we choose a beam width of $w=10$, falling between default values used in other implementations~\cite{arzamasov2021reds, lemmerich2019pysubgroup}.

\paragraph{Baselines}

We also include baselines that are simpler than the heuristic search methods.
In particular, we employ our own methods \emph{MORS} (cf.~Algorithm~\ref{al:csd:mors}) and \emph{Random Search} (cf.~Algorithm~\ref{al:csd:random-search}, subsequently called \emph{Random}).
\emph{MORS} is parameter-free.
For \emph{Random}, we set the number of iterations $\mathit{n\_iters} = 1000$.

\subsection{Experimental Scenarios}
\label{sec:csd:experimental-design:scenarios}

We evaluate the subgroup-discovery methods in four experimental scenarios.
Two of the scenarios do not involve all subgroup-discovery methods.

\paragraph{Unconstrained subgroup discovery}

Our first experimental scenario (cf. Section~\ref{sec:csd:evaluation:unconstrained} for results) compares all eight subgroup-discovery methods without constraints.
This comparison assesses the effectiveness of the solver-based search method \emph{SMT} for conventional subgroup discovery and serves as a reference point for subsequent experiments with constraints.
All methods except \emph{MORS} use WRAcc (cf.~Equation~\ref{eq:csd:wracc}) as the subgroup-quality function~$Q(\mathit{lb}, \mathit{ub}, X, y)$ for search.
\emph{MORS} optimizes its built-in objective (cf.~Definition~\ref{def:csd:minimal-optimal-recall-subgroup-discovery}).

\paragraph{Solver timeouts}

Our second experimental scenario (cf.~Section~\ref{sec:csd:evaluation:timeouts} for results) takes a deeper dive into \emph{SMT} as the subgroup-discovery method.
In particular, we analyze whether setting solver timeouts enables finding solutions with reasonable quality in a shorter time frame.
If the solver does not finish optimization within a given timeout, we record the currently best solution at this time, which may be suboptimal.
Note that the timeout only applies to the optimization procedure, while our runtime measurements also include the time for formulating the optimization problem upfront.

We evaluate twelve exponentially scaled timeout values, i.e., \{1~s, 2~s, 4~s, $\dots$, 2048~s\}.
In the three other experimental scenarios, we employ the maximum timeout of 2048~s for \emph{SMT}.
Since the heuristic search methods and baselines are significantly faster, we do not conduct a timeout analysis for them.
The two exhaustive algorithmic search methods \emph{BSD} and \emph{SD-Map} are usually fast as well.
However, they did not finish within two days in the unconstrained scenario for five datasets, and they do not have a timeout parameter.
We replace these missing values with the results for a feature-cardinality threshold of~$k=5$.

\paragraph{Feature-cardinality constraints}

Our third experimental scenario (cf.~Section~\ref{sec:csd:evaluation:cardinality} for results) analyzes feature-cardinality constraints (cf.~Section~\ref{sec:csd:approach:cardinality}) for all eight subgroup-discovery methods.
In particular, we study $k \in \{1, 2, 3, 4, 5\}$ selected features.
These values of~$k$ are upper bounds (cf.~Equation~\ref{eq:csd:smt-constraint-feature-cardinalty}), i.e., the subgroup-discovery methods may select fewer features if selecting more does not improve subgroup quality.

\paragraph{Alternative subgroup descriptions}

Our fourth experimental scenario (cf. Section~\ref{sec:csd:evaluation:alternatives} for results) studies alternative subgroup descriptions (cf.~Section~\ref{sec:csd:approach:alternatives}) for \emph{SMT} and \emph{Beam}, i.e., one solver-based and one heuristic search method.
This scenario still optimizes WRAcc (cf.~Equation~\ref{eq:csd:wracc}) for the original subgroup but normalized Hamming similarity (cf.~Equation~\ref{eq:csd:hamming-general}) for the alternatives.
The deselection dissimilarity (cf.~Equation~\ref{eq:csd:constraint-dissimilarity}) ensures that the alternatives are dissimilar enough.
We limit the feature cardinality to~$k=3$, which yields reasonably high subgroup quality (cf.~Section~\ref{sec:csd:evaluation:cardinality}).
We search for $a=5$ alternative subgroup descriptions with a dissimilarity threshold $\tau_{\text{abs}} \in \{1, 2, 3\}$.
Since each dataset has $n \geq 20$ features (cf.~Section~\ref{sec:csd:experimental-design:datasets}), our choices of~$a$, $k$, and~$\tau$ ensure that there always is a valid alternative.

\subsection{Evaluation Metrics}
\label{sec:csd:experimental-design:metrics}

\paragraph{Subgroup quality}

We use \emph{nWRAcc} (cf.~Equation~\ref{eq:csd:wracc-normalized}) to report subgroup quality.
To analyze how well the subgroup descriptions generalize, we conduct a stratified five-fold cross-validation.
In particular, each run of a subgroup-discovery method uses only 80\% of a dataset's data objects as training data, while the remaining data objects serve as test data.
Based on the bounds of each found subgroup, we determine subgroup membership for all data objects and compute \emph{training-set nWRAcc} and \emph{test-set nWRAcc} on the corresponding part of the data separately, using the true class labels~$y$.

\paragraph{Subgroup similarity}

For evaluating alternative subgroup descriptions, we consider not only their quality but also their induced subgroup's similarity to the original subgroup.
To this end, we use \emph{normalized Hamming similarity} (cf.~Equation~\ref{eq:csd:hamming-general}) and \emph{Jaccard similarity} (cf.~Equation~\ref{eq:csd:jaccard}) to compare subgroup membership of data objects between the original and the alternative.

\paragraph{Runtime}

As \emph{runtime}, we report the training time of the subgroup-discovery methods.
In particular, we measure how long the search for each subgroup takes.
For solver-based search, we also record whether the solver timed out.

\begin{table}[p]
	\centering
	\caption{
		Datasets from PMLB used in our experiments.
		$m$~denotes the number of data objects and $n$~the number of features.
		In dataset names, we replaced \emph{GAMETES\_Epistasis} with  \emph{GE\_} and \emph{GAMETES\_Heterogeneity} with \emph{GH\_} to reduce the table's width.
		\emph{Timeouts} indicates whether at least one timeout occurred with \emph{SMT} as the subgroup-discovery method and the highest timeout setting (2048~s), optimizing the original subgroup without cardinality constraints (\emph{Max~$k$}) or in any cardinality setting (\emph{Any~$k$}).
	}
	\begin{tabular}{lrrll}
		\toprule
		\multirow{2}{*}{Dataset} & \multirow{2}{*}{$m$} & \multirow{2}{*}{$n$} & \multicolumn{2}{c}{Timeouts} \\
		\cmidrule(lr){4-5}
		& & & Max~$k$ & Any~$k$ \\
		\midrule
		backache & 180 & 32 & No & No \\
		chess & 3196 & 36 & No & No \\
		churn & 5000 & 20 & Yes & Yes \\
		clean1 & 476 & 168 & No & No \\
		clean2 & 6598 & 168 & No & No \\
		coil2000 & 9822 & 85 & Yes & Yes \\
		credit\_g & 1000 & 20 & Yes & Yes \\
		dis & 3772 & 29 & No & No \\
		GE\_2\_Way\_20atts\_0.1H\_EDM\_1\_1 & 1600 & 20 & Yes & Yes \\
		GE\_2\_Way\_20atts\_0.4H\_EDM\_1\_1 & 1600 & 20 & No & No \\
		GE\_3\_Way\_20atts\_0.2H\_EDM\_1\_1 & 1600 & 20 & Yes & Yes \\
		GH\_20atts\_1600\_Het\_0.4\_0.2\_50\_EDM\_2\_001 & 1600 & 20 & Yes & Yes \\
		GH\_20atts\_1600\_Het\_0.4\_0.2\_75\_EDM\_2\_001 & 1600 & 20 & Yes & Yes \\
		Hill\_Valley\_with\_noise & 1212 & 100 & Yes & Yes \\
		horse\_colic & 368 & 22 & No & No \\
		hypothyroid & 3163 & 25 & No & No \\
		ionosphere & 351 & 34 & No & No \\
		molecular\_biology\_promoters & 106 & 57 & No & No \\
		mushroom & 8124 & 22 & No & No \\
		ring & 7400 & 20 & Yes & Yes \\
		sonar & 208 & 60 & No & Yes \\
		spambase & 4601 & 57 & No & Yes \\
		spect & 267 & 22 & No & No \\
		spectf & 349 & 44 & No & Yes \\
		tokyo1 & 959 & 44 & No & Yes \\
		twonorm & 7400 & 20 & Yes & Yes \\
		wdbc & 569 & 30 & No & No \\
		\bottomrule
	\end{tabular}
	\label{tab:csd:datasets}
\end{table}

\subsection{Datasets}
\label{sec:csd:experimental-design:datasets}

We use binary-classification datasets from the Penn Machine Learning Benchmarks (PMLB)~\cite{olson2017pmlb, romano2021pmlb}.
If the class labels occur with different frequencies, we encode the minority class as positive, i.e., assign~1 as its class label.
To avoid prediction scenarios that may be too easy or do not have enough features for alternative subgroup descriptions, we only select datasets with at least 100 data objects and 20 features.
Next, we exclude one dataset with 1000 features, which has a significantly higher dimensionality than all remaining datasets and would dominate runtime.
Finally, we manually exclude datasets that seem to be duplicated or modified versions of other datasets in our experiments.

Based on these criteria, we obtain 27 datasets with 106 to 9822 data objects and 20 to 168 features (cf.~Table~\ref{tab:csd:datasets}).
The datasets do not contain any missing values.
Further, PMLB encodes categorical features ordinally by default.

\subsection{Implementation and Execution}
\label{sec:csd:experimental-design:implementation}

We implemented all subgroup-discovery methods, experiments, and evaluations in Python~3.8.
The code is available on \emph{GitHub}\footnote{\url{https://github.com/Jakob-Bach/Constrained-Subgroup-Discovery}} and additionally backed up in the \emph{Software Heritage archive}\footnote{\href{https://archive.softwareheritage.org/swh:1:dir:8119c420804e9027b12cf3aaf82ce12157c049e1;origin=https://github.com/Jakob-Bach/Constrained-Subgroup-Discovery;visit=swh:1:snp:d9fd1a1757f1966b5e254154f41ddb4af476e124;anchor=swh:1:rev:bc3aafc904cc27214d458f93782104484ffc372d}{swh:1:dir:8119c420804e9027b12cf3aaf82ce12157c049e1}}.
A requirements file in our repository specifies the versions of all dependencies.
Further, we released the subgroup-discovery methods and evaluation metrics as the Python package \emph{csd}\footnote{\url{https://pypi.org/project/csd/}} on \emph{PyPI} to ease reuse.
Finally, we published all experimental data\footnote{\url{https://doi.org/10.35097/8ppb5x50nyvw1wa7}}.

Our experimental pipeline parallelizes over datasets, cross-validation folds, and subgroup-discovery methods, while each of these experimental tasks runs single-threaded.
We ran the pipeline on a server with 160~GB RAM and an \emph{AMD EPYC 7551} CPU, having 32~physical cores and a base clock of 2.0~GHz.
With this hardware, the parallelized pipeline run took approximately 34~hours.

\section{Evaluation}
\label{sec:csd:evaluation}

In this section, we evaluate our experiments.
In particular, we cover our four experimental scenarios, i.e., unconstrained subgroup discovery (cf.~Section~\ref{sec:csd:evaluation:unconstrained}), solver timeouts (cf.~Section~\ref{sec:csd:evaluation:timeouts}), feature-cardinality constraints (cf.~Section~\ref{sec:csd:evaluation:cardinality}), and alternative subgroup descriptions (cf.~Section~\ref{sec:csd:evaluation:alternatives}).
Finally, we summarize key experimental results (cf.~Section~\ref{sec:csd:evaluation:summary}).

\subsection{Unconstrained Subgroup Discovery}
\label{sec:csd:evaluation:unconstrained}

In this section, we compare all eight subgroup-discovery methods in the experimental scenario without constraints.
\emph{SMT} uses its default maximum solver timeout of 2048~s.

\begin{figure}[t]
	\centering
	\begin{subfigure}[t]{0.48\textwidth}
		\centering
		\includegraphics[width=\textwidth, trim=15 55 15 15, clip]{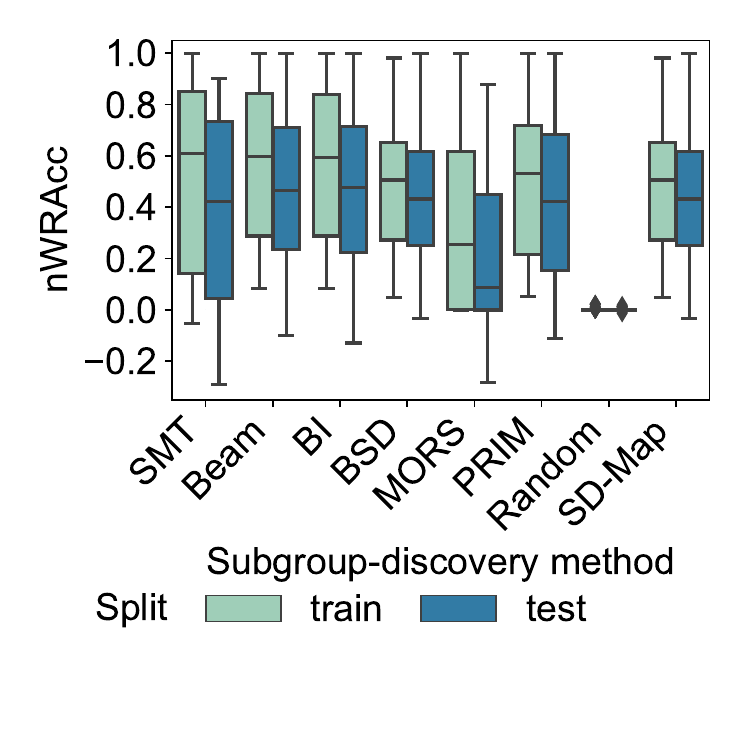}
		\caption{All 27 datasets.}
		\label{fig:csd:unconstrained-nwracc-all-datasets}
	\end{subfigure}
	\hfill
	\begin{subfigure}[t]{0.48\textwidth}
		\centering
		\includegraphics[width=\textwidth, trim=15 55 15 15, clip]{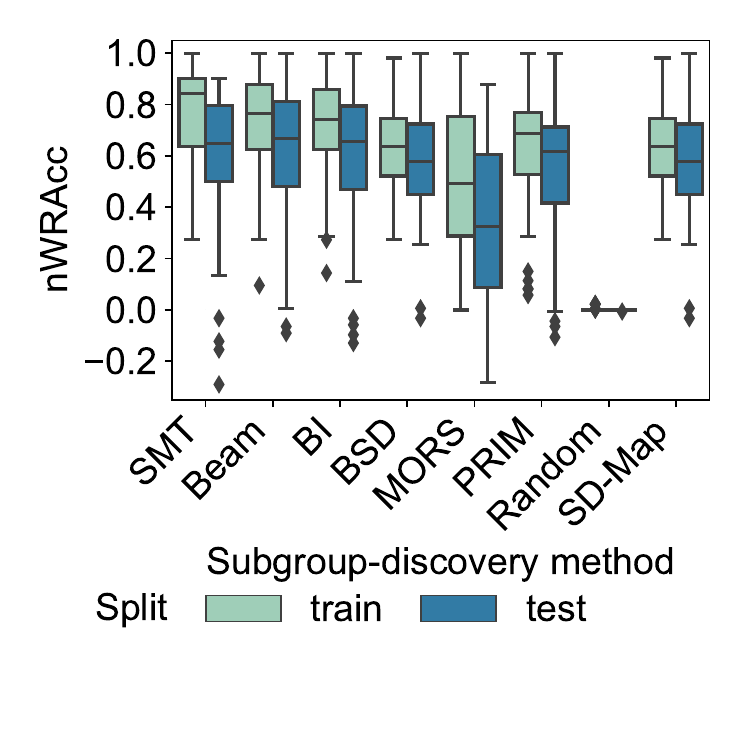}
		\caption{17 datasets without \emph{SMT} timeouts.}
		\label{fig:csd:unconstrained-nwracc-no-timeout-datasets}
	\end{subfigure}
	\caption{
		Distribution of subgroup quality over datasets and cross-validation folds, by subgroup-discovery method.
		Results from the unconstrained experimental scenario.
	}
	\label{fig:csd:unconstrained-nwracc}
\end{figure}

\paragraph{Subgroup quality}

Figure~\ref{fig:csd:unconstrained-nwracc-all-datasets} visualizes the subgroup quality achieved by the eight subgroup-discovery methods.
On the training set, the two heuristic search methods \emph{Beam} and \emph{BI} have roughly the same median nWRAcc as the solver-based search method \emph{SMT}.
In particular, the heuristics are even better than \emph{SMT} on some datasets but worse on others.
The former can only happen when \emph{SMT} yields suboptimal solutions due to timeouts, as we analyze later (cf.~Section~\ref{sec:csd:evaluation:timeouts}).
However, even if we limit our analysis to the 17 datasets without \emph{SMT} timeouts, \emph{Beam} and \emph{BI} are still remarkably close to the optimum quality (cf.~Figure~\ref{fig:csd:unconstrained-nwracc-no-timeout-datasets}).
This result is not specific to \emph{SMT} but also sets up \emph{Beam} and \emph{BI} as serious competitors for any other exhaustive search method.

On the test set, \emph{Beam} and \emph{BI} are even better than \emph{SMT} on median, also excluding timeout datasets, since their training-test nWRAcc difference is smaller.
This result indicates that \emph{Beam} and \emph{BI} are less susceptible to overfitting, so their solutions generalize better.
In detail, the average difference between training-set nWRAcc and test-set nWRAcc is 0.122 for \emph{SMT}, 0.101 for \emph{BI}, 0.095 for \emph{Beam}, 0.094 for \emph{MORS}, 0.068 for \emph{PRIM}, 0.042 for \emph{SD-Map}, 0.041 for \emph{BSD}, and 0.001 for \emph{Random}.

\emph{BSD} and \emph{SD-Map} yield worse average subgroup quality than \emph{Beam} and \emph{BI} as well.
While the former two are exhaustive, they require discretizing numeric features.
Thus, they effectively only have a fixed set of intervals to use as bounds instead of being able to choose the bounds independently at each possible feature value.
As a positive side-effect, this limitation may have reduced their overfitting and runtime.
\emph{PRIM}'s worse subgroup quality may equally arise from a reduced search space.
Although it follows an iterative subgroup-refinement procedure like  \emph{Beam} and \emph{BI}, its refinement options are more limited.
In particular, \emph{PRIM} always has to remove a fixed fraction~$\alpha$ of data objects from the subgroup (cf.~Algorithm~\ref{al:csd:prim}), while \emph{Beam} and \emph{BI} can remove more or fewer data objects.
Finally, the two baselines \emph{MORS} and \emph{Random} yield the lowest quality of all eight subgroup-discovery methods.
While \emph{Random} mostly yields the same quality as not restricting the subgroup at all, i.e., an nWRAcc of~0, \emph{MORS} is considerably above~0 and thus a suitable baseline for future studies comparing subgroup-discovery methods.

\begin{table}[t]
	\centering
	\caption{
		Aggregated runtime (in seconds) over datasets and cross-validation folds, by subgroup-discovery method.
		Results from the unconstrained experimental scenario.
	}
	\begin{subtable}{\textwidth}
		\centering
		\caption{
			All 27 datasets.
		}
		\begin{tabular}{lrrr}
			\toprule
			Method & Mean & Standard dev. & Median \\
			\midrule
			BI & 34.95 & 103.61 & 2.60 \\
			BSD & 55.70 & 151.46 & 1.06 \\
			Beam & 30.47 & 85.69 & 2.95 \\
			MORS & 0.01 & 0.00 & 0.01 \\
			PRIM & 1.26 & 1.51 & 0.66 \\
			Random & 0.91 & 0.95 & 0.51 \\
			SD-Map & 367.43 & 1143.05 & 4.08 \\
			SMT & 849.02 & 929.60 & 254.21 \\
			\bottomrule
		\end{tabular}
		\label{tab:csd:unconstrained-runtime-all-datasets}
	\end{subtable}
	\\ \vspace{\baselineskip}
	\begin{subtable}{\textwidth}
		\centering
		\caption{
			17 datasets without \emph{SMT} timeouts.
		}
		\begin{tabular}{lrrr}
			\toprule
			Method & Mean & Standard dev. & Median \\
			\midrule
			BI & 12.40 & 21.17 & 2.60 \\
			BSD & 44.02 & 128.33 & 1.57 \\
			Beam & 11.77 & 20.47 & 2.95 \\
			MORS & 0.01 & 0.00 & 0.01 \\
			PRIM & 1.29 & 1.62 & 0.80 \\
			Random & 0.82 & 0.89 & 0.56 \\
			SD-Map & 109.92 & 290.69 & 12.56 \\
			SMT & 168.13 & 243.11 & 57.23 \\
			\bottomrule
		\end{tabular}
		\label{tab:csd:unconstrained-runtime-no-timeout-datasets}
	\end{subtable}
	\label{tab:csd:unconstrained-runtime}
\end{table}

\paragraph{Runtime}

Table~\ref{tab:csd:unconstrained-runtime} summarizes the runtimes of the subgroup-discovery methods.
On average, \emph{SMT} is an order of magnitude slower than \emph{Beam} and \emph{BI}.
The exhaustive search methods with discretization, i.e., \emph{BSD} and \emph{SD-Map}, fall in between.
The heuristic \emph{PRIM} and the baseline \emph{Random} are yet another order of magnitude faster than \emph{Beam} and \emph{BI}.
Finally, the baseline \emph{MORS} runs in negligible time and, therefore, is a good tool for instantaneously obtaining a lower bound on subgroup quality.
Taking subgroup quality into consideration, the heuristic search methods offer a good quality in a short time.
Among the three heuristics, \emph{PRIM} is the fastest but yields the lowest subgroup quality, so users should decide which runtime is acceptable.

For \emph{SMT}, the overall runtime not only comprises optimization but also formulating the optimization problem.
Since the latter depends on the dataset size, e.g., involves $O(m)$ constraints with length $O(n)$ each to define the subgroup-membership variables~$b_i$ (cf.~Equation~\ref{eq:csd:smt-constraint-subgroup-membership}), the preparation time can become considerable for large datasets.
In our experiments, formulating the \emph{SMT} problem took 45~s on average, with a maximum of 379~s.
This average preparation time is already greater than the average total runtime of the heuristics.

\begin{table}[t]
	\centering
	\caption{
		Spearman correlation between runtime and metrics for dataset size,
		over datasets and cross-validation folds, by subgroup-discovery method.
		Results from the unconstrained experimental scenario, using the 17 datasets without \emph{SMT} timeouts.
	}
	\begin{tabular}{lrrrr}
		\toprule
		Method & $\Sigma n^u$ & $m \cdot n$ & $m$ & $n$ \\
		\midrule
		BI & 0.95 & 0.51 & 0.32 & 0.67 \\
		BSD & 0.46 & 0.60 & 0.44 & 0.42 \\
		Beam & 0.96 & 0.49 & 0.30 & 0.66 \\
		MORS & 0.27 & 0.57 & 0.51 & 0.26 \\
		PRIM & 0.84 & 0.56 & 0.29 & 0.76 \\
		Random & 0.58 & 0.69 & 0.42 & 0.77 \\
		SD-Map & 0.43 & 0.65 & 0.47 & 0.45 \\
		SMT & 0.39 & 0.73 & 0.70 & 0.23 \\
		\bottomrule
	\end{tabular}
	\label{tab:csd:unconstrained-runtime-correlation}
\end{table}

To determine which factors influence runtime, we analyze the Spearman correlation between runtime and four simple metrics for dataset size.
In particular, Table~\ref{tab:csd:unconstrained-runtime-correlation} considers the number of data objects~$m$, the number of features~$n$, the product of these two quantities~$m \cdot n$, and the number of unique values per feature summed over the features~$\Sigma n^u$.
For the three heuristic search methods, the latter metric shows a high correlation to runtime, while the three exhaustive search methods exhibit the highest runtime correlation to~$m \cdot n$.

\subsection{Solver Timeouts}
\label{sec:csd:evaluation:timeouts}

In this section, we evaluate the impact of solver timeouts for \emph{SMT} search.

\begin{figure}[t]
	\centering
	\begin{subfigure}[t]{0.48\textwidth}
		\centering
		\includegraphics[width=\textwidth, trim=15 25 15 10, clip]{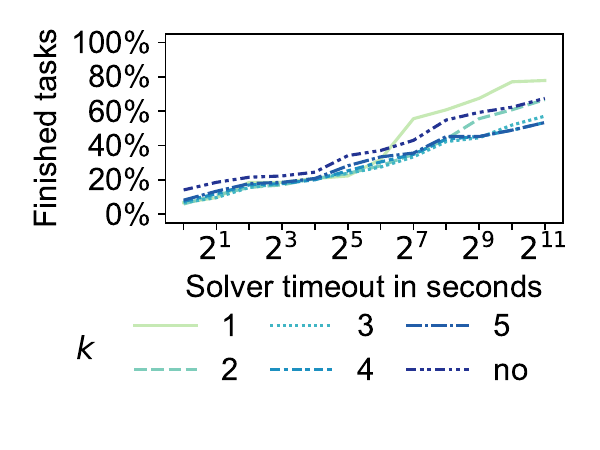}
		\caption{
			Frequency of finished \emph{SMT} tasks over datasets and cross-validation folds, by feature cardinality~$k$.
		}
		\label{fig:csd:timeouts-finished-tasks}
	\end{subfigure}
	\hfill
	\begin{subfigure}[t]{0.48\textwidth}
		\centering
		\includegraphics[width=\textwidth, trim=15 25 15 10, clip]{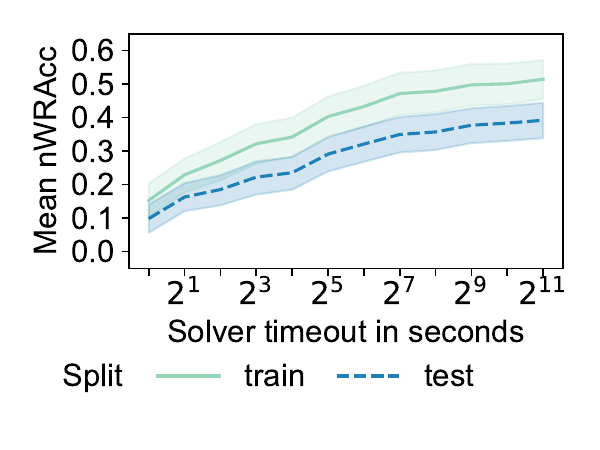}
		\caption{
			Mean subgroup quality, with 95\% confidence intervals based on datasets and cross-validation folds.
			Results from the unconstrained experimental scenario.
		}
		\label{fig:csd:timeouts-nwracc}
	\end{subfigure}
	\caption{
		Impact of solver timeouts for \emph{SMT} as the subgroup-discovery method.
		Results from the search for original subgroups.
	}
	\label{fig:csd:timeouts}
\end{figure}

\paragraph{Finished tasks}

Figure~\ref{fig:csd:timeouts-finished-tasks} displays how many of the \emph{SMT} optimization tasks for original subgroups finished within the evaluated solver timeouts.
Besides the unconstrained tasks, we also consider tasks with different feature-cardinality thresholds, though the overall trend is the same.
In particular, the number of finished tasks only increases slowly over time, and some tasks take orders of magnitude longer than others.
E.g., in the unconstrained experimental scenario, 21.5\% of the \emph{SMT} tasks finished within 4~s, 24.4\% within 16~s, 37.0\% within 64~s, 54.8\% within 256~s, and 62.2\% within 1024~s.
For the maximum timeout setting of 2048~s, 67.4\% of the \emph{SMT} tasks finished, and 17 out of 27 datasets did not encounter timeouts (cf.~Table~\ref{tab:csd:datasets}).

\paragraph{Subgroup quality}

Figure~\ref{fig:csd:timeouts-nwracc} visualizes the subgroup quality over solver timeouts for unconstrained \emph{SMT} search.
This plot uses the quality of the optimal solution for finished tasks and of the currently best solution for unfinished tasks.
As for the number of finished tasks (cf.~Figure~\ref{fig:csd:timeouts-finished-tasks}), the largest gains occur in the first minute.
E.g., the mean test-set nWRAcc over datasets and cross-validation folds is 0.10 for 1~s, 0.19 for 4~s, 0.24 for 16~s, 0.32 for 64~s, and 0.39 for the maximum solver timeout of 2024~s.
The main cause for this trend is that many tasks finish relatively early (cf.~Figure~\ref{fig:csd:timeouts-finished-tasks}), and finished tasks cannot improve their quality for higher solver timeouts.
In contrast, if we only consider the tasks where the solver did not finish even within the maximum solver timeout, the quality increase of the currently best solution over time is marginal.

Further, even \emph{SMT} with a timeout does not compare favorably to fast heuristic search methods.
E.g., with a solver timeout of 64~s, corresponding to an average overall runtime of 88~s, \emph{SMT} achieves a mean training-set nWRAcc of 0.43, compared to 0.56 for \emph{Beam} with an average runtime of 30~s (cf.~Table~\ref{tab:csd:unconstrained-runtime-all-datasets}).

Finally, setting a lower solver timeout decreases overfitting, i.e., the difference between training-set nWRAcc and test-set nWRAcc increases over time (cf.~Figure~\ref{fig:csd:timeouts-nwracc}).
However, since the test-set nWRAcc still increases with the timeout, choosing lower timeouts does not help quality-wise.

\subsection{Feature-Cardinality Constraints}
\label{sec:csd:evaluation:cardinality}

In this section, we compare all eight subgroup-discovery methods in the experimental scenario with feature-cardinality constraints.
\emph{SMT} uses its default maximum solver timeout of 2048~s.

\begin{figure}[p]
	\centering
	\begin{subfigure}[t]{0.48\textwidth}
		\centering
		\includegraphics[width=\textwidth, trim=15 60 15 15, clip]{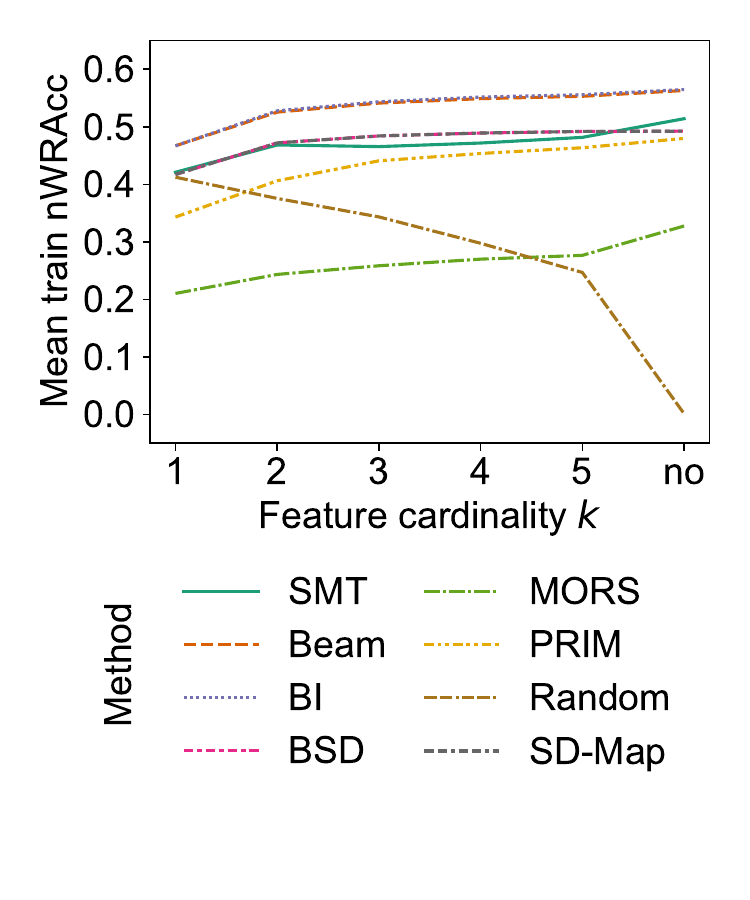}
		\caption{All 27 datasets, training set.}
		\label{fig:csd:cardinality-train-nwracc-all-datasets}
	\end{subfigure}
	\hfill
	\begin{subfigure}[t]{0.48\textwidth}
		\centering
		\includegraphics[width=\textwidth, trim=15 60 15 15, clip]{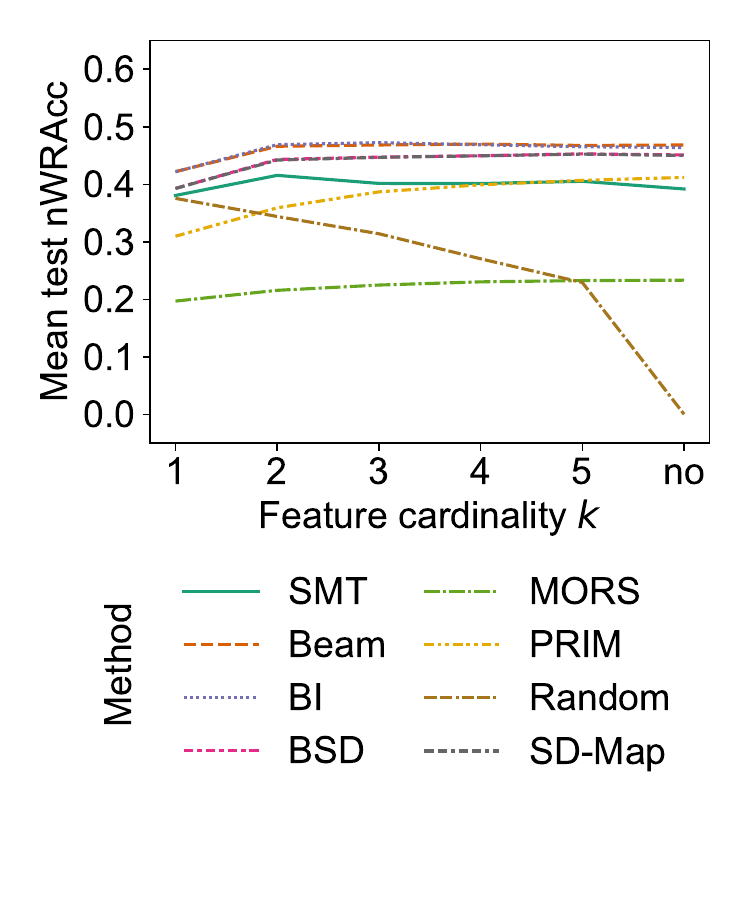}
		\caption{All 27 datasets, test set.}
		\label{fig:csd:cardinality-test-nwracc-all-datasets}
	\end{subfigure}
	\\ \vspace{\baselineskip}
	\begin{subfigure}[t]{0.48\textwidth}
		\centering
		\includegraphics[width=\textwidth, trim=15 60 15 15, clip]{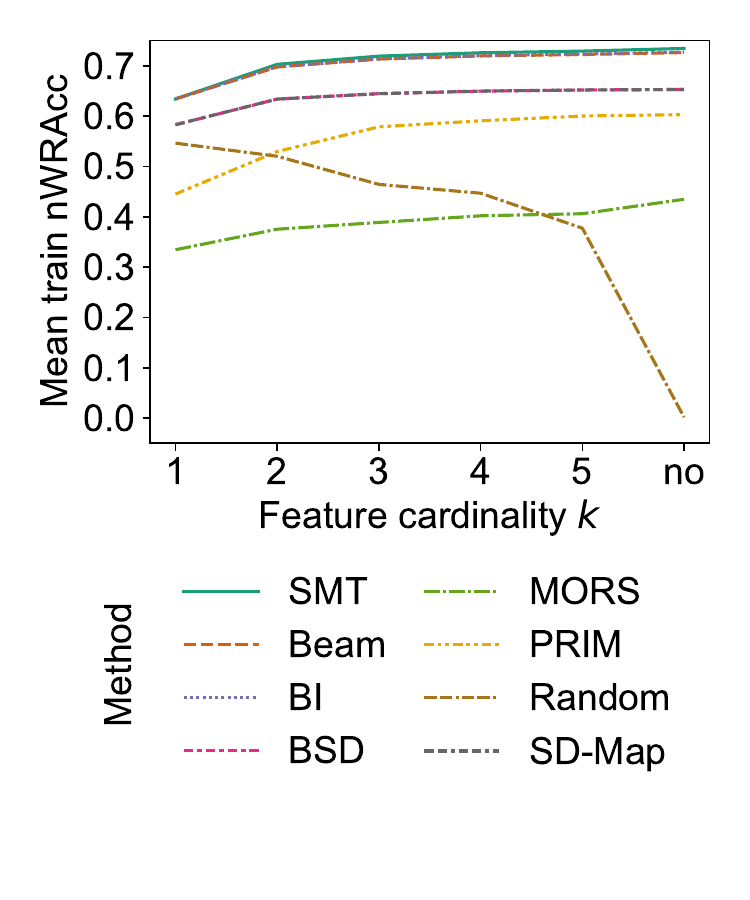}
		\caption{13 datasets without \emph{SMT} timeouts, training set.}
		\label{fig:csd:cardinality-train-nwracc-no-timeout-datasets}
	\end{subfigure}
	\hfill
	\begin{subfigure}[t]{0.48\textwidth}
		\centering
		\includegraphics[width=\textwidth, trim=15 60 15 15, clip]{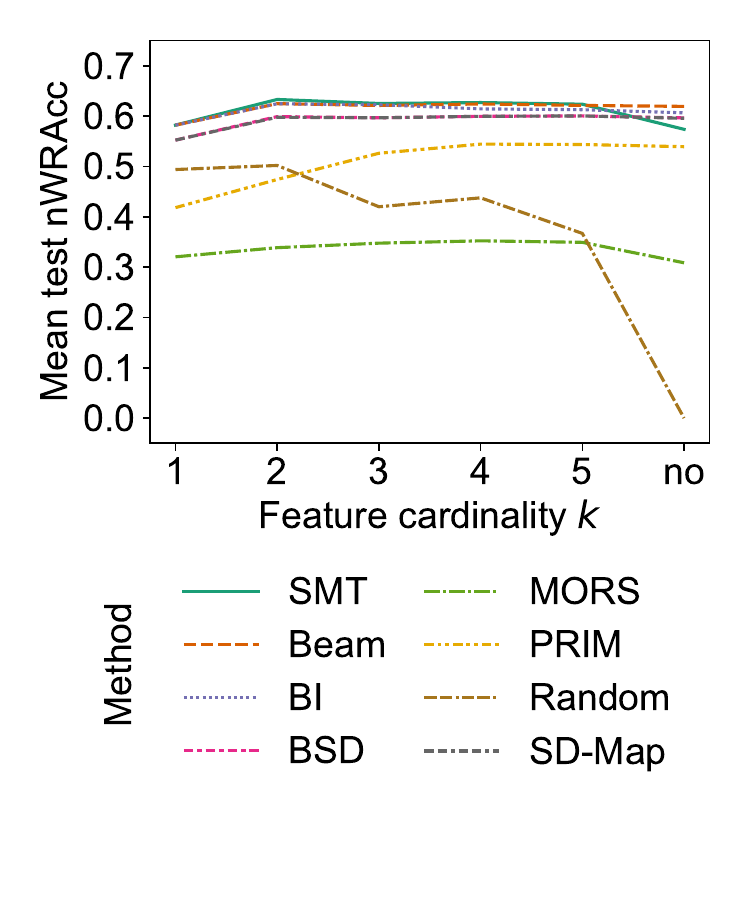}
		\caption{13 datasets without \emph{SMT} timeouts, test set.}
		\label{fig:csd:cardinality-test-nwracc-no-timeout-datasets}
	\end{subfigure}
	\caption{
		Mean subgroup quality over datasets and cross-validation folds, by subgroup-discovery method and feature cardinality~$k$.
		Results from the search for original subgroups.
	}
	\label{fig:csd:cardinality-nwracc}
\end{figure}

\paragraph{Subgroup quality}

Figure~\ref{fig:csd:cardinality-nwracc} displays the mean subgroup quality, averaging over datasets and cross-validation folds, for different values of the feature-cardinality threshold~$k$.
For most subgroup-discovery methods, mean training-set nWRAcc (cf.~Figure~\ref{fig:csd:cardinality-train-nwracc-all-datasets}) increases with~$k$, though the marginal utility decreases.
In particular, even with $k=1$, the mean nWRAcc is already clearly above 50\% of the nWRAcc achieved with all features.
Further, the quality increase between $k=1$ and $k=2$ is usually the largest.
On the test set (cf.~Figure~\ref{fig:csd:cardinality-test-nwracc-all-datasets}), the benefit of a larger~$k$ is even smaller:
The mean test-set nWRAcc of all methods except \emph{PRIM} barely improves beyond $k=2$.
These results indicate that sparse subgroup descriptions already yield a high subgroup quality.

The baseline \emph{Random} even improves subgroup quality with lower~$k$ due to its particular design (cf.~Algorithm~\ref{al:csd:random-search}):
It randomly samples bounds for each feature independently.
Thus, each feature excludes a certain fraction of data objects from the subgroup.
The more features are used in the subgroup description, the smaller the expected number of data objects in the subgroup becomes.
Since the number of subgroup members is one factor in WRAcc (cf.~Equation~\ref{eq:csd:wracc}), quality naturally decreases for smaller subgroups.

Figures~\ref{fig:csd:cardinality-train-nwracc-all-datasets} and~\ref{fig:csd:cardinality-test-nwracc-all-datasets} also reveal that the heuristic search methods \emph{Beam} and \emph{BI} still yield higher average subgroup quality than the solver-based search \emph{SMT} due to timeouts, for any feature-cardinality setting.
Even excluding the datasets with \emph{SMT} timeouts (cf.~Figures~\ref{fig:csd:cardinality-train-nwracc-no-timeout-datasets} and~\ref{fig:csd:cardinality-test-nwracc-no-timeout-datasets}), these two heuristics yield nearly the same average subgroup quality as \emph{SMT} for constrained~$k$ and have an advantage on the test set with unconstrained~$k$.
\emph{BSD} and \emph{SD-Map} do not beat \emph{Beam} and \emph{BI} either, suffering from their discretized search space.
The heuristic \emph{PRIM} exhibits a larger increase of subgroup quality over~$k$ than \emph{Beam} and \emph{BI}, thereby narrowing the quality gap to the latter.
The baseline \emph{MORS} displays the least effect of~$k$ on mean test-set nWRAcc, showing very stable subgroup quality.

Finally, the results indicate that limiting~$k$ reduces overfitting.
For example, for \emph{Beam}, the mean difference between training-set and test-set nWRAcc is 0.095 without a feature-cardinality constraint, 0.073 for $k=3$, and 0.045 for $k=1$.
The increasing tendency to overfit with larger~$k$ explains why mean training-set nWRAcc increases more than mean test-set nWRAcc over~$k$ in Figure~\ref{fig:csd:cardinality-nwracc}.
From the eight subgroup-discovery methods, \emph{BSD} and \emph{SD-Map} show the smallest increase of overfitting over~$k$, \emph{MORS} and \emph{SMT} the largest.

\begin{table}[t]
	\centering
	\caption{
		Mean runtime (in seconds) over datasets and cross-validation folds, by subgroup-discovery method and feature cardinality~$k$.
		Results from the search for original subgroups.
	}
	\begin{tabular}{lrrrrrr}
		\toprule
		$k$ & 1 & 2 & 3 & 4 & 5 & no \\
		\midrule
		BI & 7.8 & 11.7 & 14.2 & 16.7 & 18.7 & 35.0 \\
		BSD & 0.9 & 0.9 & 0.9 & 2.7 & 29.5 & 55.7 \\
		Beam & 6.8 & 10.1 & 12.8 & 14.6 & 16.1 & 30.5 \\
		MORS & 0.0 & 0.0 & 0.0 & 0.0 & 0.0 & 0.0 \\
		PRIM & 0.1 & 0.2 & 0.3 & 0.3 & 0.5 & 1.3 \\
		Random & 0.6 & 0.6 & 0.6 & 0.7 & 0.7 & 0.9 \\
		SD-Map & 2.3 & 3.3 & 9.6 & 54.0 & 345.2 & 367.4 \\
		SMT & 648.2 & 911.3 & 1091.7 & 1113.4 & 1117.4 & 849.0 \\
		\bottomrule
	\end{tabular}
	\label{tab:csd:cardinality-runtime}
\end{table}

\paragraph{Runtime}

As Table~\ref{tab:csd:cardinality-runtime} displays, the three heuristic search methods as well as \emph{BSD} and \emph{SD-Map} become faster the smaller~$k$ is.
The baseline \emph{Random} shows a similar trend, though less prominent, while \emph{MORS} yields results instantaneously in any case.
In contrast, the picture for the solver-based search method \emph{SMT} is less clear.
While its average runtime clearly increases from~$k=1$ till $k=3$, it roughly remains constant for $k \in \{4, 5\}$ and even decreases without a feature-cardinality constraint, only remaining higher than for $k = 1$.

\subsection{Alternative Subgroup Descriptions}
\label{sec:csd:evaluation:alternatives}

In this section, we analyze alternative subgroup descriptions for the subgroup-discovery methods \emph{Beam} and \emph{SMT}.
Both employ a feature-cardinality threshold of~$k=3$.
\emph{SMT} uses its default maximum solver timeout of 2048~s.

\begin{figure}[t]
	\centering
	\begin{subfigure}[t]{0.48\textwidth}
		\centering
		\includegraphics[width=\textwidth, trim=15 50 15 15, clip]{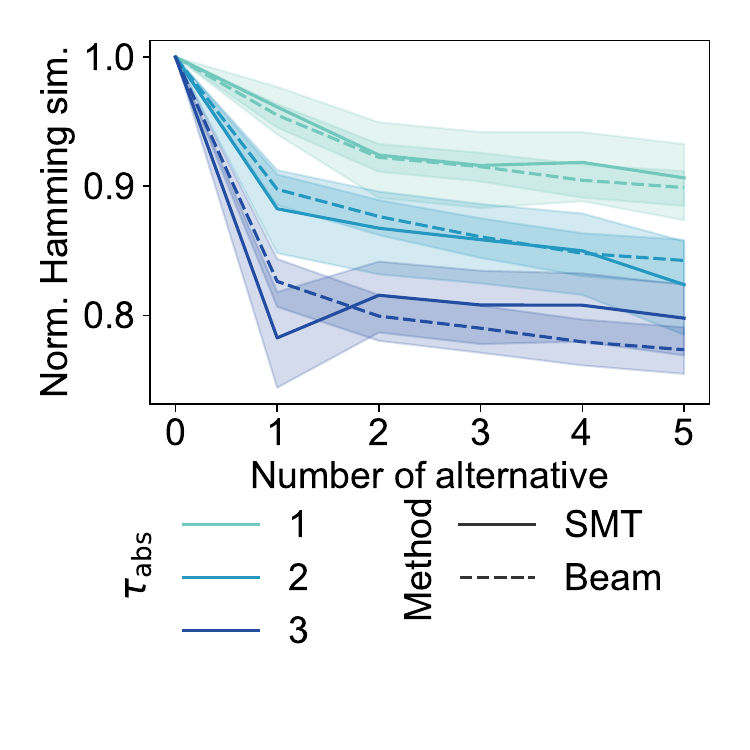}
		\caption{Normalized Hamming similarity.}
		\label{fig:csd:alternatives-hamming}
	\end{subfigure}
	\hfill
	\begin{subfigure}[t]{0.48\textwidth}
		\centering
		\includegraphics[width=\textwidth, trim=15 50 15 15, clip]{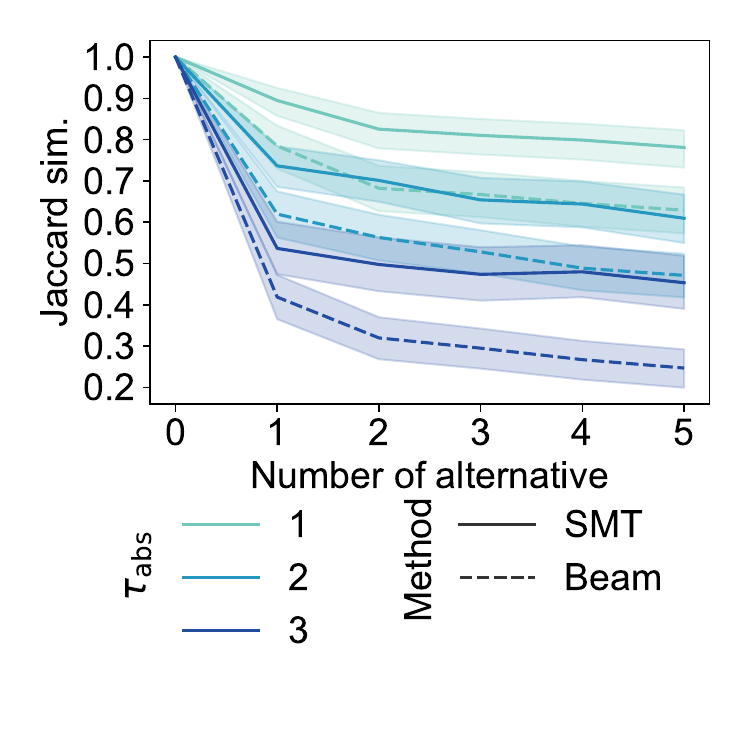}
		\caption{Jaccard similarity.}
		\label{fig:csd:alternatives-jaccard}
	\end{subfigure}
	\caption{
		Mean subgroup similarity of alternative subgroup descriptions to the original subgroup, with 95\% confidence intervals based on datasets and cross-validation folds, by subgroup-discovery method, number of alternative, and dissimilarity threshold~$\tau_{\text{abs}}$.
	}
	\label{fig:csd:alternatives-similarity}
\end{figure}

\paragraph{Subgroup similarity}

Figure~\ref{fig:csd:alternatives-similarity} visualizes the average similarity between the original subgroups and those induced by alternative subgroup descriptions.
As one would expect, the subgroup-membership similarity decreases the more alternatives one desires and the more the selected features in subgroup descriptions should differ.
Further, the decrease is strongest from the original subgroup, i.e., the zeroth alternative, to the first alternative but smaller beyond.
This observation indicates that one may find several alternative subgroup descriptions of comparable similarity to the original.

These trends hold for both similarity measures, i.e., the normalized Hamming similarity we optimize (cf.~Equation~\ref{eq:csd:hamming-general} and Figure~\ref{fig:csd:alternatives-hamming}) as well as the Jaccard similarity (cf.~Equation~\ref{eq:csd:jaccard} and Figure~\ref{fig:csd:alternatives-jaccard}).
The latter yields lower similarity values than the former since it ignores data objects that are not contained in either compared subgroup.
Further, the observed trends exist for the solver-based search method \emph{SMT} as well as the heuristic search method \emph{Beam}.
\emph{SMT} yields clearly more similar subgroups than \emph{Beam} for the Jaccard similarity, while the normalized Hamming similarity does not show a clear winner.

\begin{figure}[t]
	\centering
	\begin{subfigure}[t]{0.48\textwidth}
		\centering
		\includegraphics[width=\textwidth, trim=15 50 15 15, clip]{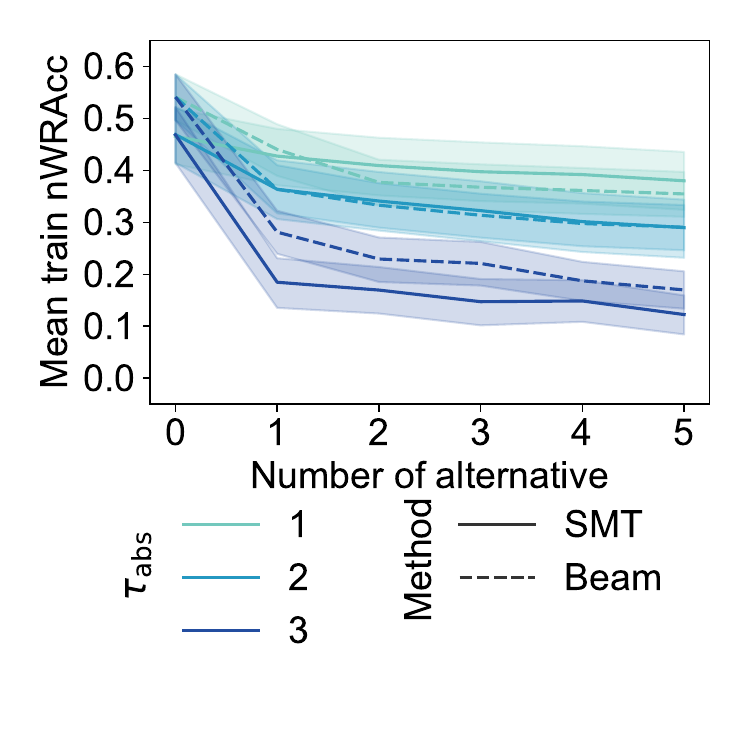}
		\caption{Training set.}
		\label{fig:csd:alternatives-train-nwracc}
	\end{subfigure}
	\hfill
	\begin{subfigure}[t]{0.48\textwidth}
		\centering
		\includegraphics[width=\textwidth, trim=15 50 15 15, clip]{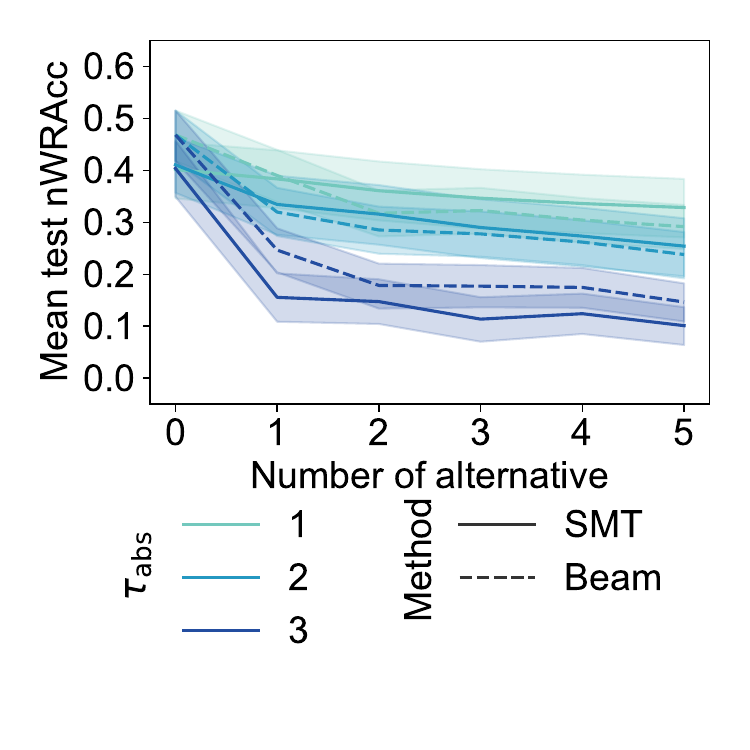}
		\caption{Test set.}
		\label{fig:csd:alternatives-test-nwracc}
	\end{subfigure}
	\caption{
		Mean subgroup quality of alternative subgroup descriptions, with 95\% confidence intervals based on datasets and cross-validation folds, by subgroup-discovery method, number of alternative, and dissimilarity threshold~$\tau_{\text{abs}}$.
	}
	\label{fig:csd:alternatives-nwracc}
\end{figure}

\paragraph{Subgroup quality}

The average subgroup quality of alternative subgroup descriptions (cf.~Figure~\ref{fig:csd:alternatives-nwracc}) shows similar trends as subgroup similarity (cf.~Figure~\ref{fig:csd:alternatives-similarity}).
In particular, quality decreases over the dissimilarity threshold~$\tau_{\text{abs}}$ and over the number of alternatives~$a$, with the largest decrease to the first alternative.
For the highest dissimilarity threshold~$\tau_{\text{abs}} = 3$, \emph{Beam} consistently yields higher average quality than \emph{SMT} for the original subgroup and each alternative.
In contrast, the other two values of the dissimilarity threshold do not clearly favor either subgroup-discovery method.
The observed trends on the test set (cf.~Figure~\ref{fig:csd:alternatives-test-nwracc}) are very similar to those on the training set (cf.~Figure~\ref{fig:csd:alternatives-train-nwracc}).
For both subgroup-discovery methods, overfitting, as measured by the training-test difference in nWRAcc, is lower for the alternative subgroup descriptions than for the original subgroups.
This phenomenon may result from the alternative subgroup descriptions not directly optimizing subgroup quality.

\begin{table}[t]
	\centering
	\caption{
		Mean runtime (in seconds) over datasets and cross-validation folds, by subgroup-discovery method, dissimilarity threshold~$\tau_{\text{abs}}$, and number of alternative.
		Results from the search for alternative subgroup descriptions.
	}
	\begin{tabular}{llrrrrrr}
		\toprule
		\multirow{2}{*}{Method} & \multirow{2}{*}{$\tau_{\text{abs}}$} & \multicolumn{6}{c}{Number of alternative} \\
		\cmidrule(lr){3-8}
		& &  0 & 1 & 2 & 3 & 4 & 5 \\
		\midrule
		\multirow[t]{3}{*}{Beam} & 1 & 12.8 & 8.0 & 7.6 & 7.3 & 7.3 & 7.3 \\
		& 2 & 12.8 & 7.7 & 7.4 & 7.2 & 7.0 & 6.8 \\
		& 3 & 12.8 & 5.8 & 5.1 & 4.7 & 4.1 & 3.5 \\
		\multirow[t]{3}{*}{SMT} & 1 & 1091.7 & 166.0 & 221.5 & 239.6 & 258.1 & 277.9 \\
		& 2 & 1105.2 & 377.5 & 463.5 & 537.5 & 599.4 & 658.3 \\
		& 3 & 1107.4 & 869.1 & 670.8 & 597.6 & 588.1 & 557.6 \\
		\bottomrule
	\end{tabular}
	\label{tab:csd:alteratives-runtime}
\end{table}

\paragraph{Runtime}

Table~\ref{tab:csd:alteratives-runtime} displays the average runtime for searching original and alternative subgroup descriptions.
The search for alternatives is faster for both analyzed search methods, i.e., \emph{Beam} and \emph{SMT}.
As for the original subgroups, \emph{Beam} search for alternatives is one to two orders of magnitude faster than the solver-based \emph{SMT} search.
For \emph{Beam}, runtime tends to decrease over the number of alternatives, while \emph{SMT} shows a less clear behavior.
In particular, its runtime increases over alternatives for~$\tau_{\text{abs}} \in \{1, 2\}$, i.e., settings that allow reusing features from previous subgroup descriptions.
In contrast, runtime decreases over alternatives for $\tau_{\text{abs}} = k = 3$, which forbids reusing any feature selected before.
Finally, the number of \emph{SMT} tasks finished within the solver timeout behaves similarly to the runtime.
In particular, there are more finished tasks when searching for alternative subgroup descriptions than for original subgroups.

\subsection{Summary}
\label{sec:csd:evaluation:summary}

\paragraph{Unconstrained subgroup discovery (cf.~Section~\ref{sec:csd:evaluation:unconstrained})}

We recommend using the heuristic search methods \emph{Beam} and \emph{BI}, which were overall best.
These two methods from the literature yielded close-to-optimal quality while being significantly faster than our exhaustive, solver-based search method \emph{SMT}.
Additionally, heuristic solutions were less prone to overfitting than optimal solutions from exhaustive search.
These results highlight the heuristics as serious competitors for any exhaustive search method from the literature, not only our method \emph{SMT}.
Further, \emph{Beam} and \emph{BI} beat the quality of the exhaustive competitors \emph{BSD} and \emph{SD-Map}, which required discretizing numeric features beforehand.
The heuristic search method \emph{PRIM} was faster than \emph{Beam} and \emph{BI} but yielded lower subgroup quality.
Our novel baseline \emph{MORS} provided instantaneous, non-trivial lower bounds for subgroup quality.

\paragraph{Solver timeouts (cf.~Section~\ref{sec:csd:evaluation:timeouts})}

Setting larger solver timeouts showed a decreasing marginal utility regarding the number of finished \emph{SMT} tasks and subgroup quality, i.e., most gains occurred within the first few seconds or dozens of seconds.
About half the \emph{SMT} tasks that finished at all finished in under a minute.
However, the average subgroup quality with this solver timeout was lower than for heuristic search methods with even lower runtime.

\paragraph{Feature-cardinality constraints (cf.~Section~\ref{sec:csd:evaluation:cardinality})}

Using more features in subgroup descriptions showed a decreasing marginal utility regarding subgroup quality.
For most subgroup-discovery methods, subgroups with as few as $k=2$ features in their description often yielded already a similar quality as unconstrained subgroups, i.e., using all features.
Further, feature-cardinality constraints reduced overfitting and sped up the heuristic search methods as well as \emph{BSD} and \emph{SD-Map}.
These results speak for using small feature sets in subgroup descriptions, which may also benefit interpretability for users.

\paragraph{Alternative subgroup descriptions (cf.~Section~\ref{sec:csd:evaluation:alternatives})}

The heuristic \emph{Beam} was one to two orders of magnitude faster than solver-based \emph{SMT} when searching for alternative subgroup descriptions, while both search methods found alternatives faster than original subgroups.
The quality and similarity of alternative subgroup descriptions strongly depended on two user parameters, i.e., the number of alternatives~$a$ and the dissimilarity threshold on feature selection~$\tau_{\text{abs}}$.
The difference in quality and similarity between the original and the first alternative was higher than among the first few alternatives.
In particular, there may be several promising alternatives from which users may choose one.

\section{Related Work}
\label{sec:csd:related-work}

In this section, we review related work.
Next to the literature on subgroup discovery (cf.~Section~\ref{sec:csd:related-work:subgroup-discovery}), we also discuss relevant work from the adjacent field of feature selection (cf.~Section~\ref{sec:csd:related-work:feature-selection}) and other related areas (cf.~Section~\ref{sec:csd:related-work:other}).

\subsection{Subgroup Discovery}
\label{sec:csd:related-work:subgroup-discovery}

In this section, we present related work from the field of subgroup discovery.
First, we discuss algorithmic search methods (cf.~Section~\ref{sec:csd:related-work:subgroup-discovery:algorithmic-methods}) as well as white-box formulations (cf.~Section~\ref{sec:csd:related-work:subgroup-discovery:white-box}) for this problem.
Second, we cover constrained subgroup discovery in general (cf.~Section~\ref{sec:csd:related-work:subgroup-discovery:constraints}) and for the two constraint types we focus on, i.e., feature-cardinality constraints (cf.~Section~\ref{sec:csd:related-work:subgroup-discovery:cardinality}) and alternative subgroup descriptions (cf.~Section~\ref{sec:csd:related-work:subgroup-discovery:alternatives}).

\subsubsection{Algorithmic Search Methods}
\label{sec:csd:related-work:subgroup-discovery:algorithmic-methods}

The terms `subgroup' and `subgroup discovery' have different meanings in different communities, and similar concepts have been reinvented over time.
Our problem definition (Section~\ref{sec:csd:fundamentals:problem}) builds on related work in data mining; see~\cite{atzmueller2015subgroup, helal2016subgroup, herrera2011overview, ventura2018subgroup} for broad surveys of subgroup discovery in this field.
There are also recent works independent from this research~\cite{asudeh2019assessing, lawless2022interpretable, pastor2021looking, sagadeeva2021sliceline}.
These works are only loosely related to ours since their problem definition differs in the optimization objective, type of subgroup description, and employed constraints.

Nearly all existing subgroup-discovery methods are algorithmic.
In particular, there are heuristic search methods, like PRIM~\cite{friedman1999bump} and Best Interval~\cite{mampaey2012efficient}, as well as exhaustive search methods, like SD-Map~\cite{atzmueller2009fast, atzmueller2006sd}, MergeSD~\cite{grosskreutz2009subgroup}, and BSD~\cite{lemmerich2016fast, lemmerich2010fast}.
To the best of our knowledge, optimizing subgroup discovery with a general-purpose SMT solver is a novel concept.
There are a few other white-box formulations of particular variants of subgroup discovery, which differ from our work in several aspects, as we discuss next.

\subsubsection{White-Box Formulations}
\label{sec:csd:related-work:subgroup-discovery:white-box}

\paragraph{Maximum box problem}

\cite{eckstein2002maximum}~formulates an integer program for the \textsc{Maximum Box} problem, which is about finding a hyperrectangle containing as many positive data objects as possible but no negative data objects, i.e., no false positives.
This problem is an intermediate between subgroup discovery (cf.~Definition~\ref{def:csd:subgroup-discovery}), which allows false positives and false negatives, and perfect-subgroup discovery (cf.~Definition~\ref{def:csd:perfect-subgroup-discovery}), which allows neither.
Also, the problem defines a kind of inverse scenario to minimal-optimal-recall-subgroup discovery (cf.~Definition~\ref{def:csd:minimal-optimal-recall-subgroup-discovery}), which is about finding a subgroup with as \emph{few negative} data objects as possible but \emph{all positive} data objects.
While the latter problem is in~$\mathcal{P}$ (cf.~Proposition~\ref{prop:csd:complexity-unconstrained-minimal-optimal-recall-subgroup-discovery}), \cite{eckstein2002maximum}~proves $\mathcal{NP}$-hardness of the \textsc{Maximum Box} problem by reduction from the \textsc{Maximum Independent Set}~\cite{tarjan1977finding} problem.
In their evaluation, the authors only use a customized branch-and-bound algorithm but neither solver-based nor heuristic search methods.
Further, they consider neither feature-cardinality constraints nor alternative descriptions.

\paragraph{Maximum $\alpha$-pattern problem}

\cite{bonates2008maximum}~investigates the \textsc{Maximum $\alpha$-Pattern} problem.
This problem is similar to the \textsc{Maximum Box} problem but involves a binary dataset and requires a user-selected data object~$\alpha$ to be a subgroup member.
Again, cardinality constraints or alternative descriptions are not considered.
The authors formulate two integer programs as well as two heuristics.
They evaluate their approaches, but no existing subgroup-discovery methods, on generated and benchmark datasets, comparing subgroup quality as well as runtime.
Similar to us, they find that heuristics may reach a subgroup quality similar to a solver-based search with orders of magnitude less runtime.

\paragraph{Box search problem}

\cite{louveaux2014combinatorial}~proposes two integer-programming formulations for the \textsc{Box Search} problem, which is about finding a hyperrectangle that optimizes the sum of the target variable of all contained data objects.
In particular, the target variable is continuous rather than binary, and the chosen objective differs from ours.
Further, there are no constraints on feature cardinality or for alternative descriptions.
In their evaluation, the authors compare solver-based search to multiple versions of a new branch-and-bound approach but not heuristic search methods.
They use synthetically generated datasets rather than typical machine-learning benchmark datasets.

\paragraph{Discriminative itemset mining problem}

Besides three related problems, \cite{koccak2020exploiting}~formulates \textsc{Closed Discriminative Itemset Mining} and \textsc{Relevant Subgroup Discovery} with the constraint-specification language \emph{Essence}~\cite{frisch2008ssence}.
Both these formulations are close to frequent itemset mining, where items correspond to binary features and itemsets to subgroup descriptions.
There is no optimization objective but constraints on the frequency of itemsets and their relevance, excluding dominated solutions.
The authors propose a transformation of their problem specification to enable using standard propositional-satisfiability (SAT) or constraint-programming (CP) solvers.
Their evaluation exclusively analyzes solver runtime rather than subgroup quality and does not compare against heuristic search methods.
Finally, they do not search for alternative descriptions.
Instead of feature-cardinality constraints, they randomly generate costs and values for items and corresponding bounds for itemsets.

\cite{guns2011itemset} also provides a constraint-programming formulation of \textsc{Discriminative Itemset Mining}.
The authors compare two configurations of a constraint solver against existing itemset-mining algorithms.
They evaluate runtime but not quality, as they enumerate all non-redundant itemsets, and do not include constraints for feature cardinality or alternative descriptions.

\subsubsection{Constrained Subgroup Discovery}
\label{sec:csd:related-work:subgroup-discovery:constraints}

Section~\ref{sec:csd:approach:constraint-types} has already discussed various constraint types in subgroup discovery.
Typically, constraints are not formulated declaratively for solver-based optimization but integrated into algorithmic search methods.
\cite{atzmueller2007using}~expresses domain knowledge with the logic programming language Prolog, creating a knowledge base composed of facts and rules.
However, the authors do not use a solver to optimize subgroup discovery.
In the following, we discuss particular related work for the two constraint types we analyze in detail.

\subsubsection{Feature-Cardinality Constraints}
\label{sec:csd:related-work:subgroup-discovery:cardinality}

\paragraph{Formulation}

Feature cardinality is a common constraint type~\cite{meeng2021real} and a well-known metric for subgroup complexity~\cite{helal2016subgroup, herrera2011overview, ventura2018subgroup}.
However, our SMT formulation of this constraint type is novel.
\cite{li2015efficient}~formulates a quadratic program to select non-redundant features for subgroups, but this is only a subroutine within an algorithmic search for subgroups.
Also, the authors define a continuous optimization problem with real-valued feature weights as decision variables, while feature selection within our SMT formulation is discrete (cf.~Equation~\ref{eq:csd:smt-constraint-feature-selection}).

\paragraph{Empirical studies}

While several articles on subgroup discovery use a feature-cardinality constraint in their experiments~\cite{arzamasov2022pedagogical, mampaey2012efficient, lavrac2006relevancy, leeuwen2012diverse, leeuwen2013discovering}, there is a lack of studies that analyze the impact of different feature-cardinality thresholds on different subgroup-discovery methods broadly and systematically.
\cite{friedman1999bump} analyzes the subgroup quality over eliminating a different number of redundant features as a post-processing step, but only for PRIM and one dataset.
\cite{lemmerich2010fast} evaluates different search depths for their algorithm BSD but only regarding runtime and number of subgroups after quality-based pruning, not subgroup quality.
\cite{proencca2022robust} compares multiple search depths for their algorithm SSD++, which returns a list of multiple subgroups, regarding an information-theoretic quality measure.
\cite{helal2016subgroup} compares five subgroup-discovery methods with categorical datasets and also evaluates feature cardinality but does not systematically constrain the latter to different values.
\cite{meeng2021real} evaluates three subgroup-discovery methods, including a beam search and an exhaustive search.
The authors use feature-cardinality constraints with~$k \in \{1, 2, 3, 4\}$ but mainly focus their evaluation on comparing strategies for handling numeric data.
Also, they only use six classification datasets, five of them with at most ten numeric features, while we employ more and higher-dimensional datasets.
Additionally, we compare subgroup discovery with feature-cardinality constraints to an unconstrained setting.

\subsubsection{Alternative Subgroup Descriptions}
\label{sec:csd:related-work:subgroup-discovery:alternatives}

To the best of our knowledge, alternative subgroup descriptions in the sense of this article are a novel concept.
In particular, we aim to maximize the set similarity of contained data objects relative to an original subgroup while using a different subgroup description (cf.~Definition~\ref{def:csd:alternative-subgroup-description-discovery}).
In contrast, there are various existing approaches striving for alternatives in the sense of diverse or non-redundant sets of subgroups, which aim to minimize rather than maximize the overlap of contained data objects~\cite{atzmueller2015subgroup} (cf.~Section~\ref{sec:csd:approach:constraint-types}).
In the following, we discuss approaches that focus on subgroup descriptions.

\paragraph{Description-based diverse subgroup set selection}

\cite{leeuwen2012diverse} introduces six strategies to foster diversity while searching for multiple subgroups simultaneously.
Besides strategies assessing the subgroup members and the information-theoretic compression achieved by subgroups, two strategies refer to subgroup descriptions.
The first excludes subgroup descriptions that have the same quality and differ in only one condition from an existing subgroup description.
The second uses a global upper bound on how often a feature may be selected in a set of subgroup descriptions rather than controlling pairwise dissimilarity.
Both these strategies give users less control over the overlap of subgroup descriptions than our dissimilarity parameter~$\tau$ does.
Further, \cite{leeuwen2012diverse} targets at simultaneous beam search, optimizing subgroup quality and using the diversity strategies only to prune certain solution candidates.
In contrast, we search for alternative descriptions sequentially, optimize similarity to the original subgroup, and consider a solved-based search method in addition to heuristic search.

\paragraph{Diverse top-k characteristics lists}

\cite{lopez2023discovering}~introduces the notion of \emph{diverse top-k characteristic lists}, which is a set of lists, each containing multiple patterns, e.g., subgroups.
Within each list, the subgroups should be alternative to each other in terms of data objects contained.
Between lists, subgroup descriptions should be diverse.
However, the latter goal is implemented with a very simple notion of diversity, i.e., exactly the same subgroup description must not appear in two lists, but any other overlap is allowed.

\paragraph{Equivalent subgroup descriptions of minimal length}

\cite{boley2009non}~introduces the notion of \emph{equivalent subgroup descriptions of minimal length}, which is stricter than our notion of alternative subgroup descriptions.
In particular, the former descriptions need to cover exactly the same set of data objects, like our notion of perfect alternative subgroup descriptions (cf.~Definition~\ref{def:csd:perfect-alternative}), instead of maximizing similarity.
Further, the original feature set should be minimized, i.e., a subset be found, instead of using a different feature set subject to a dissimilarity constraint.
The authors prove $\mathcal{NP}$-hardness and propose two algorithms for their problem but do not pursue a solver-based search.
We adapt their hardness proof to the perfect-subgroup-discovery problem with a feature-cardinality constraint (cf.~Proposition~\ref{prop:csd:complexity-cardinality-np-perfect-subgroup}), based on which we derive further proofs for problems with feature-cardinality constraints and alternative subgroup descriptions.

\paragraph{Redescription mining}

Redescription mining aims to find pairs or sets of descriptions that cover exactly or approximately the same data objects~\cite{galbrun2017redescription, ramakrishnan2004turning}.
Our notion of alternative subgroup descriptions pursues a similar goal.
However, we search for alternative descriptions sequentially instead of simultaneously.
Also, our original subgroup description usually optimizes subgroup quality, while redescription mining has no target variable, i.e., is unsupervised~\cite{ramakrishnan2004turning}.
Further, redescription mining works with different dissimilarity criteria than we do, e.g., having features pre-partitioned into non-overlapping sets~\cite{galbrun2017redescription, gallo2008finding, mihelcic2023complexity} or requiring only one arbitrary part of the description to differ~\cite{parida2005redescription}.
In contrast, we allow users to control the overlap between feature sets with the parameter~$\tau$.
Also, the language for redescriptions may be more complex than for subgroups, e.g., also involve logical negation ($\lnot$) and disjunction ($\lor$)~\cite{galbrun2017redescription, gallo2008finding}, while subgroup descriptions only use the logical AND ($\land$) over features.
Finally, most existing approaches for redescription mining are algorithmic rather than using white-box optimization~\cite{galbrun2017redescription, mihelcic2023complexity}, though~\cite{guns2013kpattern} provides a constraint-programming formulation of this problem and other pattern-set mining problems.
Several formulations and parametrizations of the redescription-mining problem are $\mathcal{NP}$-hard; see \cite{mihelcic2023complexity} for a detailed analysis.

\subsection{Feature Selection}
\label{sec:csd:related-work:feature-selection}

Both constraint types we analyze, i.e., feature-cardinality constraints and alternative subgroup descriptions, relate to the features used in the subgroup description.
In the field of feature selection~\cite{guyon2003introduction, li2017feature}, constraints are a topic as well~\cite{bach2023finding, bach2024alternative, bach2022empirical}.
While limiting feature-set cardinality is very common in feature-selection methods, \cite{bach2023finding, bach2024alternative} are unique since they propose a white-box formulation of alternative feature selection.
Similar to Equation~\ref{eq:csd:smt-constraint-dissimilarity}, they use a threshold-based dissimilarity constraint on feature selection, though with a different dissimilarity measure.
Besides a sequential search for alternatives, which we use as well, they also analyze simultaneous search.

Despite the similarities, traditional feature selection generally tackles a different optimization problem than subgroup discovery.
In particular, the former problem `only' concerns selecting the features instead of also determining bounds on them.
The selected features do not form a prediction model per se but are used in another machine-learning model afterward.
For feature selection itself, a notion of feature-set quality serves as the objective function.
The latter depends on the feature-selection method but typically assesses features globally, while subgroups describe a particular region in the data.

\subsection{Other Fields}
\label{sec:csd:related-work:other}

\paragraph{Alternatives and constraints in data mining}

Working with constraints is an area of research for various sub-fields of data mining \cite{grossi2017survey}, e.g., for clustering \cite{dao2013declarative, dao2024review}, pattern mining \cite{ng1998exploratory, silva2016constrained}, explainable AI~\cite{deutch2019constraints, gorji2022sufficient, mothilal2020explaining, shrotri2022constraint}, and automated machine learning~\cite{neutatz2023automl}.
There is work in the other direction as well, i.e., using machine-learning techniques in constraint solving~\cite{popescu2022overview}.
Finding alternative or diverse solutions is also a concern in various fields, e.g., clustering~\cite{bailey2014alternative}, subspace search~\cite{fouche2021efficient}, and explainable-AI paradigms like counterfactuals~\cite{guidotti2022counterfactual}, criticisms~\cite{kim2016examples}, and semifactuals~\cite{artelt2022even}.

\paragraph{White-box classification}

There are white-box formulations for various types of classification models~\cite{ignatiev2021reasoning}.
E.g., there are formulations in propositional logic (SAT) for optimal decision trees, decision sets, and decision lists~\cite{narodytska2018learning, shati2021sat, yu2021learning}.
Similar to subgroup descriptions, these three model types also use conjunctions of conditions to form decision rules.
Creating sparse models to reduce model complexity, as we do with feature-cardinality constraints, is an issue for such models as well~\cite{yu2021learning}.
However, these model types use multiple rules to classify data globally, while subgroup discovery employs one rule to describe an interesting region.
Further, some of these particular white-box formulations target at perfect predictions rather than optimizing prediction quality.

\paragraph{Counterfactual explanations}

Searching for counterfactual explanations is an explainable-AI paradigm that targets at data objects with feature values as similar as possible to a given data object but with a different prediction of a given classifier~\cite{guidotti2022counterfactual}.
Thus, counterfactuals provide alternative explanations on the local level, i.e., for individual data objects.
In contrast, alternative subgroup descriptions aim to reproduce subgroup membership globally, striving for a similar prediction but a different feature selection.
Approaches yielding multiple counterfactuals often foster diversity, e.g., by extending the optimization objective~\cite{mothilal2020explaining} or introducing constraints~\cite{karimi2020model, mohammadi2021scaling, russell2019efficient}.
However, only some approaches have a user-friendly parameter to actively control the diversity of solutions.
In particular, \cite{mohammadi2021scaling} offers a dissimilarity threshold comparable to our parameter~$\tau$ for alternative subgroup descriptions.

\section{Conclusions and Future Work}
\label{sec:csd:conclusion}

In this section, we recap our article (cf.~Section~\ref{sec:csd:conclusion:conclusion}) and propose directions for future work (cf.~Section~\ref{sec:csd:conclusion:future-work}).

\subsection{Conclusions}
\label{sec:csd:conclusion:conclusion}

Subgroup-discovery methods constitute an important category of interpretable machine-learning models.
In this article, we analyzed constrained subgroup discovery as another step to improve interpretability.
First, we formalized subgroup discovery as an SMT optimization problem.
This formulation supports a variety of user constraints and enables a solver-based search for subgroups.
In particular, we studied two constraint types, i.e., limiting the number of features used in subgroups and searching for alternative subgroup descriptions.
For the latter constraint type, we let users control the number of alternatives and a dissimilarity threshold.
We showed how to integrate these constraint types into our SMT formulation as well as existing heuristic search methods for subgroup discovery.
Further, we proved $\mathcal{NP}$-hardness of the optimization problem with constraints.
Finally, we evaluated heuristic and solver-based search with 27 binary-classification datasets.
In particular, we analyzed four experimental scenarios:
unconstrained subgroup discovery, our two constraint types, and timeouts for solver-based search.
Section~\ref{sec:csd:evaluation:summary} summarized key results.

\subsection{Future Work}
\label{sec:csd:conclusion:future-work}

\paragraph{Datasets}

Our evaluation used over two dozen generic benchmark datasets (cf.~Section~\ref{sec:csd:experimental-design:datasets}).
While such an evaluation shows general trends, the impact of constraints naturally depends on the dataset.
Thus, our results may not transfer to each particular scenario.
This caveat calls for domain-specific case studies.
In such studies, one could also interpret alternative subgroup descriptions qualitatively, i.e., from the domain perspective.

\paragraph{Constraint types}

We formalized, analyzed, and evaluated two constraint types, i.e., feature-cardinality constraints (cf.~Sections~\ref{sec:csd:approach:cardinality} and~\ref{sec:csd:evaluation:cardinality}) and alternative subgroup descriptions (cf.~Sections~\ref{sec:csd:approach:alternatives} and~\ref{sec:csd:evaluation:alternatives}).
While our solver-based approach was not convincing for these two, it retains the conceptual advantage that constraints can be added and combined declaratively instead of needing to integrate individual constraint types into the search procedure.
The two particular constraint types studied in this article were antimonotonic, which eased their integration into heuristic search methods.
As mentioned in Section~\ref{sec:csd:approach:constraint-types}, there are further constraint types one could investigate, e.g., domain-specific constraints, secondary objectives, or alternatives in the sense of covering different data objects rather than covering the same data objects differently.

For alternative subgroup descriptions, one could analyze other dissimilarities, e.g., symmetric ones rather than the asymmetric deselection dissimilarity we used (cf.~Equation~\ref{eq:csd:constraint-dissimilarity}).
While the SMT encoding of subgroup discovery is relatively flexible regarding dissimilarities, integrating them into heuristic search methods may be challenging, e.g., if the dissimilarity is not antimonotonic.

\paragraph{Formalization}

In the solver-based search for subgroups, we used an SMT encoding (cf.~Section~\ref{sec:csd:approach:smt}) and one particular solver.
Different white-box encodings or solvers may speed up the search and lead to fewer timeouts, potentially improving the subgroup quality.
We already proposed MILP and MaxSAT encodings (cf.~Appendices~\ref{sec:csd:appendix:further-encodings:milp} and~\ref{sec:csd:appendix:further-encodings:max-sat}), though without evaluation.

In our article, two assumptions for subgroup discovery were numerical features and a binary target (cf.~Section~\ref{sec:csd:fundamentals:problem}).
One could adapt the SMT encoding to multi-valued categorical features (cf.~Appendix~\ref{sec:csd:appendix:further-encodings:smt-categorical}) and continuous targets.

\paragraph{Time complexity}

We established $\mathcal{NP}$-hardness for subgroup discovery with a feature-cardinality constraint.
In particular, we tackled the search problem for perfect subgroups (cf.~Proposition~\ref{prop:csd:complexity-cardinality-np-perfect-subgroup}) and the general optimization problem (cf.~Proposition~\ref{prop:csd:complexity-cardinality-np}).
While our proof for the latter reduced from the former, one could try to explicitly prove hardness for optimizing imperfect subgroups.
Further, while we presented a polynomial-time algorithm for finding perfect subgroups without constraints (cf.~Proposition~\ref{prop:csd:complexity-unconstrained-perfect-subgroup}), we did not analyze the unconstrained optimization problem of subgroup discovery (cf.~Definition~\ref{def:csd:subgroup-discovery}).

We also showed $\mathcal{NP}$-hardness for finding alternative subgroup descriptions.
Again, we tackled the search problem for perfect alternatives first (cf.~Propositions~\ref{prop:csd:complexity-perfect-alternatives-np-perfect-subgroup} and~\ref{prop:csd:complexity-perfect-alternatives-np-imperfect-subgroup}) before reducing to the general optimization problem later (cf.~Proposition~\ref{prop:csd:complexity-alternatives-np}).
One could try to explicitly prove hardness for optimizing imperfect alternatives.
Also, our proofs focused on scenarios where all originally selected features must not be selected in the alternative subgroup description, i.e., a specific value of the dissimilarity threshold~$\tau$.
One could analyze scenarios with overlapping feature sets explicitly.
Additionally, our proofs assumed a feature-cardinality constraint when searching alternative subgroup descriptions.
One could examine scenarios without this constraint type.

For parameterized complexity, we established membership in the relatively broad complexity class $\mathcal{XP}$ for the unconstrained scenario, feature-cardinality constraints, and alternative subgroup descriptions (cf.~Propositions~\ref{prop:csd:complexity-unconstrained-xp}, \ref{prop:csd:complexity-cardinality-xp}, and~\ref{prop:csd:complexity-alternatives-xp}).
One may attempt to tighten these results.

Finally, while we described how one can integrate feature-cardinality constraints and alternative subgroup descriptions into heuristic search methods (cf.~Sections~\ref{sec:csd:approach:cardinality:heuristics} and~\ref{sec:csd:approach:alternatives:heuristics}), we did not provide quality guarantees relative to the exact optimum.
In that regard, one could seek an approximation complexity result, e.g., membership in the complexity class~$\mathcal{APX}$, as established for the problem of finding equivalent subgroup descriptions of minimal length~\cite{boley2009non}.


\appendix

\section{Appendix}
\label{sec:csd:appendix}

In this section, we provide supplementary materials.
Appendix~\ref{sec:csd:appendix:further-encodings} describes further problem encodings of subgroup discovery, complementing Section~\ref{sec:csd:approach:smt}.
Appendix~\ref{sec:csd:appendix:proofs} contains proofs for propositions from Section~\ref{sec:csd:approach}.
Appendix~\ref{sec:csd:appendix:competitor-runtime} summarizes the runtime experiments with subgroup-discovery packages that led to integrating \emph{BSD} and \emph{SD-Map} into the main experiments (cf.~Section~\ref{sec:csd:evaluation}).

\subsection{Further Problem Encodings of Subgroup Discovery}
\label{sec:csd:appendix:further-encodings}

In this section, we provide additional white-box encodings of subgroup discovery beyond the SMT encoding from Section~\ref{sec:csd:approach:smt}.
First, we describe how to encode categorical features within the SMT formulation (cf.~Section~\ref{sec:csd:appendix:further-encodings:smt-categorical}).
Next, we discuss encodings as a mixed integer linear program (cf.~Section~\ref{sec:csd:appendix:further-encodings:milp}) and a maximum-satisfiability problem (cf.~Section~\ref{sec:csd:appendix:further-encodings:max-sat}).

\subsubsection{Handling Categorical Features in the SMT Encoding}
\label{sec:csd:appendix:further-encodings:smt-categorical}

In general, there are many different options to encode categorical data in machine learning numerically~\cite{matteucci2023benchmark}.
Similarly, there are also multiple options for considering categorical features in an SMT formulation of subgroup discovery.
We present three of them in the following.

\paragraph{Two variables per categorical feature}

As a straightforward option, one may map all categories, i.e., unique values, of each categorical feature to distinct integers before instantiating the optimization problem.
One can directly apply our existing SMT formulation (cf.~Equation~\ref{eq:csd:smt-problem-unconstrained-complete}) to such an ordinally encoded dataset, at least technically.
In particular, there would be two integer-valued bound variables for each encoded categorical feature.
However, the ordering of categories should be semantically meaningful since it influences which categories may jointly be included in the subgroup.
In particular, only sets of categories that form contiguous integer ranges in the ordinal encoding may define subgroup membership.
I.e., the subgroup may comprise the encoded categories~$\{3,4,5\}$, but not only~$\{3,5 \}$ since it needs to include all values between a lower and an upper bound.
Thus, if there is no meaningful ordering of categories, one should choose a different encoding.

\paragraph{Two variables per categorical feature value}

One can achieve more flexibility by introducing separate bound variables for each category of a feature rather than only for each feature.
This approach corresponds to a one-hot encoding of the dataset, which creates one new binary feature for each category.
Thus, the bound variables are effectively binary as well.
By default, our SMT encoding uses a logical AND ($\land$) over the binary features, i.e., categories.
The interpretation of bound values for one binary Feature~$j$ is as follows:

(Case 1) $\mathit{lb}_j = \mathit{ub}_j = 1$ means that data objects that assume the corresponding category for Feature~$j$ are members of the subgroup.
In practice, this case may apply to at most one category of each feature.
Otherwise, the AND ($\land$) operator would require each data object to assume multiple categories for one feature, which is unsatisfiable.
Thus, this encoding cannot directly express that a set of categories is included in the subgroup.

(Case 2) $\mathit{lb}_j = \mathit{ub}_j = 0$ means that data objects that do \emph{not} assume the corresponding category for Feature~$j$ are members of the subgroup.
I.e., data objects assuming the corresponding category are not subgroup members.
Other than Case~1, this case can apply to multiple categories of each feature, i.e., the subgroup may explicitly exclude multiple categories.
Further, if one category is actively included in the subgroup (Case~1), then Case-2 bounds on other categories are redundant since they are implied by the former.

(Case 3) $\mathit{lb}_j = 0,~\mathit{ub}_j = 1$ explicitly deselects a binary feature, i.e., both binary values do not restrict subgroup membership.

(Case 4) $\mathit{lb}_j = 1,~\mathit{ub}_j = 0$ cannot occur since it violates the bound constraints (cf.~Equation~\ref{eq:csd:smt-constraint-bounds-monotonic}).

Finally, note that binary features allow us to slightly simplify the subgroup-membership expression (cf.~Equation~\ref{eq:csd:smt-constraint-subgroup-membership}).
In general, we need to check the lower and upper bound for a feature.
However, if a binary feature assumes the value~0 for a data object, checking the upper bound is unnecessary since it is always satisfied.
Similarly, if a binary feature assumes the value~1 for a data object, checking the lower bound is unnecessary since it is always satisfied.
Both these simplifications assume that the bounds are explicitly defined as binary or at least in $[0, 1]$, which can be enforced with straightforward constraints.
Otherwise, the bounds may theoretically be placed outside the feature's range and exclude all data objects, producing an empty subgroup.

\paragraph{One variable per categorical feature value}

In some scenarios, it does not make sense to include the absence of a category in the subgroup, i.e., to permit $\mathit{lb}_j = \mathit{ub}_j = 0$.
In particular, some existing subgroup-discovery methods for categorical data assume that only the presence of categories is interesting~\cite{atzmueller2015subgroup}.
In this case, introducing one instead of two bound variable(s) for each category suffices.
Assume the categorical Feature~$j$ has $|c_j| \in \mathbb{N}$ different categories~$\{c^1_j, \dots, c^{|c_j|}_j\}$.
Let $\mathit{cb}_j \in \{0, 1\}^{|c_j|}$ denote the corresponding bound variables, which denote whether a category is included in the subgroup.
The ordering of categories in this vector is arbitrary but fixed.

As a difference to previously described encodings, the subgroup-membership expression (cf.~Equation~\ref{eq:csd:smt-constraint-subgroup-membership}) should still use a logical AND ($\land$) over features but not over categories belonging to the same feature.
Otherwise, the expression would be unsatisfiable since each data object only assumes one category for each feature.
Instead, we replace the numeric bound check~$\left( X_{ij} \geq \mathit{lb}_j \right) \land \left( X_{ij} \leq \mathit{ub}_j \right)$ for Feature~$j$ with the following OR ($\lor$) expression:
\begin{equation}
	\bigvee_{l \in \{1, \dots, |c_j|\}} \left( \mathit{cb}^l_j \land \left(  X_{ij} = c^l_j \right) \right)
	\label{eq:csd:category-constraint:or}
\end{equation}
Since the equality holds for exactly one category, all conjunctions except one are false, and the expression simplifies to one variable~$\mathit{cb}^{l'}_j$, where $l'$ is the index of the category~$X_{ij}$.
I.e., for each categorical feature, a data object can only be a subgroup member if the variable belonging to its category is~1.

In general, multiple $\mathit{cb}^l_j$ for Feature~$j$ may be~1, representing multiple categories included in the subgroup, which is an advantage over the previous encoding.
If all categories are in the subgroup, the feature becomes deselected.
Thus, for a categorical Feature~$j$, Equation~\ref{eq:csd:smt-constraint-feature-selection} for feature selection becomes:
\begin{equation}
	s_j \leftrightarrow \lnot \bigwedge_{l \in \{1, \dots, |c_j|\}} \mathit{cb}^l_j
	\label{eq:csd:category-constraint:feature-selection}
\end{equation}
One can also constrain the number of categories in the subgroup, e.g., to either include one category of Feature~$j$ in the subgroup or deselect the feature altogether by including all categories:
\begin{equation}
	 \left( \left( \sum_{l=1}^{|c_j|} \mathit{cb}^l_j \right) = 1 \right) \lor \left( \left( \sum_{l=1}^{|c_j|} \mathit{cb}^l_j \right) = |c_j| \right)
	\label{eq:csd:category-constraint:cardinality}
\end{equation}

\subsubsection{Mixed Integer Linear Programming (MILP)}
\label{sec:csd:appendix:further-encodings:milp}

We start from the SMT formulation and introduce additional variables and constraints to linearize certain logical expressions.

\paragraph{Unconstrained subgroup discovery}

We can keep all decision variables from the corresponding SMT formulation (cf.~Equation~\ref{eq:csd:smt-problem-unconstrained-complete}):
the binary variables~$b_i$ for subgroup membership and the real-valued variables~$\mathit{lb}_j$ and~$\mathit{ub}_j$ for the bounds.
The bound constraints (cf.~Equation~\ref{eq:csd:smt-constraint-bounds-monotonic}) remain unchanged as well.
Further, we retain the optimization objective, which already is linear in~$b_i$ (cf.~Equations~\ref{eq:csd:smt-wracc} and~\ref{eq:csd:smt-constraint-m-as-sum}).
However, we need to linearize the logical AND operators ($\land$) in the definition of subgroup membership~$b_i$ (cf.~Equation~\ref{eq:csd:smt-constraint-subgroup-membership}) by introducing auxiliary variables and further constraints.
In particular, we supplement the variables~$b \in \{0, 1\}^m$ by $b^{\text{lb}} \in \{0, 1\}^{m \times n}$ and $b^{\text{ub}} \in \{0, 1\}^{m \times n}$.
These new binary variables indicate whether a particular data object satisfies the lower or upper bound for a particular feature.
Using linearization techniques for constraint satisfaction and AND operators from~\cite{mosek2022modeling}, we obtain the following set of constraints to replace Equation~\ref{eq:csd:smt-constraint-subgroup-membership}:
\begin{equation}
	\begin{aligned}
		\forall i~\forall j: & & X_{ij} + m_j \cdot b^{\text{lb}}_{ij} &\leq \mathit{lb}_j - \varepsilon_j \\
	 	\forall i~\forall j: & & \mathit{lb}_j &\leq X_{ij} + M_j \cdot \left(1 - b^{\text{lb}}_{ij} \right) \\
	 	\forall i~\forall j: & & \mathit{ub}_j + m_j \cdot b^{\text{ub}}_{ij} &\leq X_{ij} - \varepsilon_j \\
	 	\forall i~\forall j: & & X_{ij} &\leq \mathit{ub}_j + M_j \cdot \left(1 - b^{\text{ub}}_{ij} \right) \\
	 	\forall i~\forall j: & & b_i &\leq b^{\text{lb}}_{ij} \\
	 	\forall i~\forall j: & & b_i &\leq b^{\text{ub}}_{ij} \\
	 	\forall i: & & \sum_{j=1}^{n} \left( b^{\text{lb}}_{ij} + b^{\text{ub}}_{ij} \right) &\leq b_i + 2n - 1 \\
		\text{with indices:} & & i &\in \{1, \dots, m\} \\
		& & j &\in \{1, \dots, n\}
	\end{aligned}
	\label{eq:csd:milp-constraint-subgroup-membership}
\end{equation}
The first two inequalities ensure that $b^{\text{lb}}_{ij} = 1$ if and only if $\mathit{lb}_j \leq X_{ij}$.
The following two inequalities perform a corresponding check for~$b^{\text{ub}}_{ij}$.
The values~$\varepsilon_j \in \mathbb{R}_{> 0}$ are small constants that turn strict inequalities into non-strict inequalities since a MILP solver may only be able to handle the latter.
One possible choice, which we used in a demo implementation, is sorting all unique feature values and taking the minimum difference between two consecutive values in that order.

The values~$M_j \in \mathbb{R}_{> 0}$ and $m_j \in \mathbb{R}_{< 0}$ are large positive and negative constants, respectively.
They allow us to express logical implications between real-valued and binary-valued expressions, compensating the latter's smaller range.
One choice for~$M_j$ is a value larger than the difference between the feature's minimum and maximum, which can be pre-computed before optimization:
\begin{equation}
	\begin{aligned}
		\forall j \in \{1, \dots, n\} & & M_j &:= 2 \cdot \left( \max_{i \in \{1, \dots, m\}} X_{ij} - \min_{i \in \{1, \dots, m\}} X_{ij} \right) \\
		\forall j \in \{1, \dots, n\} & & m_j &:= 2 \cdot \left( \min_{i \in \{1, \dots, m\}} X_{ij} - \max_{i \in \{1, \dots, m\}} X_{ij} \right) \\
	\end{aligned}
	\label{eq:csd:milp-big-m}
\end{equation}
In particular, the difference between the subgroup's bounds and arbitrary feature values must be smaller than $M_j$ and larger than $m_j$, unless the bounds are placed outside the feature's value range.
Since the latter does not improve the subgroup's quality in any case, we prevent it with additional constraints on the bound variables~$\mathit{lb}_j$ and~$\mathit{ub}_j$:
\begin{equation}
	\begin{aligned}
		\forall j \in \{1, \dots, n\} & & \min_{i \in \{1, \dots, m\}} X_{ij} &\leq \mathit{lb}_j &\leq \max_{i \in \{1, \dots, m\}} X_{ij} \\
		\forall j \in \{1, \dots, n\} & & \min_{i \in \{1, \dots, m\}} X_{ij} &\leq \mathit{ub}_j &\leq \max_{i \in \{1, \dots, m\}} X_{ij} \\
	\end{aligned}
	\label{eq:csd:milp-constraint-bounds-in-range}
\end{equation}
Finally, the last three inequalities in Equation~\ref{eq:csd:milp-constraint-subgroup-membership} tie $b^{\text{lb}}_{ij}$ and $b^{\text{ub}}_{ij}$ to $b_i$ and linearize the logical AND operators ($\land$) from Equation~\ref{eq:csd:smt-constraint-subgroup-membership}.
In particular, these constraints ensure that a data object is a member of the subgroup, i.e., $b_i = 1$, if and only if all feature values of the data object observe the lower and upper bounds, i.e., all corresponding $b^{\text{lb}}_{ij} = 1$ and $b^{\text{ub}}_{ij} = 1$.

\paragraph{Feature-cardinality constraints}

The feature-cardinality constraint of the SMT formulation (cf.~Equation~\ref{eq:csd:smt-constraint-feature-cardinalty}) already is a linear expression in the feature-selection variables~$s_j$, so we can keep it as-is.
However, the constraints defining~$s_j$ (cf.~Equation~\ref{eq:csd:smt-constraint-feature-selection}) contain a logical OR ($\lor$) operator and comparison ($<$) expressions.
We linearize these constraints as follows:
\begin{equation}
	\begin{aligned}
		\forall i~\forall j: & & 1 - b^{\text{lb}}_{ij} &\leq s^{\text{lb}}_j \\
		\forall i~\forall j: & & 1 - b^{\text{ub}}_{ij} &\leq s^{\text{ub}}_j \\
		\forall j: & & s^{\text{lb}}_j &\leq s_j \\
		\forall j: & & s^{\text{ub}}_j &\leq s_j \\
		\forall j: & & s_j &\leq 2m - \sum_{i=1}^{m} \left( b^{\text{lb}}_{ij} + b^{\text{ub}}_{ij} \right) \\
		\text{with indices:} & & i &\in \{1, \dots, m\} \\
		& & j &\in \{1, \dots, n\}
	\end{aligned}
	\label{eq:csd:milp-constraint-feature-selection}
\end{equation}
The first four inequalities ensure that a feature is selected, i.e., $s_j = 1$, if any data object's feature value lies outside the subgroup's bounds, i.e., any $b^{\text{lb}}_{ij} = 0$ or $b^{\text{ub}}_{ij} = 0$.
The last inequality covers the other direction of the logical equivalence:
If a feature is selected, then at least one data object's feature value lies outside the subgroup's bounds.

\paragraph{Alternative subgroup descriptions}

The objective function for alternative subgroup descriptions in the SMT formulation (cf.~Equation~\ref{eq:csd:smt-hamming}) is already linear.
We only need to replace the logical negation operators ($\lnot$):
\begin{equation}
	\text{sim}_{\text{nHamm}}(b^{(a)}, b^{(0)}) = \frac{1}{m} \cdot \Big( \sum\limits_{\substack{i \in \{1, \dots, m\} \\ b_i^{(0)} = 1}} b_i^{(a)} + \sum\limits_{\substack{i \in \{1, \dots, m\} \\ b_i^{(0)} = 0}} \left( 1 - b_i^{(a)} \right) \Big)
	\label{eq:csd:mip-hamming}
\end{equation}
The same replacement also applies to the dissimilarity constraints (cf.~Equation~\ref{eq:csd:smt-constraint-dissimilarity}), which now look as follows:
\begin{equation}
	\forall l \in \{0, \dots, a-1\}:~ \text{dis}_{\text{des}}(s^{(a)}, s^{(l)}) = \sum_{\substack{j \in \{1, \dots, n\} \\ s^{(l)}_j = 1}} \left(1 - s^{(a)}_j \right) \geq \min \left( \tau_{\text{abs}},~k^{(l)} \right)
	\label{eq:csd:mip-constraint-dissimilarity}
\end{equation}
Otherwise, this expression is linear as well, so no further auxiliary variables or constraints are necessary.

\paragraph{Implementation}

Our published code (cf.~Section~\ref{sec:csd:experimental-design:implementation}) contains a MILP implementation for unconstrained and feature-cardinality-constrained subgroup discovery.
We use the package \emph{OR-Tools}~\cite{perron2022or-tools} with \emph{SCIP}~\cite{bestuzheva2021scip} as the optimizer.
However, in preliminary experiments, this implementation was (on average) slower than the SMT implementation or yielded worse subgroup quality in the same runtime.
Further, it sometimes finished considerably after the prescribed timeout or ran out of memory after consuming dozens of gigabytes.
Thus, we stuck to the SMT implementation for our main experiments (cf.~Section~\ref{sec:csd:experimental-design:methods}).

\subsubsection{Maximum Satisfiability (MaxSAT)}
\label{sec:csd:appendix:further-encodings:max-sat}

Our SMT formulation of subgroup discovery with and without constraints uses a combination of propositional logic and linear arithmetic.
However, if all feature values are binary or binarized, i.e., $X \in \{0, 1\}^{m \times n}$, we can also define a partial weighted MaxSAT problem \cite{bacchus2021maximum, li2021maxsat}.
This formulation involves hard constraints in propositional logic and an objective function containing weighted clauses, i.e., OR terms.
In our case, it even is a \textsc{Max One}~\cite{khanna1997complete} problem since the `clauses' in the objective are plain binary variables.

\paragraph{Unconstrained subgroup discovery}

For binary feature values, the bound variables $\mathit{lb}_j$ and $\mathit{ub}_j$ become binary rather than real-valued as well.
The subgroup membership variables~$b_i$ were binary already (cf.~Equation~\ref{eq:csd:smt-problem-unconstrained-complete}).
In the hard constraints, all less-or-equal inequalities ($\leq$) become logical implications ($\rightarrow$).
Thus, the bound constraints (cf.~Equation~\ref{eq:csd:smt-constraint-bounds-monotonic}) become:
\begin{equation}
	\forall j \in \{1, \dots, n\}:~ \mathit{lb}_j \rightarrow \mathit{ub}_j
	\label{eq:csd:maxsat-constraint-bounds-monotonic}
\end{equation}
I.e., if the lower bound is~1, then the upper bound also needs to be~1; otherwise, the upper bound may be~0 or~1.

The subgroup-membership expressions (cf.~Equation~\ref{eq:csd:smt-constraint-subgroup-membership}) turn into:
\begin{equation}
	\forall i \in \{1, \dots, m\}:~ b_i\leftrightarrow \bigwedge_{j \in \{1, \dots, n\}} \left( \left( \mathit{lb}_j \rightarrow X_{ij} \right) \land \left( X_{ij} \rightarrow \mathit{ub}_j \right) \right)
	\label{eq:csd:maxsat-constraint-subgroup-membership}
\end{equation}
Since all values~$X_{ij}$ are known, we can remove and simplify terms in the definition of~$b_i$.
In particular, if $X_{ij} = 1$, then $\mathit{lb}_j \rightarrow X_{ij}$ is a tautology, which we can remove, and $X_{ij} \rightarrow \mathit{ub}_j$ becomes $\mathit{ub}_j$.
Vice, versa, if $X_{ij} = 0$, then $X_{ij} \rightarrow \mathit{ub}_j$ is a tautology and $\mathit{lb}_j \rightarrow X_{ij}$ becomes $\lnot \mathit{lb}_j$.

Further, having determined the bound values, the final subgroup description can be expressed as a plain conjunction of propositional literals, e.g., $b_i \leftrightarrow \left( X_{i2} \land \lnot X_{i5} \land X_{i6} \right)$.
In particular, there are four cases:
(1)~If $\mathit{lb}_j = 0$ and $\mathit{ub}_j = 1$, then the feature's value does not restrict subgroup membership and therefore does not need to be checked in the final subgroup description.
(2)~If $\mathit{lb}_j = \mathit{ub}_j = 0$, then only $X_{ij} = 0$ is in the subgroup, i.e., a negative literal becomes part of the final subgroup description.
(3)~If $\mathit{lb}_j = \mathit{ub}_j = 1$, then only $X_{ij} = 1$ is in the subgroup, i.e., a positive literal becomes part of the final subgroup description.
(4)~The combination $\mathit{lb}_j = 1$ and $\mathit{ub}_j = 0$ violates the bound constraints and will therefore not appear in a valid solution.

Finally, the objective function is already a weighted sum of the subgroup-membership variables~$b_i$, which form the soft constraints for the problem.
In particular, we can re-formulate Equation~\ref{eq:csd:smt-wracc} as follows:
\begin{equation}
	\text{WRAcc} = \frac{1}{m} \cdot \sum_{\substack{i \in \{1, \dots, m\} \\ y_i = 1 }} b_i - \frac{m^+}{m^2} \cdot \sum_{i=1}^{m} b_i
	\label{eq:csd:maxsat-wracc}
\end{equation}
Thus, for negative data objects, i.e., with $y_i = 0$, the weight is $-m^+ / m^2$.
For positive data objects, i.e., with $y_i = 1$, the weight is $(m - m^+) / m^2$.
Since~$m$ is a constant, we can also multiply with~$m^2$ to obtain integer-valued weights.

\paragraph{Feature-cardinality constraints}

For binary features, the definition of the feature selection variables~$s_j$ (cf.~Equation~\ref{eq:csd:smt-constraint-feature-selection}), which are binary by default, amounts to:
\begin{equation}
	\begin{aligned}
		\forall j: & & s^{\text{lb}}_j &\leftrightarrow \left( \mathit{lb}_j \land \lnot \left( \bigwedge_{i \in \{1, \dots, m\}} X_{ij} \right) \right) \\
		\forall j: & &s^{\text{ub}}_j &\leftrightarrow \left( \lnot \mathit{ub}_j \land \left( \bigvee_{i \in \{1, \dots, m\}} X_{ij} \right) \right) \\
		\forall j: & & s_j &\leftrightarrow \left( s^{\text{lb}}_j \lor s^{\text{ub}}_j \right) \\
		\text{with index:} & & j &\in \{1, \dots, n\}
	\end{aligned}
	\label{eq:csd:maxsat-constraint-feature-selection}
\end{equation}
I.e., a feature is selected regarding its lower bound if the lower bound is set to~1 and at least one feature value is~0, i.e., at least one feature value is excluded from the subgroup.
Vice versa, a feature is selected regarding its upper bound if the upper bound is set to~0 and at least one feature value is~1, i.e., at least one feature value is excluded from the subgroup.
Since all values~$X_{ij}$ are known, we can evaluate the corresponding AND and OR terms before optimization.
If a feature is~0 and~1 for at least one data object each, which should usually be the case, Equation~\ref{eq:csd:maxsat-constraint-feature-selection} becomes a much simpler expression:
\begin{equation}
		s_j \leftrightarrow \left( \mathit{lb}_j\lor \lnot \mathit{ub}_j \right)
	\label{eq:csd:maxsat-constraint-feature-selections-simplified}
\end{equation}

To transform the actual feature-cardinality constraint (cf.~Equation~\ref{eq:csd:smt-constraint-feature-cardinalty}), which sums up the variables~$s_j$ and compares them to a user-defined~$k$, into propositional logic, we can use a cardinality encoding from the literature~\cite{sinz2005towards}.

\paragraph{Alternative subgroup descriptions}

The objective function for alternative subgroup descriptions (cf.~Equation~\ref{eq:csd:smt-hamming}) already is a weighted sum of the subgroup-membership variables~$b_i^{(a)}$.
In particular, for negative data objects, i.e., with $y_i = 0$, the weight of the literal $\lnot b_i^{(a)}$ is~$1 / m$.
For positive data objects, i.e., with $y_i = 1$, the weight of the literal $b_i^{(a)}$ is~$1 / m$.
Since~$m$ is a constant, we can also use~1 as the weight.

We can encode the dissimilarity constraint on the feature selection (cf.~Equation~\ref{eq:csd:smt-constraint-dissimilarity}) with a cardinality encoding from the literature~\cite{sinz2005towards}.

\paragraph{Non-binary features}

While we discussed binary features up to now, we can also encode multi-valued features in a way suitable for a MaxSAT formulation.
In Section~\ref{sec:csd:appendix:further-encodings:smt-categorical}, we already addressed how categorical features may be represented binarily.
For numeric features, we can introduce two binary variables for each numeric value:
Let the numeric Feature~$j$ have $|v_j| \in \mathbb{N}$ distinct values~$\{v^1_j, \dots, v^{|v_j|}_j\}$, with higher superscripts denoting higher values.
Next, let $\mathit{lb}_j \in \{0, 1\}^{|v_j|}$ and $\mathit{ub}_j \in \{0, 1\}^{|v_j|}$ denote the corresponding binary bound variables.
I.e., instead of two bound variables per feature, there are two bound variables for each unique feature value now.
$\mathit{lb}^l_j$ indicates whether the $l$-th unique value of Feature~$j$ is the lower bound.
Vice versa, $\mathit{ub}^l_j$ indicates whether the $l$-th unique value of Feature~$j$ is the upper bound.
If this encoding generates too many variables, one may discretize the feature first, e.g., by binning its values and representing each bin by one value, e.g., the bin's mean.

The bound constraints (cf.~Equations~\ref{eq:csd:smt-constraint-bounds-monotonic} and~\ref{eq:csd:maxsat-constraint-bounds-monotonic}) take the following form:
\begin{equation}
	\begin{aligned}
		\forall j: & & \sum_{l=1}^{|v_j|} \mathit{lb}^l_j &= 1 \\
		\forall j: & & \sum_{l=1}^{|v_j|} \mathit{ub}^l_j &= 1 \\
		\forall j~ \forall l_1 \in \{1, \dots, |v_j|\}: & & \mathit{ub}^{l_1}_j &\rightarrow \bigvee_{l_2 \in \{1, \dots, l_1\}} \mathit{lb}^{l_2}_j \\
		\text{with index:} & & j &\in \{1, \dots, n\}
	\end{aligned}
	\label{eq:csd:maxsat-numeric-constraint-bounds-monotonic}
\end{equation}
The first two constraints ensure that exactly one value of Feature~$j$ is chosen as the lower bound and upper bound, respectively.
These constraints can be encoded into propositional logic with a cardinality encoding from the literature~\cite{sinz2005towards}.
The third constraint enforces that the value chosen as the lower bound is less than or equal to the value chosen as the upper bound.
Alternatively, one could also formulate that the value chosen as the upper bound is greater than or equal to the value chosen as the lower bound.

We formulate the subgroup-membership expressions (cf.~Equations~\ref{eq:csd:smt-constraint-subgroup-membership} and~\ref{eq:csd:maxsat-constraint-subgroup-membership}) as follows:
\begin{equation}
	\forall i \in \{1, \dots, m\}:~ b_i\leftrightarrow \bigwedge_{j \in \{1, \dots, n\}} \Big( \Big( \bigvee_{\substack{l \in \{1, \dots, \bar{l}\} \\ X_{ij} = v_{\bar{l}} }} \mathit{lb}^l_j \Big) \land \Big( \bigvee_{\substack{l \in \{\bar{l}, \dots, |v_j|\} \\ X_{ij} = v_{\bar{l}} }} \mathit{ub}^l_j \Big) \Big)
	\label{eq:csd:maxsat-numeric-constraint-subgroup-membership}
\end{equation}
In particular, for a data object to be a subgroup member, each feature's lower bound needs to be lower or equal to the actual value~$X_{ij}$, while the upper bound needs to be higher or equal.
For the binary lower-bound variables $\mathit{lb}^l_j$, this means that any of the bound variables representing values lower or equal to $X_{ij}$ needs to be~1; vice versa for the upper bounds.

Finally, for feature-cardinality constraints, we define the feature-selection variables~$s_j$ (cf.~Equations~\ref{eq:csd:smt-constraint-feature-selection} and~\ref{eq:csd:maxsat-constraint-feature-selection}) as follows:
\begin{equation}
	\forall j \in \{1, \dots, n\}:~ s_j \leftrightarrow \left( \lnot \mathit{lb}^1_j \lor \lnot \mathit{ub}^{|v_j|}_j \right)
	\label{eq:csd:maxsat-numeric-constraint-feature-selection}
\end{equation}
In particular, we check whether the lower bound is not the minimum or the upper bound is not the maximum value of that feature, which indicates whether the bounds exclude at least one data object from the subgroup or not.
The actual feature-cardinality constraint (cf.~Equation~\ref{eq:csd:smt-constraint-feature-cardinalty}) does not need to be specifically adapted for non-binary features in MaxSAT.
The same goes for the definition of alternative subgroup descriptions (cf.~Equations~\ref{eq:csd:smt-hamming} and Equation~\ref{eq:csd:smt-constraint-dissimilarity}), which only uses the original binary decision variables by default.

\subsection{Proofs}
\label{sec:csd:appendix:proofs}

In this section, we provide proofs for propositions from Section~\ref{sec:csd:approach}, particularly for the complexity results for subgroup discovery with a feature-cardinality constraint and for searching alternative subgroup descriptions.

\subsubsection{Proof of Proposition~\ref{prop:csd:complexity-cardinality-np-perfect-subgroup}}
\label{sec:csd:appendix:proofs:complexity-cardinality-np-perfect-subgroup}

\begin{proof}
Let an arbitrary problem instance~$I$ of the decision problem \textsc{Set Covering}~\cite{karp1972reducibility} be given.
$I$ consists of a set of elements~$E = \{e_1, \dots, e_m\}$, a set of sets~$\mathbb{S} = \{S_1,  \dots, S_n\}$ with $E = \bigcup_{S \in \mathbb{S}} S$, and a cardinality~$k \in \mathbb{N}$.
The decision problem \textsc{Set Covering} asks whether a subset $\mathbb{C} \subseteq \mathbb{S}$ exists with $|\mathbb{C}| \leq k$ and $E = \bigcup_{S \in \mathbb{C}} S$, i.e., a subset of~$\mathbb{S}$ which contains (= covers) each element from~$E$ in at least one set and consist of at most $k$~sets.

We transform~$I$ into a problem instance~$I'$ of the perfect-subgroup-discovery problem (cf.~Definition~\ref{def:csd:perfect-subgroup-discovery}) with a feature-cardinality constraint (cf.~Definition~\ref{def:csd:feature-cardinality-constraint}).
To this end, we define a binary dataset~$X \in \{0, 1\}^{(m + 1) \times n}$, prediction target~$y \in \{0, 1\}^{m+1}$, and retain the number of sets~$k \in \mathbb{N}$ as feature cardinality~$k$.
In particular, data objects represent elements from~$E$, and features represent sets from~$\mathbb{S}$.
I.e., $X_{ij}$ denotes $e_i \in S_j$, i.e., membership of Element~$i$ in Set~$j$.
The additional index $i = m + 1$ represents a \emph{dummy element} that is not part of any set, so all its feature values~$X_{ij}$ are set to~0.
Further, we define the prediction target~$y \in \{0, 1\}^{m+1}$ as $y_{m+1} = 1$ and $y_i = 0$ for all other indices $i \in \{1, \dots, m\}$.
This prediction target represents whether an element should \emph{not} be covered by the set of sets~$\mathbb{C} \subseteq \mathbb{S}$.
In particular, all actual elements from~$E$ should be covered but not the new dummy element.
This `inverted' definition of the prediction target stems from the different nature of set covers and subgroup descriptions:
Set covers include elements from selected sets, with the empty cover $\mathbb{C} = \emptyset$ containing no elements.
There is a logical OR ($\lor$) respectively set union over the selected sets.
In contrast, subgroup descriptions exclude data objects based on bounds for their selected features, with the unrestricted subgroup containing all data objects.
There is a logical AND ($\land$) over the features' bounds.

A perfect subgroup (cf.~Definition~\ref{def:csd:perfect-subgroup}) exactly replicates the prediction target as subgroup membership.
Here, it only contains the data object representing the dummy element but no data objects representing actual elements.
Further, as all feature values of this dummy data object are~0, the subgroup description only consists of the bounds $\mathit{lb}_j = \mathit{ub}_j = 0$ for selected features and $\mathit{lb}_j = 0 < 1 = \mathit{ub}_j$ for unselected features.
Therefore, the data objects described by the selected features represent elements not contained in any of the selected sets, which only applies to the dummy element.
Vice versa, all remaining data objects represent elements that are part of at least one selected set, which applies to all actual elements from~$E$.
Further, the feature-cardinality constraint (cf.~Definition~\ref{def:csd:feature-cardinality-constraint}) ensures that at most $k$~features are selected, which means that at most $k$~sets are selected.
Thus, if the feature-cardinality constraint is satisfied in the perfect subgroup, the selected features represent sets forming a valid set cover~$\mathbb{C}$.

In contrast, if no feature set of the desired size~$k$ can describe a perfect subgroup, then at least one data object with prediction target~$y_i = 0$ has to be part of the subgroup.
Thus, at least one element is not contained in any set forming the set cover, so no valid set cover of size~$k$ exists.

Overall, a solution to the instance~$I'$ of the perfect-subgroup discovery problem (cf.~Definition~\ref{def:csd:perfect-subgroup-discovery}) with a feature-cardinality constraint (cf.~Definition~\ref{def:csd:feature-cardinality-constraint}) also solves the instance~$I$ of the decision problem \textsc{Set Covering} ~\cite{karp1972reducibility} with negligible computational overhead.
In particular, an efficient solution algorithm for the former would also efficiently solve the latter.
However, since the latter problem is $\mathcal{NP}$-hard~\cite{karp1972reducibility}, the former is as well.
To be more precise, the perfect-subgroup-discovery problem with a feature-cardinality constraint resides in the complexity class $\mathcal{NP}$ and therefore is $\mathcal{NP}$-complete.
In particular, checking a solution induces a polynomial cost of~$O(m \cdot n)$, requiring one pass over the dataset to determine subgroup membership and feature selection.
\end{proof}
This proof is an adaptation of the proof of \cite{boley2009non} for minimizing the feature cardinality of a given subgroup description.
The latter proof reduces from the optimization problem \textsc{Minimum Set Cover}, while we use the decision problem \textsc{Set Covering} since perfect-subgroup discovery (cf.~Definition~\ref{def:csd:perfect-subgroup-discovery}) is not an optimization problem.
Further, we replace the notion of a given subgroup description~\cite{boley2009non} with the notion of a perfect subgroup.
Also, we employ inequalities with lower and upper bounds in the subgroup description, while \cite{boley2009non}~uses `feature=value' conditions.
However, this difference is irrelevant for binary datasets, where selected features have $\mathit{lb}_j = \mathit{ub}_j$ bounds and thereby implicitly select a feature value instead of a range.
The hardness result naturally extends to real-valued datasets, which generalize binary datasets.

Note that the hardness reduction does not work for the special case $k=n$.
For \textsc{Set Covering}, this case allows all sets to be selected, which leads to a trivial solution since the complete set of sets~$\mathbb{S}$ contains all elements from~$E$ by definition.
Vice versa, being able to use all features in the subgroup description leads to the unconstrained problem of perfect-subgroup discovery (cf.~Definition~\ref{def:csd:perfect-subgroup-discovery}), which admits a polynomial-time solution (cf.~Proposition~\ref{prop:csd:complexity-unconstrained-perfect-subgroup}).

\subsubsection{Proof of Proposition~\ref{prop:csd:complexity-cardinality-np}}
\label{sec:csd:appendix:proofs:complexity-cardinality-np}

\begin{proof}
Let an arbitrary problem instance~$I$ of the perfect-subgroup-discovery problem (cf.~Definition~\ref{def:csd:perfect-subgroup-discovery}) with a feature-cardinality constraint (cf.~Definition~\ref{def:csd:feature-cardinality-constraint}) be given.
We transform~$I$ into a problem instance~$I'$ of the subgroup-discovery problem (cf.~Definition~\ref{def:csd:subgroup-discovery}) with the same constraint.
In particular, we define the objective as optimizing a subgroup-quality function~$Q(\mathit{lb}, \mathit{ub}, X, y)$ rather than searching for a perfect subgroup (cf.~Definition~\ref{def:csd:perfect-subgroup}) that may or may not exist.
The other inputs of the problem instance ($X$, $y$, and $k$) remain the same.

Based on the assumption we made on~$Q(\mathit{lb}, \mathit{ub}, X, y)$ in Proposition~\ref{prop:csd:complexity-cardinality-np}, the optimal solution for~$I'$ is a perfect subgroup if the latter exists.
Thus, if the optimal subgroup for~$I'$ is not perfect, then a perfect subgroup does not exist at all.
Checking whether a subgroup is perfect entails a cost of $O(m \cdot n)$, i.e., computing subgroup membership and checking for false positives and false negatives.
Overall, an algorithm for subgroup discovery (cf.~Definition~\ref{def:csd:subgroup-discovery}) with a feature-cardinality constraint (cf.~Definition~\ref{def:csd:feature-cardinality-constraint}) solves perfect-subgroup discovery (cf.~Definition~\ref{def:csd:perfect-subgroup-discovery}) with the same constraint with negligible overhead.
Since the latter problem is $\mathcal{NP}$-complete (cf.~Proposition~\ref{prop:csd:complexity-cardinality-np-perfect-subgroup}) and the former resides in the complexity class $\mathcal{NP}$, the former is $\mathcal{NP}$-complete as well.
\end{proof}
As an alternative proof, one could reduce from the optimization problem \textsc{Maximum Coverage}~\cite{chekuri2004maximum} instead of the search problem of perfect-subgroup discovery (cf.~Definition~\ref{def:csd:perfect-subgroup-discovery}) with a feature-cardinality constraint (cf.~Definition~\ref{def:csd:feature-cardinality-constraint}).
This proof idea is strongly related to the proof for Proposition~\ref{prop:csd:complexity-cardinality-np-perfect-subgroup} (cf.~Section~\ref{sec:csd:appendix:proofs:complexity-cardinality-np-perfect-subgroup}), which reduces from the decision problem \textsc{Set Covering}~\cite{karp1972reducibility} to perfect-subgroup discovery (cf.~Definition~\ref{def:csd:perfect-subgroup-discovery}) with a feature-cardinality constraint (cf.~Definition~\ref{def:csd:feature-cardinality-constraint}).
In contrast to \textsc{Set Covering}, the $k \in \mathbb{N}$~selected subsets in \textsc{Maximum Coverage} need not cover all elements but should cover as many elements as possible.
In the terminology of subgroup discovery, the latter objective corresponds to a particular notion of subgroup quality:
maximizing the number of true negatives or minimizing the number of false positives, i.e., excluding as many negative data objects from the subgroup as possible.
We introduced this problem as minimal-optimal-recall-subgroup discovery (cf.~Definition~\ref{def:csd:minimal-optimal-recall-subgroup-discovery}), which resides in~$\mathcal{P}$ without a feature-cardinality constraint (cf.~Proposition~\ref{prop:csd:complexity-unconstrained-minimal-optimal-recall-subgroup-discovery}) due to the baseline \emph{MORS} (cf.~Algorithm~\ref{al:csd:mors}).
When equipping \emph{MORS} with feature-cardinality constraints (cf.~Section~\ref{sec:csd:approach:cardinality:baselines}), existing heuristics for the \textsc{Maximum Coverage} problem may provide approximation guarantees.

However, minimizing the number of false positives is a simpler objective than WRAcc (cf.~Equation~\ref{eq:csd:wracc}), which we focus on in this article.
Our proof approach chosen above is more general regarding the notion of subgroup quality but more narrow in the sense that it reduces from a search problem, assuming a particular value of the objective function, instead of an optimization problem.

\subsubsection{Proof of Proposition~\ref{prop:csd:complexity-perfect-alternatives-np-perfect-subgroup}}
\label{sec:csd:appendix:proofs:complexity-perfect-alternatives-np-perfect-subgroup}

\begin{proof}
Let an arbitrary problem instance~$I$ of the perfect-subgroup-discovery problem (cf.~Definition~\ref{def:csd:perfect-subgroup-discovery}) with a feature-cardinality constraint (cf.~Definition~\ref{def:csd:feature-cardinality-constraint}) be given.	
We transform~$I$ into a problem instance~$I'$ of the perfect-alternative-subgroup-description-discovery problem (cf.~Definition~\ref{def:csd:perfect-alternative-subgroup-description-discovery}) with the same constraint.
In particular, we retain the feature-cardinality threshold~$k \in \mathbb{N}$.
We will slightly modify the dataset~$X \in \mathbb{R}^{m \times n}$, as explained later.

Based on the assumptions we made in Proposition~\ref{prop:csd:complexity-perfect-alternatives-np-perfect-subgroup}, we define the original subgroup for~$I'$ to be perfect (cf.~Definition~\ref{def:csd:perfect-subgroup}), i.e., having subgroup membership $b^{(0)} = y$.
Also, we choose the dissimilarity threshold~$\tau \in \mathbb{R}_{\geq 0}$ high enough that the alternative subgroup description may not select any features that were selected in the original subgroup description.
This choice of~$\tau$ depends on the choice of the dissimilarity measure~$\text{dis}(\cdot)$ for feature-selection vectors.
E.g., we can choose the deselection dissimilarity used in our article (cf.~Equation~\ref{eq:csd:constraint-dissimilarity}) and $\tau_{\text{abs}} = k$.
Note that we do not even need to explicitly define the actual feature selection for the original subgroup description since we must not select these features in the alternative subgroup description anyway.
For the sake of completeness, we can define dataset~$X' \in \mathbb{R}^{m \times (n+k)}$ of problem instance~$I'$ as dataset~$X \in \mathbb{R}^{m \times n}$ of problem instance~$I$ with $k$~extra \emph{perfect features} added.
In particular, we define the Features $n+1, \dots, n+k$ to be identical to the binary prediction target~$y$.
Choosing the bounds $\mathit{lb}_j = \mathit{ub}_j = 1$ on any of these extra features produces the desired original subgroup membership $b^{(0)} = y$.
We further assume that all extra features were selected in the original subgroup description but none of the actual features from~$X$ was, i.e., $\forall j \in \{1, \dots, n\}:~ s^{(0)}_j = 0$ and $\forall j \in \{n+1, \dots, n+k\}:~ s^{(0)}_j = 1$.

A solution for problem instance~$I'$ is also a solution for problem instance~$I$.
In particular, the perfect alternative subgroup description (cf.~Definition~\ref{def:csd:perfect-alternative}) defines a perfect subgroup since it perfectly replicates the original subgroup membership, which constitutes a perfect subgroup.
I.e., $b^{(a)} = b^{(0)} = y$.
Due to the dissimilarity constraint, the alternative subgroup description only selects features from dataset~$X$, not those newly added to create~$X'$.
Finally, both~$I$ and~$I'$ use a feature-cardinality constraint with threshold~$k$.
Thus, if a perfect alternative subgroup description for~$I'$ exists, it also solves~$I$.
If it does not exist, then there is also no other perfect subgroup for~$I$.

Thus, an efficient solution algorithm for the perfect-alternative-subgroup-description-discovery problem (cf.~Definition~\ref{def:csd:perfect-alternative-subgroup-description-discovery}) with a feature-cardinality constraint (cf.~Definition~\ref{def:csd:feature-cardinality-constraint}) would also efficiently solve perfect-subgroup discovery (cf.~Definition~\ref{def:csd:perfect-subgroup-discovery}) with the same constraint.
However, we already established that the latter problem is $\mathcal{NP}$-complete (cf.~Proposition~\ref{prop:csd:complexity-cardinality-np-perfect-subgroup}).
Further, evaluating a solution for the former problem entails a polynomial cost of $O(m \cdot n)$ for checking subgroup membership, the bound constraints, the feature-cardinality constraint, and the dissimilarity constraint, placing the problem in complexity class~$\mathcal{NP}$.
Thus, perfect-alternative-subgroup-description discovery (cf.~Definition~\ref{def:csd:perfect-alternative-subgroup-description-discovery}) with a feature-cardinality constraint (cf.~Definition~\ref{def:csd:feature-cardinality-constraint}) is $\mathcal{NP}$-complete.
\end{proof}

\subsubsection{Proof of Proposition~\ref{prop:csd:complexity-perfect-alternatives-np-imperfect-subgroup}}
\label{sec:csd:appendix:proofs:complexity-perfect-alternatives-np-imperfect-subgroup}

\begin{proof}
Let an arbitrary problem instance~$I$ of the perfect-alternative-subgroup-description-discovery problem (cf.~Definition~\ref{def:csd:perfect-alternative-subgroup-description-discovery}) with a feature-cardinality constraint (cf.~Definition~\ref{def:csd:feature-cardinality-constraint}) and a perfect original subgroup (cf.~Definition~\ref{def:csd:perfect-subgroup}) be given.
We transform~$I$ into a problem instance~$I'$ of the same problem but with an imperfect original subgroup.
In particular, we retain all inputs of the problem as-is except defining dataset~$X' \in \mathbb{R}^{(m + 1) \times n}$ of problem instance~$I'$ as dataset~$X \in \mathbb{R}^{m \times n}$ of problem instance~$I$ plus an additional \emph{imperfect data object}.
This special data object has the label $y_{m+1}=0$ but exactly the same feature values as an arbitrary existing data object~$X_{i\cdot}$ with $y_i=1$.
In particular, such a data object makes it impossible to find a perfect subgroup.
However, we assume this data object to be a member of the original subgroup, i.e., $b^{(0)}_{m+1} = 1$, while subgroup membership of all other data objects corresponds to their prediction target, i.e., $\forall i \in \{1, \dots, m\}:~ b^{(0)}_i = y_i$.

If there is a solution for problem instance~$I'$, we can easily transform it to a solution for~$I$.
In particular, since the solution is a perfect alternative subgroup description (cf.~Definition~\ref{def:csd:perfect-alternative}), it replicates~$b^{(0)}$, i.e., assigns all positive data objects of~$I$ to the alternative subgroup and places all negative data objects of~$I$ outside the subgroup.
The additional imperfect data object is also a member of the alternative subgroup in~$I'$ but does not exist in~$I$.
Thus, the solution is a perfect subgroup for~$I$.
On the other hand, if no solution for problem instance~$I'$ exists, then there is also no solution for~$I$.

Overall, an efficient solution algorithm for the problem of perfect-alternative-subgroup-description discovery (cf.~Definition~\ref{def:csd:perfect-alternative-subgroup-description-discovery}) with a feature-cardinality constraint (cf.~Definition~\ref{def:csd:feature-cardinality-constraint}) and an imperfect original subgroup (cf.~Definition~\ref{def:csd:perfect-subgroup}) could also be used to efficiently solve this problem for a perfect original subgroup.
However, we proved that the latter problem is $\mathcal{NP}$-complete (cf.~Proposition~\ref{prop:csd:complexity-perfect-alternatives-np-perfect-subgroup}), making the former, which resides in $\mathcal{NP}$ as well, also $\mathcal{NP}$-complete.
\end{proof}

\subsubsection{Proof of Proposition~\ref{prop:csd:complexity-alternatives-np}}
\label{sec:csd:appendix:proofs:complexity-alternatives-np}

The following proof is similar to the proof of Proposition~\ref{prop:csd:complexity-cardinality-np}
(cf.~Section~\ref{sec:csd:appendix:proofs:complexity-cardinality-np}), which reduced the search problem of perfect-subgroup discovery with a feature-cardinality constraint (cf.~Definitions~\ref{def:csd:perfect-subgroup-discovery} and ~\ref{def:csd:feature-cardinality-constraint}) to the optimization problem of subgroup discovery (cf.~Definition~\ref{def:csd:subgroup-discovery}) with the same constraint.
\begin{proof}
Let an arbitrary problem instance~$I$ of the perfect-alternative-subgroup-description-discovery problem (cf.~Definition~\ref{def:csd:perfect-alternative-subgroup-description-discovery}) with a feature-cardinality constraint (cf.~Definition~\ref{def:csd:feature-cardinality-constraint}) be given.
We transform~$I$ into a problem instance~$I'$ of the alternative-subgroup-description-discovery problem (cf.~Definition~\ref{def:csd:alternative-subgroup-description-discovery}) with the same constraint.
In particular, we define the objective as optimizing the subgroup-membership similarity~$\text{sim}(\cdot)$ rather than asking for a perfect alternative subgroup description (cf.~Definition~\ref{def:csd:perfect-alternative}) that may or may not exist.
The other inputs of the problem instance remain the same.

Based on the assumption we made on~$\text{sim}(\cdot)$ in Proposition~\ref{prop:csd:complexity-alternatives-np}, the optimal solution for~$I'$ is a perfect alternative subgroup description if the latter exists.
Thus, if the optimal alternative subgroup description for~$I'$ is not a perfect alternative, then a perfect alternative subgroup description does not exist.
Overall, an algorithm for alternative-subgroup-description discovery (cf.~Definition~\ref{def:csd:alternative-subgroup-description-discovery}) with a feature-cardinality constraint (cf.~Definition~\ref{def:csd:feature-cardinality-constraint}) solves perfect-alternative-subgroup-description discovery (cf.~Definition~\ref{def:csd:perfect-alternative-subgroup-description-discovery}) with the same constraint with negligible overhead.
Since the latter problem is $\mathcal{NP}$-complete (cf.~Propositions~\ref{prop:csd:complexity-perfect-alternatives-np-perfect-subgroup} and~\ref{prop:csd:complexity-perfect-alternatives-np-imperfect-subgroup}) and the former resides in~$\mathcal{NP}$, the former is $\mathcal{NP}$-complete.
\end{proof}

\subsection{Competitor-Runtime Experiments}
\label{sec:csd:appendix:competitor-runtime}

In this section, we describe the runtime experiments we conducted to find exhaustive subgroup-discovery methods as competitors for our solver-based approach~\emph{SMT} in the main experiments (cf.~Section~\ref{sec:csd:experimental-design:methods}).
Section~\ref{sec:csd:appendix:competitor-runtime:experimental-design} explains the experimental design, and Section~\ref{sec:csd:appendix:competitor-runtime:evaluation} presents the results.

\subsubsection{Experimental Design}
\label{sec:csd:appendix:competitor-runtime:experimental-design}

\paragraph{Goal}

The competitor-runtime experiments aim to find subgroup-discovery methods that are suitable competitors for solver-based search in our main experimental pipeline.
An ideal competitor should satisfy several criteria to enable a fair comparison:
\begin{enumerate}[noitemsep, label=(\arabic*)]
	\item Support optimizing WRAcc~\cite{lavravc1999rule} as the subgroup-quality metric.
	\item Support setting a feature-cardinality threshold.
	\item Not have additional constraints, e.g., on the number of subgroup members.
	\item Support a continuous search space where subgroup descriptions define intervals on features (cf.~Definition~\ref{def:csd:subgroup}).
	\item Be reasonably fast.
\end{enumerate}
Note that these criteria do not require the competitors to yield a certain subgroup quality, which we would study in the main experiments anyway.
Instead, (1)-(4) ensure that the competitors address the same optimization problem as we do, and (5) concerns the experiments' feasibility.

\begin{table}[t]
	\centering
	\caption{
		Datasets used in our competitor-runtime experiments.
		$m$~denotes the number of data objects and $n$~the number of features.
	}
	\begin{tabular}{lrr}
		\toprule
		Dataset & $m$ & $n$ \\
		\midrule
		backache & 180 & 32 \\
		horse\_colic & 368 & 22 \\
		ionosphere & 351 & 34 \\
		iris & 150 & 4 \\
		spect & 267 & 22 \\
		spectf & 349 & 44 \\
		\bottomrule
	\end{tabular}
	\label{tab:csd:competitor-runtime-datasets}
\end{table}

\paragraph{Subgroup-discovery methods}

We consider competitors from four Python packages for subgroup discovery, i.e., \emph{pysubdisc}, \emph{pysubgroup}~\cite{lemmerich2019pysubgroup}, \emph{sd4py}~\cite{hudson2023subgroup}, and \emph{subgroups}~\cite{lopez2024subgroups}.
All four packages offer exhaustive subgroup-discovery methods that support WRAcc as the optimization objective.
Such methods are particularly interesting competitors for our solver-based search method \emph{SMT}.
Additionally, \emph{pysubdisc}, \emph{pysubgroup}, and \emph{sd4py} provide beam search as a heuristic search method.
For comparison, we also include \emph{Beam} and \emph{SMT} from our main experiments (cf.~Section~\ref{sec:csd:experimental-design:methods}).
Overall, we evaluate the following 17 subgroup-discovery methods in our competitor-runtime experiments:
\begin{itemize}[noitemsep]
	\item \emph{csd} (our package): \emph{Beam}, \emph{SMT}
	\item \emph{pysubdisc}: \emph{Beam}, \emph{BestFirst}, \emph{BreadthFirst}, \emph{DepthFirst}
	\item \emph{pysubgroup}: \emph{Apriori}, \emph{Beam}, \emph{DepthFirst}, \emph{GPGrowth}
	\item \emph{sd4py}: \emph{Beam}, \emph{BSD}, \emph{SDMap}
	\item \emph{subgroups}: \emph{BSD}, \emph{SDMap}, \emph{SDMapStar}, \emph{VLSD}
\end{itemize}
Note that we harmonized the method names between the packages, i.e., the actual function names in each package may differ slightly.

The methods from \emph{sd4py} and \emph{subgroups} require discretizing numeric features.
To this end, we create up to 50 bins per feature, with fewer bins if a feature has fewer unique values.
\emph{sd4py} has a built-in equal-width discretization, while we manually employ equal-frequency discretization for \emph{subgroups}.

The exhaustive search methods from \emph{subgroups} do not support a feature-cardinality threshold, so we run them without it.

Finally, we generally leave the hyperparameters of all subgroup-discovery methods at their defaults except to adapt them to our main experiments.
For example, we always set WRAcc as the objective and use a beam width of $w=10$ for all beam-search methods.

\paragraph{Experimental task}

We focus on finding competitors with a reasonable runtime.
To this end, each experimental task in the competitor-runtime experiments combines one subgroup-discovery method with one cross-validation fold of a dataset (in five-fold cross-validation).
Within each experimental task, we sequentially measure the runtime for a feature-cardinality threshold of $k \in \{1, 2, 3, 4, 5\}$.
In particular, we exclude the unconstrained setting, which may take significantly longer.
We run these experiments separately for each dataset and grant each task 9~hours of runtime to process these five cardinality settings.
This runtime limit is considerably higher than the maximum solver timeout of 2048~s in our main experiments, even though the latter timeout applies to each cardinality setting individually instead of jointly.

\paragraph{Datasets}

To accompany potentially slow subgroup-discovery methods, we employ smaller datasets than in our main experiments.
In particular, we use five of the smallest datasets from our main experiments, all with $m < 500$ data objects and $n < 50$ features, plus the even smaller \emph{iris} dataset with $m = 150$ and $n = 4$.
Table~\ref{tab:csd:competitor-runtime-datasets} lists these datasets.
For comparison: The maxima in our main experiments are considerably higher, i.e., $m = 9822$ and $n = 168$ (cf.~Table~\ref{tab:csd:datasets}).

\paragraph{Implementation}

We implemented a dedicated experimental pipeline for the competitor-runtime experiments.
This pipeline parallelizes over the experimental tasks.
The code and data are available together with the main experiments (cf.~Section~\ref{sec:csd:experimental-design:implementation}).
We used the same hardware but Python~3.9 instead of Python~3.8 to support recent versions of all subgroup-discovery packages.

\subsubsection{Evaluation}
\label{sec:csd:appendix:competitor-runtime:evaluation}

\begin{table}[t]
	\centering
	\caption{
		Which subgroup-discovery methods finished their experimental task (feature-cardinality thresholds $k \in \{1, 2, 3, 4, 5\}$ evaluated sequentially) within 9~hours on which dataset?
	}
	\begin{tabular}{lllllll}
		\toprule
		Method & back. & horse. & iono. & iris & spect & spectf \\
		\midrule
		csd.Beam & \checkmark & \checkmark & \checkmark & \checkmark & \checkmark & \checkmark \\
		csd.SMT & \checkmark & \checkmark & \checkmark & \checkmark & \checkmark &  \\
		pysubdisc.Beam & \checkmark & \checkmark & \checkmark & \checkmark & \checkmark & \checkmark \\
		pysubdisc.BestFirst & \checkmark & \checkmark &  & \checkmark & \checkmark &  \\
		pysubdisc.BreadthFirst & \checkmark & \checkmark &  & \checkmark & \checkmark &  \\
		pysubdisc.DepthFirst & \checkmark & \checkmark &  & \checkmark & \checkmark &  \\
		pysubgroup.Apriori &  &  &  & \checkmark & \checkmark &  \\
		pysubgroup.Beam & \checkmark & \checkmark & \checkmark & \checkmark & \checkmark & \checkmark \\
		pysubgroup.DepthFirst &  &  &  & \checkmark & \checkmark &  \\
		pysubgroup.GPGrowth &  &  &  & \checkmark &  &  \\
		sd4py.BSD & \checkmark & \checkmark & \checkmark & \checkmark & \checkmark & \checkmark \\
		sd4py.Beam & \checkmark & \checkmark & \checkmark & \checkmark & \checkmark & \checkmark \\
		sd4py.SDMap & \checkmark & \checkmark & \checkmark & \checkmark & \checkmark & \checkmark \\
		subgroups.BSD & \checkmark & \checkmark &  & \checkmark & \checkmark &  \\
		subgroups.SDMap &  &  &  & \checkmark &  &  \\
		subgroups.SDMapStar &  &  &  & \checkmark &  &  \\
		subgroups.VLSD &  &  &  & \checkmark &  &  \\
		\bottomrule
	\end{tabular}
	\label{tab:csd:competitor-timeouts}
\end{table}

\paragraph{Timeouts}

Table~\ref{tab:csd:competitor-timeouts} displays which experimental tasks finished within the 9-hour timeout.
Generally, each subgroup-discovery method finished either on all or no cross-validation folds of a dataset.
All methods finished on the \emph{iris} dataset, but fewer methods on the five other datasets.
From the exhaustive search methods, i.e., without beam search, only \emph{sd4py.BSD} and \emph{sd4py.SDMap} finished on each of the six small datasets.

\begin{table}[t]
	\centering
	\caption{
		Runtime (in seconds) on the dataset~\emph{spect}, averaged over five cross-validation folds.
		Missing subgroup-discovery methods compared to Table~\ref{tab:csd:competitor-timeouts} did not finish their experimental task within 9~hours.
	}
	\begin{tabular}{lrrrrr}
		\toprule
		Method & $k=1$ & $k=2$ & $k=3$ & $k=4$ & $k=5$ \\
		\midrule
		csd.Beam & 0.03 & 0.05 & 0.06 & 0.08 & 0.09 \\
		csd.SMT & 3.58 & 3.92 & 3.63 & 3.49 & 3.50 \\
		pysubdisc.Beam & 2.40 & 0.20 & 0.17 & 0.32 & 0.26 \\
		pysubdisc.BestFirst & 1.61 & 0.27 & 2.12 & 26.39 & 829.56 \\
		pysubdisc.BreadthFirst & 2.35 & 0.33 & 1.79 & 6.23 & 47.98 \\
		pysubdisc.DepthFirst & 1.96 & 0.29 & 1.95 & 27.60 & 948.02 \\
		pysubgroup.Apriori & 0.02 & 0.07 & 0.81 & 8.93 & 70.20 \\
		pysubgroup.Beam & 0.02 & 0.12 & 0.23 & 0.35 & 0.50 \\
		pysubgroup.DepthFirst & 0.02 & 0.23 & 3.65 & 45.06 & 412.13 \\
		sd4py.BSD & 0.36 & 0.66 & 0.07 & 0.10 & 0.18 \\
		sd4py.Beam & 0.45 & 0.85 & 0.26 & 0.27 & 0.19 \\
		sd4py.SDMap & 0.45 & 0.89 & 0.29 & 0.55 & 0.49 \\
		subgroups.BSD & 0.15 & 0.86 & 6.40 & 36.90 & 166.24 \\
		\bottomrule
	\end{tabular}
	\label{tab:csd:spect-runtime}
\end{table}

\paragraph{Runtime}

For a more detailed view, Table~\ref{tab:csd:spect-runtime} displays the exact runtimes for the dataset~\emph{spect}, which had the second-most methods finishing within the 9-hour timeout.
This table clearly shows an exponential runtime increase over the feature-cardinality threshold~$k$ for the exhaustive search methods from the packages \emph{pysubdisc}, \emph{pysubgroup}, and \emph{subgroups}, while \emph{sd4py} exhibits considerably smaller and less varying runtimes.

\paragraph{Conclusions}

Overall, these competitor-runtime experiments show that only the exhaustive search methods from \emph{sd4py} can be expected to have a reasonable runtime in our main experiments.
In particular, the latter involve larger datasets and also an unrestricted feature-cardinality threshold, which corresponds to~$k = n \geq 20$ (number of features in the dataset, which is at least~20).
Thus, we integrated \emph{sd4py}'s \emph{BSD}~\cite{lemmerich2010fast} and \emph{SDMap}~\cite{atzmueller2006sd} into our main experiments (cf.~Section~\ref{sec:csd:experimental-design:methods}).
Both methods retain the potential weakness that they require discretizing numeric features, which reduces their search space.
Thus, they may yield a lower quality than an exhaustive search on the full space but also be faster.
In our main experiments, we rely on \emph{sd4py}'s built-in equal-width discretization but optimize subgroup quality over ten different numbers of bins.

\renewcommand*{\bibfont}{\small} 
\printbibliography

\end{document}